\documentclass[journal]{IEEEtran}

\ifCLASSINFOpdf
   \usepackage[pdftex]{graphicx}
\else

\fi

\usepackage{graphicx}
\usepackage{amsmath,amssymb} 
\usepackage{color}
\usepackage{subfigure}
\usepackage{epstopdf}
\usepackage{epsfig}
\usepackage{float}
\usepackage{makecell}
\begin{document}

\title{Constrained Manifold Learning for Hyperspectral Imagery Visualization}

\author{Danping~Liao,
 Yuntao~Qian~\IEEEmembership{Member,~IEEE,}
 and Yuan Yan~Tang,~\IEEEmembership{Fellow,~IEEE}

\thanks {This work was supported by the National Natural Science Foundation of China 61571393,
the Research Grants of University of Macau MYRG2015-00049-FST, MYRG2015-00050-FST.}
\thanks{D. Liao and Y. Qian are with the Institute of Artificial Intelligence, College of Computer Science, Zhejiang University, Hangzhou 310027,
P.R. China. Corresponding author: Y. Qian (ytqian@zju.edu.cn)}
\thanks{Y. Y. Tang is with the Faculty of Science and Technology, University of
Macau, Macau 999078, China.}
}
\markboth{IEEE Journal of Selected Topics in Applied Earth Observations and Remote Sensing}%
{~~}

\maketitle

\begin{abstract}
Displaying the large number of bands in a hyperspectral image (HSI) on a trichromatic monitor is important for HSI processing and analysis system.
The visualized image shall convey as much information as possible
from the original HSI and meanwhile facilitate image interpretation.
However, most existing methods display HSIs in false color, which contradicts with user experience and expectation.
In this paper, we propose a visualization approach based on constrained manifold learning, whose goal is to learn a visualized image that not only preserves the manifold structure of the HSI but also has natural colors.
Manifold learning preserves the image structure by forcing pixels with similar signatures to be displayed with similar colors.
A composite kernel is applied in manifold learning to incorporate both the spatial and spectral information of HSI in the embedded space.
The colors of the output image are constrained by a corresponding natural-looking  RGB image, which can either be  generated from the HSI itself (e.g., band selection from the visible wavelength) or be captured by a separate device.
Our method can be done at instance-level and feature-level.
Instance-level learning directly obtains the RGB coordinates for the pixels in the HSI while feature-level learning learns an explicit mapping function from the high dimensional spectral space to the RGB space.
Experimental results demonstrate the advantage of the proposed method in information preservation and natural color visualization.
\end{abstract}

\begin{IEEEkeywords}
Hyperspectral image, visualization, manifold learning, composite kernel
\end{IEEEkeywords}

\IEEEpeerreviewmaketitle

\section{Introduction}

Hyperspectral imaging sensors acquire images with tens or hundreds of light wavelength indexed bands, which enable more accurate target detection and classification.
In order  for human observers to easily interpret and analyze hyperspectral images (HSIs), it is crucial that the displayed HSIs on the output device best enable human interaction with them.
However, the large number of bands in an HSI is far beyond the capability of a trichromatic display device.
A common solution is to consider HSI visualization as a dimension reduction problem where the number of bands in an HSI is reduced to 1 for a representative grayscale output or 3 for color visualization.

A simple way to visualize an HSI is to average all bands to produce a grayscale image.
This approach preserves the basic scene structure but suffers from \textit{metamerism},
where different high dimensional pixel values (or called spectral signatures) are assigned with the same output intensity.
To alleviate metamerism, a better way is to present an HSI as a color image.
A straightforward way for color representation is to select three of the original bands as R, G and B composites.
Some softwares provide interactive tools for users to manually pick three bands to display~\cite{MultiSpec,ENVI}.
More sophisticated band selection methods~\cite{demir2009low,le2011constrained,su2014hyperspectral} aim to highlight expected features so that human perceptual bands or the most informative bands are selected for visualization.

Band selection methods only take the selected spectrum into account.
As a result, the information in other bands is ignored.
To preserve the information across the full wavelength range, some HSI visualization approaches use feature transformation to condense the original spectral bands into three new bands.
Several classic linear dimension reduction methods such as independent component analysis (ICA)~\cite{zhu2007evaluation} and  principal component analysis (PCA)~\cite{tyo2003principal,tsagaris2005fusion} have been applied to map the HSI to a 3-D subspace, whose basis is then rotated so that the final 3-D coordinates form a plausible RGB image.

\begin{figure*}[t]
\centering
\subfigure[]{
\includegraphics[width=0.48\linewidth]{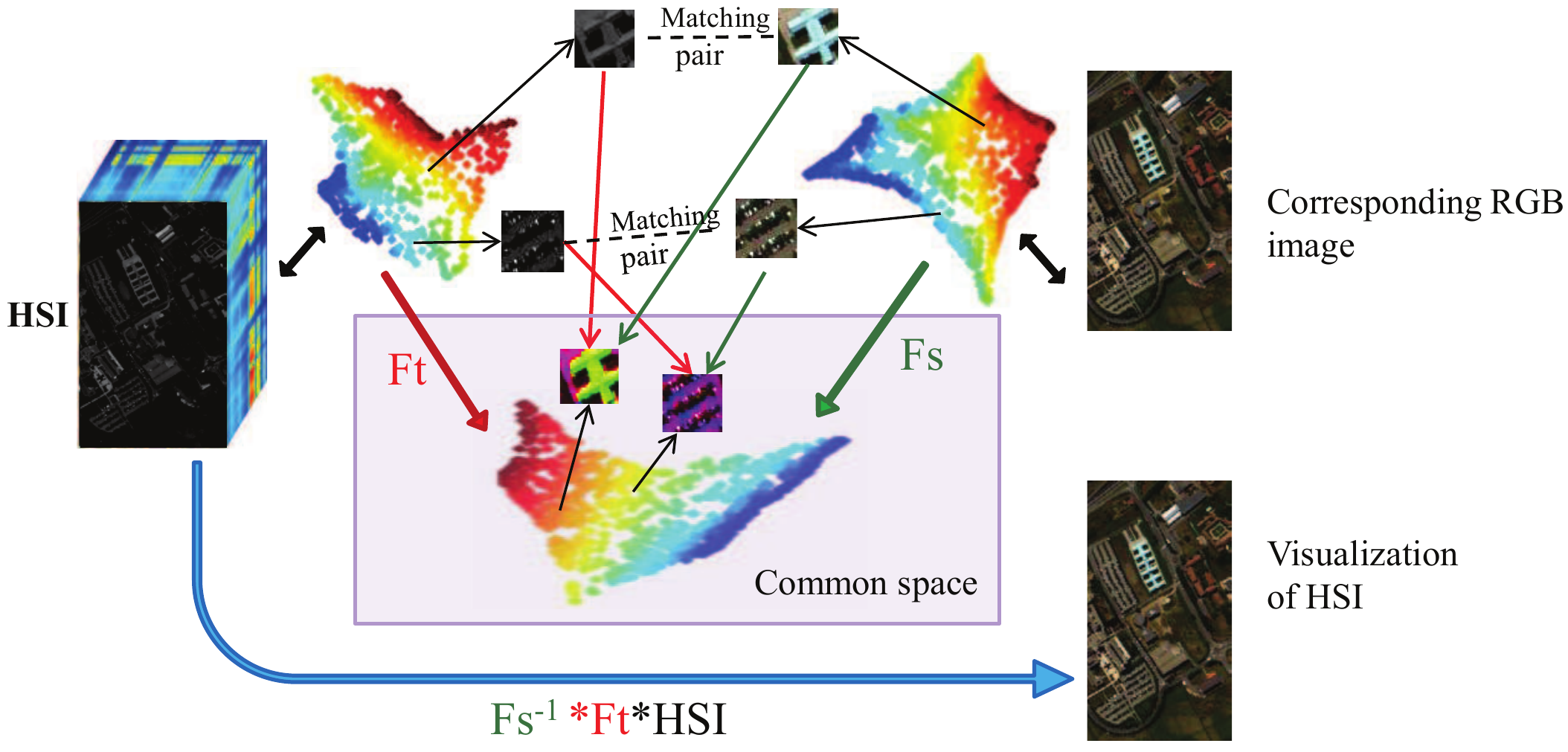}
\label{maworkflow}}
\subfigure[]{
\includegraphics[width=0.48\linewidth]{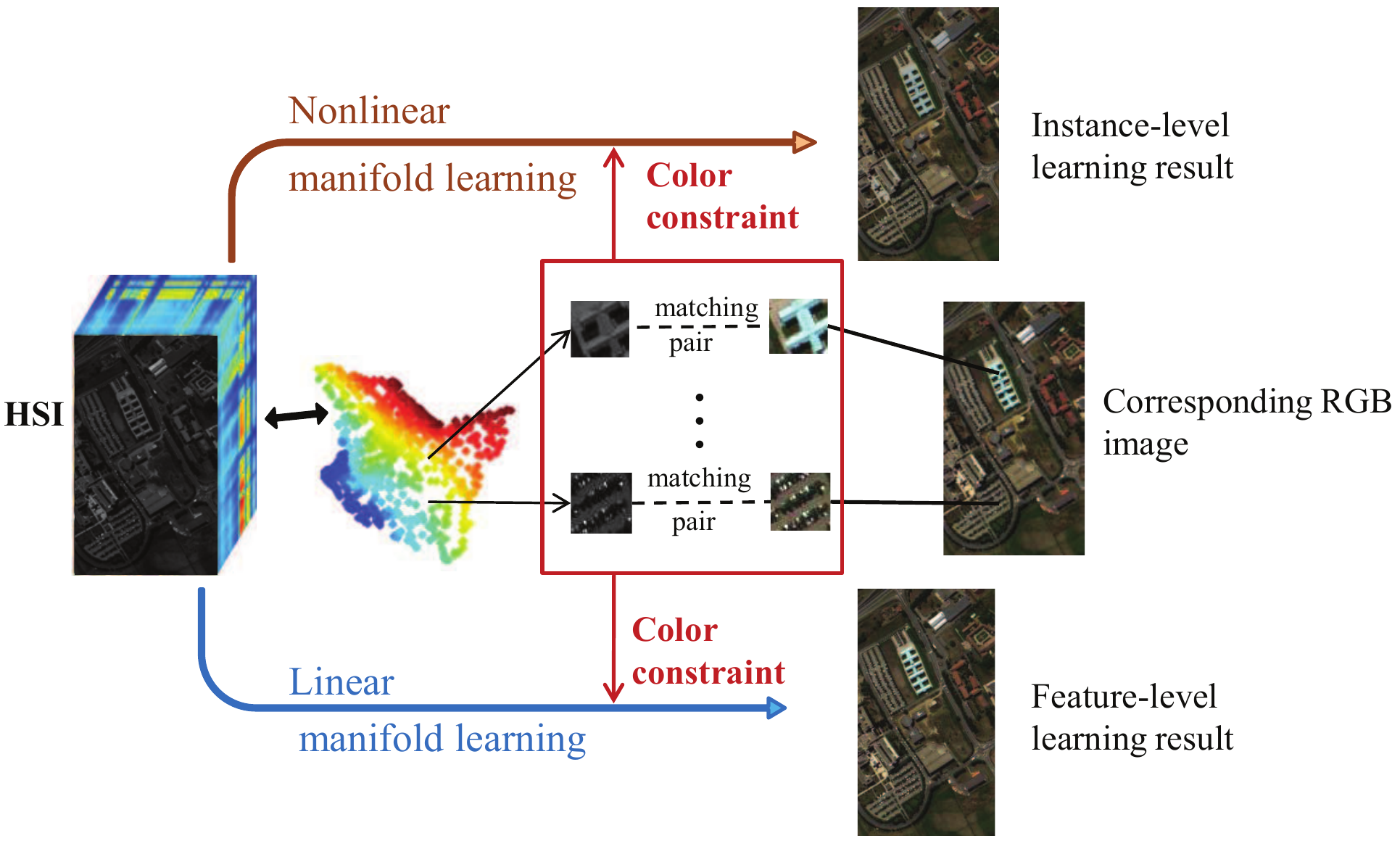}
\label{cml_workflow}}
\caption{The frameworks of two manifold learning-based methods for HSI visualization.
(a) manifold alignment model. (b) The proposed model.}
\end{figure*}

Linear methods do not consider the nonlinear characteristics of the hyperspectral data.
To model the nonlinear data structure in HSIs~\cite{bachmann2005exploiting}, manifold learning methods have been proposed to represent the topology of high dimensional HSIs in lower dimensions for visualization and dimension reduction.
Isometric feature mapping
(ISOMAP)~\cite{bachmann2005exploiting,najim2015fspe}, kernel principal component analysis (KPCA)~\cite{Sch2014Nonlinear}, Laplacian Eigenmaps~\cite{belkin2003laplacian}, Locality Preserving Projection (LPP)~\cite{niyogi2004locality} and locally linear embedding (LLE) ~\cite{Roweis2000Nonlinear,crawford2011exploring}  have received much attention because of their firm theoretical foundation associated with the kernel and eigenspectrum framework.

Similarly, some nonlinear visualization methods aim to preserve the pairwise distances
between pixels.
This task is usually posed as a constrained optimization problem.
Mignotte solved the optimization problem by a nonstationary Markov random field model~\cite{mignotte2010multiresolution}
and later extended this approach to preserve the spectral distances and separate dissimilar features~\cite{mignotte2012bicriteria}.
Edge information is also a significant local structure.
Kotwal and Chaudhuria used a nonlinear bilateral filter with the edge preserving characteristic to obtain the weights of bands at each pixel for band image fusion~\cite{kotwal2010visualization}.
Meka and Chaudhuri computed the weights at each pixel for band fusion to maintain edge information and to maximize the Shannon entropy of the fused image~\cite{Meka2015A}.

While most existing methods try to convey as much information as possible in the visualization, the HSIs are always displayed in false colors.
In general false-color visualizations are hard to interpret when object colors are very different from their natural appearance.
For example, one may be confused when the grass is shown in red or the sea is shown in yellow.
Moreover, most data-driven methods suffer the problem of ``inconsistent rendering'', i.e., the same objects/materials being displayed with very different colors in different visualizations, which also hinders the interpretation of HSIs.
Therefore, ``natural palette'' and ``consistent rendering'' gradually become two important criteria for evaluating HSI visualization.

Jacobson \emph{et al}.~\cite{jacobson2005design,jacobson2007linear} proposed a fixed linear spectral weighting envelope to generate consistent natural-looking images.
The spectral weighting envelope is a stretched version of the
CIE 1964 tristimulus color matching functions in the visible range, which represents human sensitivity to wavelength.
However, the stretched color matching function (CMF) is too simple to represent the
complex physical mechanism of spectral imaging, which makes it only
applicable to some specific hyperspectral imaging sensors.

Another approach for displaying HSIs with natural colors was proposed by Connah \emph{et al}.~\cite{connah2014spectral}, which takes advantage of a corresponding RGB image to produce a natural-looking output image.
The method preserves the image structure by mapping the structure tensor of the HSI exactly to the  output image.
To generate natural colors, the gradient of each pixel in the output image is constrained to be the same with that of its matching pixel in the corresponding RGB image.
This method requires accurate pixel-wise matching between the HSI and the corresponding RGB image, which might be difficult to achieve especially when the two images are acquired by sensors mounted on different platforms, such as the airplane and satellite, with different geometrical distortions.

We previously proposed a manifold alignment method to visualize HSI with natural colors~\cite{liaomanifold}.
It makes use of the color and structure information of a corresponding high resolution color image (HRCI) to generate natural colors and fine details in the visualization.
The workflow of the manifold alignment approach is shown in Fig.~\ref{maworkflow}. The model first computes the manifold structures of an input HSI and a corresponding HRCI, and then projects the two images to a common embedding space by manifold alignment, where the matching pixels of the two manifolds are aligned.
Finally the embedding of HSI is mapped from the common space to the RGB space to generate an natural-looking image.

In this paper, we propose a constrained manifold learning approach.
 The method takes advantage of the color information of a corresponding natural-looking RGB image, which may either be generated from the HSI itself (e.g. stacking three channels from 450-515nm (blue), 525-605 nm (green), and 630-690 nm (red)) or be captured by a different sensor (e.g. a RGB camera).
Our goal is to learn a visualized image that not only preserves the manifold structure of the input HSI but also shares the natural colors with the corresponding RGB image.
To achieve this, we combine a local geometry-preserving manifold learning method with a color constraint.
Manifold learning technique enables the pixels with similar spectral signatures in an HSI to be displayed with similar colors.
The color constraint is provided by forcing pixels in the output image to have similar colors with their matching pixels in the corresponding RGB image.
Combining manifold learning with the color constraint allows our method to generate a final visualization that facilitates human understanding.

The proposed constrained manifold learning can be done at instance-level and feature-level, where the ``instance'' refers to pixels and the ``feature'' refers to spectral features in different spaces.
Fig.~\ref{cml_workflow} shows the framework of the proposed model.
For illustration purpose, the matching pixels are represented by the patches centering at the pixels.
Instance-level learning is nonlinear. It directly computes the RGB coordinates of each pixels in an HSI without assuming any specific transformations.
Feature-level learning is linear. It estimates an explicit linear mapping function from the high dimensional spectral space to the RGB space.
The mapping function can be directly applied to visualize other HSIs captured by the same sensor.

The proposed method shares a similar idea with the manifold alignment approach~\cite{liaomanifold}.
The difference is that our method only makes use of the color information of the corresponding RGB image, thus the computation of the RGB image's manifold structure in the manifold alignment model is no longer needed, which makes our method more efficient.
Moreover, since manifold alignment attempts to fuse two image structures, an HRCI registered precisely is required in order to obtain a desirable result, which limits its usage in practical applications.
The same problem occurs to Connah's approach~\cite{connah2014spectral}, in which accurate pixel-wise matching between the HSI
and the corresponding RGB image is essential.
Different from these two methods, our method is more generic as it only requires a small set of matching pairs representing identical or similar class of objects/materials for color constraints. Hence, the color information can be provided by the corresponding RGB images obtained via various sources, or can even be provided by some color scribbles marked on each land cover class on the HSI.

Manifold learning methods consider the pixels to be statistically independent, ignoring the spatial relationships among them, however
the spatial contextual information of HSIs has been demonstrated to be very helpful in HSI processing~\cite{kim2008spatially,Fauvel2012A}. To take advantage of the spatial information, joint spectral-spatial methods such as Markov random Fields, vectors with stacked spectral-spatial features and morphological profiles were proposed (See~\cite{fauvel2013advances} for a comprehensive review).
In this paper, a composite kernel~\cite{camps2006composite} is applied in constructing the manifold structure of an HSI to incorporate its spectral and spatial information.
The resulting weight adjacency matrix fuses the spectral similarity and spatial similarity between pixels, allowing
better structure preservation in the visualization of the HSIs.

The rest of the paper is organized as follows.
Section~\ref{SectionManifoldLearning} briefly introduces Laplacian Eigenmaps and locality preservation projections (LPP), on which the proposed instance-level learning and feature-level learning are based, respectively.
Section~\ref{sec:Semisupervised manifold learning for HSI visualization} presents the proposed instance-level and feature-level constrained manifold learning algorithms for HSI visualization.
The experimental results and analysis are given in Section~\ref{Experiments}.
Finally, the conclusions are drawn in Section~\ref{Conclusions}.

\section{Manifold Learning for Dimension Reduction}
\label{SectionManifoldLearning}

Manifold learning is an effective dimension reduction method to extract nonlinear structures from high dimensional data, which has been widely applied in data visualization, feature extraction, denoising, and distance metrics.
Generally, the goal of manifold learning is to map a $p$-dimensional data set $X$ to a lower $q$-dimensional data set $Y$ while preserving the intrinsic geometry of the original manifold as much as possible.
Various algorithms such as ISOMAP~\cite{tenenbaum2000global}, LLE~\cite{Roweis2000Nonlinear}, Laplacian Eigenmaps~\cite{belkin2003laplacian}, LPP~\cite{niyogi2004locality} and Hessian LLE~\cite{donoho2003hessian} have their own representations for a manifold's geometry.
In this paper, the proposed instance-level and feature-level learning methods for HSI visualization are based on Laplacian Eigenmaps and LPP, respectively.

\subsection{Laplacian Eigenmaps}

Assume a set of data is represented by a matrix $\textbf{X}\in R^{p\times n}$ where $p$ is the number of features and $n$ is the number of samples.
Let $\textbf{X}_i$ denotes the $i$-th sample in $\textbf{X}$.
Both Laplacian Eigenmaps and LPP aim to find a dataset $\textbf{Y}\in R^{q\times n}$ such that $\textbf{Y}_i$ ``represents'' $\textbf{X}_i$ in the $q$ dimensional space with minimum structure loss.
The basic premise is that neighboring points in the original space should still stay close in the new space.

The first step of  manifold learning method is to construct a weighted graph $G = (V, E)$ to represent the manifold  of $\textbf{X}$.
Each node in $G$ represents a sample in the data set.
Let $\textbf{W} \in R^{n \times n}$ be the weighted adjacency matrix of $G$ where $\textbf{W}_{ij}$ is the weight of edge connecting $\textbf{X}_i$ and $\textbf{X}_j$.
Determining which nodes are connected on the
graph is based on the distances between points in the input space.
There are two variations to construct the edges of graph.
One is to connect each node to all the nodes within some fixed distance radius $\epsilon$, and the other connects each node to its $k$ nearest neighbors.
In this paper, the latter method is used, i.e., two nodes $i$ and $j$ are connected if $\textbf{X}_i$
is among the $k$ nearest neighbors of $\textbf{X}_j$ or $\textbf{X}_j$ is among the $k$ nearest neighbors of $\textbf{X}_i$.
The weight of an edge is computed by the heat kernel~\cite{Berline2003Heat}.
The weighted adjacency matrix $\textbf{W}$ can be constructed by
\begin{equation}
\label{W}
   \textbf{W}_{ij}=
    \begin{cases}
       K(\textbf{X}_i,\textbf{X}_j)& \text{if node $i$ and node $j$ are connected,}\\
    0 & \text{otherwise.}\\
    \end{cases}
\end{equation}
where
\begin{equation}
K(\textbf{X}_i,\textbf{X}_j)=e^{-\frac{\|\textbf{X}_i-\textbf{X}_j\|^2}{t}}.
\end{equation}
As will be illustrated in a later section, the adjacency matrix can be modified to include the spatial context to better preserve image information.

The objective of Laplacian Eigenmaps is to find the optimal $\textbf{Y}$ to minimize the following function:
\begin{equation}
\label{LE}
f(\textbf{Y})=\frac{1}{2}\sum_{ij}(\textbf{Y}_i - \textbf{Y}_j)^2\textbf{W}_{ij}.
\end{equation}
Minimizing the objective guarantees neighboring samples in the input space still stay close after dimension reduction.
Equation~(\ref{LE}) can be further rewritten in matrix form:
\begin{equation}
\begin{split}
f(\textbf{Y})&=\sum_{i}(\textbf{Y}_i\textbf{D}_{ii}\textbf{Y}_i^T)-\sum_{ij}(\textbf{Y}_i\textbf{W}_{ij}\textbf{Y}_i^T)\\
&=tr(\textbf{YDY}^T-\textbf{YWY}^T)\\
&=tr(\textbf{YLY}^T).
\end{split}
\end{equation}
$tr(\cdot)$ denotes the trace of a matrix. $\textbf{L}=\textbf{D}-\textbf{W}$ is the Laplacian matrix of $G$ where $\textbf{D}$ is a diagonal matrix with $\textbf{D}_{ii}=\sum_j\textbf{W}_{ij}$.

The optimal $\textbf{Y}$ is given by the matrix of eigenvectors corresponding to the $q$ lowest eigenvalues of the following generalized eigenvalue problem
\begin{equation}
\textbf{LY}=\lambda \textbf{DY}.
\end{equation}

\subsection{Locality Preservation Projections (LPP)}

LPP is a linear approximation of Laplacian Eigenmaps.
Its objective is to find an explicit linear transformation matrix $\textbf{F}$ of size $p\times q$ to map the data set from the original $p$-dimensional space to a $q$-dimensional space.

The objective function of LPP is formalized as
\begin{equation}
\label{LPPlossfunction}
 g(\textbf{F})=\sum_{ij} \|\textbf{F}^T\textbf{X}_i-\textbf{F}^T\textbf{X}_j\|^2 \textbf{W}(i,j)
\end{equation}
where $\textbf{F}^T\textbf{X}_i$ is the representation of $\textbf{X}_i$ in the low dimensional space.
Equation~(\ref{LPPlossfunction}) can be rewritten in matrix form as
\begin{equation}
\begin{split}
   g(\textbf{F})&=\sum_{i}(\textbf{F}^T\textbf{x}_i\textbf{D}_{ii}\textbf{x}_i^T\textbf{F})-\sum_{ij}(\textbf{F}^T\textbf{x}_i\textbf{W}_{ij}\textbf{x}_i^T\textbf{F})\\
   &=tr(\textbf{F}^T\textbf{X}\textbf{D}\textbf{X}^T\textbf{F}-\textbf{F}^T\textbf{X}\textbf{W}\textbf{X}^T\textbf{F})\\
   &=tr(\textbf{F}^T\textbf{X}\textbf{L}\textbf{X}^T\textbf{F}).
   \end{split}
 \end{equation}
The solution can be found by solving the following generalized eigenvalue problem:
\begin{equation}
\label{generalized eigenvector problem}
\textbf{X} \textbf{L}\textbf{X}^T\textbf{F}=\lambda \textbf{X} \textbf{D}\textbf{X}^T\textbf{F}.
\end{equation}
Since the matrices $\textbf{X} \textbf{L}\textbf{X}^T$ and $\textbf{X} \textbf{D}\textbf{X}^T$ are symmetric and positive semidefinite, the optimal mapping $\textbf{F}$ is constructed by the $q$ minimum generalized eigenvectors.

\section{Constrained Manifold Learning for HSI Visualization}
\label{sec:Semisupervised manifold learning for HSI visualization}

The goal of the proposed constrained manifold learning approach is to display the HSI with natural tones to facilitate human interpretation and meanwhile preserve the manifold structure of the HSI as much as possible.
The key idea is to combine the manifold structure of the HSI and the color information from a corresponding natural-looking RGB image to generate a new 3-D embedding space for visualization.
More specifically, the method projects an input HSI to the RGB space by manifold learning and meanwhile constrains the colors of the pixels in the output image to be similar with their matching pixels in the corresponding RGB image.
This allows the colors to propagate from the matching pixels to the rest of the image while preserving the manifold structure.
The proposed constrained manifold learning can be done at instance-level and feature-level based on Laplacian Eigenmaps and LLE, respectively.

Assume an input HSI and a corresponding RGB image are represented by two matrices $\textbf{X}\in R^{p\times n}$ and $\textbf{S}\in R^{q\times m}$, respectively, where $p$ and $q$ are the numbers of spectral bands, and $n$ and $m$ are the numbers of pixels.
The RGB image has three channels, thus $q=3$.
The first step of the proposed method is to obtain the weighted adjacency matrix $\textbf{W}\in R^{n\times n}$ to present the manifold structure of the HSI. Each entry in $\textbf{W}$ represents the similarity between a pair of pixels in the high dimensional space.
The next step is to construct the correspondence matrix $\textbf{C} \in R^{n\times m}$ to model the pixel-correspondence between the HSI and the RGB image.
We will introduce the construction of $\textbf{W}$ and  $\textbf{C}$ along with the proposed instance-level learning and feature-level learning in this section.

\subsection{Construction of Weighted Adjacency Matrix}

The heat kernel as in Equation~(\ref{W}) is commonly used to compute $\textbf{W}$.
Such kernel only takes advantage of the spectral information and ignores the spatial correlation between pixels.
In \cite{camps2006composite}, a family
of composite kernels were proposed accounting for the spatial, spectral, and cross-information between pixels or objects.
In these kernels, a pixel entity is redefined simultaneously both in the spectral domain using its spectral content and in the spatial domain using feature extraction on its surrounding area.

In this paper, a weighted summation kernel is utilized, which is a composite kernel balancing the spatial and spectral content.
Let $x_i^s$ represent the spectral feature of pixel $i$, and $x_i^w$ represent the spatial feature.
Assume that we apply kernel $K_s$ on the spectral feature and $K_w$ on the spatial feature.

According to Mercer's theorem, the weighted summation of two kernels is still a valid kernel.
The combined kernel is represented as
\begin{equation}
\label{composite}
   K(x_i,x_j)= \mu K_s (x_i^s,x_j^s)+ (1-\mu)K_w(x_i^w, x_j^w)
\end{equation}
where $\mu\in[0,1]$ constitutes a trade-off between the spatial and spectral information.
In our experiment, the spatial features $x_i^w$ is result of a Gaussian filter in a given window around pixel $x_i$, and the spectral features $x_i^s$ is the actual spectral signature.
Both $K_s$ and $K_w$ are set to the radial basis function (RBF) kernel, which is defined as
\begin{equation}
\label{RBF}
   K_{rbf}(x,x')= e^{-\frac{\|x-x'\|^2}{2\delta^2}}.
\end{equation}
With the combined kernel, the entity of the weighted adjacency matrix $\textbf{W}$ is defined as
\begin{equation}
   \textbf{W}_{ij}= K(x_i,x_j).
\end{equation}
Each entity in $\textbf{W}$ is then a fusion of the spectral and spatial similarities.

\subsection{Construction of Correspondence Matrix}

A corresponding RGB image offers the color constraint for visualization.
This image can be obtained from the HSI itself using some natural-color rendering algorithms such as band selection from the visible wavelength (450-515 nm (blue), 525-605 nm(green), and 630-690 nm (red)), or by mapping the visible wavelength part onto a set of color matching functions.
The RGB image can also be captured by a separate device, such as a RGB camera, on the same scene with the HSI.
As the proposed method only makes use of its color information but not structure information, the RGB image can even be captured from a different but similar site with the HSI.

The correspondence between an HSI and a RGB image is represented as a matrix $\textbf{C} \in R^{n\times m}$ where
\begin{equation}
\label{C}
   \textbf{C}_{ij}=
    \begin{cases}
       1 & \text{if the $i$th pixel in the HSI and the $j$th pixel in }\\
       ~&\text{the RGB image form a matching pair,}\\
    0 & \text{otherwise.}\\
    \end{cases}
\end{equation}
To construct the correspondence matrix, we need to find out a set of matching pixels between the two images.
Depending on the source of the corresponding RGB image, the matching pairs can be obtained by different ways.

When the corresponding RGB image is generated from the HSI itself, the two images are already pixel-wise matched.
We can directly use a part/all of the pixels in the images as matching pairs.

If the corresponding RGB image is captured by a separate RGB camera on the same site as the HSI, the two images might have different geometrical distortions.
In this case, image registration is required to find their matching relation.
In general, the image registration method firstly finds a few matching pixel
pairs, and then uses them to estimate a geometric transformation model so that the two images are matched in the same coordinate system. Scale-invariant feature transform (SIFT)
is widely used to detect the matching pixels between
images due to its robustness to changes in scale, orientation
and illumination~\cite{lowe2004distinctive}.
To find a set of matching pixels, we first extract SIFT key-points from each band of the HSI and the RGB image. The most similar key-points are considered as matching pairs between the two images. These matching pairs are then used to estimate a projective transformation for image registration ~\cite{Szeliski2006}, which is defined as
\begin{equation}
(x^\prime, y^\prime, 1)^\mathrm{T} = \mathbf{H}(x, y, 1)^\mathrm{T},
\end{equation}
where $\{(x, y),(x^\prime, y^\prime)\}$ are the coordinates of a matching pair between HSI and the RGB image, and
\begin{equation}\label{}
   \mathbf{H}=\begin{pmatrix}
h_1 & h_2 & h_3 \\
h_4 & h_5 & h_6 \\
h_7 & h_8 & 1 \\
\end{pmatrix}
\end{equation}
is the transformation matrix, or called homogaphy matrix. If we find $n$ matching pairs $\{(x_i, y_i),(x_i^\prime, y_i^\prime)\}, i=1,\ldots,n$ by SIFT matching, $\mathbf{H}$ can be estimated by
\begin{equation}\label{eq:H}
    \min_{\mathbf{H}} \sum\limits_{i=1}^n \| (x_i^\prime, y_i^\prime, 1)^\mathrm{T} - \mathbf{H}(x_i, y_i, 1)^\mathrm{T} \|^2.
\end{equation}
However, SIFT feature-based matching is not always precise. The mismatched pairs will lead to inaccuracies in estimating $\mathbf{H}$.
To deal with this problem,  RANdom SAmple Consensus (RANSAC) technique~\cite{Szeliski2006,Brown2007} is widely used to rule out mismatched pairs and in this way produce a more robust transformation estimation.

If the HSI and the corresponding RGB image are captured from different sites,
pixels belong to similar type of objects/materials can be considered as matching pairs.
In this case, interactive tools can be developed for users to manually pick the matching pairs.

\subsection{Instance-level Constrained Manifold Learning}
\label{nonlinearsection}

The proposed instance-level manifold learning builds on Laplacian Eigenmaps.
It has the same goal as Laplacian Eigenmaps to preserve the local structure of the original data in the low dimensional data.
Besides, it constrains the output data to be aligned with a referencing data set.
In the task of HSI visualization, a corresponding RGB image is used as the referencing data set to produce natural colors.
Assume an input HSI and a corresponding RGB image are represented by two matrices $\textbf{X}\in R^{p\times n}$ and $\textbf{S}\in R^{q\times m}$.
Let matrix $\textbf{Y}\in R^{n\times 3}$ represents the visualized image to be obtained.
The objective of instance-level learning is to find the optimal $\textbf{Y}$ that minimizes the following function:
\begin{equation}
\label{lossfunction1}
 H(\textbf{Y})=\frac{1}{2}\sum_{ij} \|\textbf{Y}_i-\textbf{Y}_j\|^2 \times \textbf{W}_{ij}+\lambda \sum_{ik}\|\textbf{Y}_i-\textbf{S}_k\|^2 \times \textbf{C}_{ik}.
\end{equation}
The first term on the right hand side is the same with Laplacian Eigenmaps, which is to preserve the local structure of the manifold.
For HSI visualization, it encourages pixels with similar spectral signatures in an HSI to be presented with similar colors in the visualized image.

The second term is a constraint making the embedding $\textbf{Y}$ to be aligned with the referencing data set $\textbf{S}$.
For HSI visualization, this constraint forces pixels in the output image to have similar colors with their matching pixels in the corresponding RGB image. As $\textbf{C}_{ik}=1$ if $\textbf{Y}_i$ and  $\textbf{S}_k$ form a matching pair, to minimize the objective, $\textbf{Y}_i$ should be similar to $\textbf{S}_k$.
The two terms together allow the target colors to spread from the matching pixels to the rest of the image while preserving the manifold structure of the HSI.

The objective function is convex and differentiable. It can be rewritten as the following matrix form:
\begin{equation}
\label{lossfunctionnew1}
\begin{aligned}
 H(\textbf{Y})=tr(\textbf{YLY}^T)+
 \lambda~tr(\textbf{YC}_1\textbf{Y}^T+\textbf{SC}_2\textbf{S}^T-2\textbf{YCS}^T)
 \end{aligned}
\end{equation}
where $\textbf{C}_1$ is a diagonal matrix with $\textbf{C}_1(i,i) =\sum_j \textbf{C}(i,j)$.
$\textbf{C}_2$ is a diagonal matrix with $\textbf{C}_2(j,j) =\sum_i \textbf{C}(i,j)$.
$\textbf{L}$ is the Laplacian matrix of the input HSI, which can be obtained by
$\textbf{L}=\textbf{D}-\textbf{W}$
where $\textbf{D}$ is a diagonal matrix with $\textbf{D}(i,i) =\sum_j \textbf{W}(i,j)$.

The derivative of the function with respect to $\textbf{Y}$ is
\begin{equation}
\frac{\partial{H}}{\partial{\textbf{Y}}}=2\textbf{YL}+2\lambda \textbf{YC}_1-2\lambda \textbf{SC}^T.
\end{equation}
The optimal $\textbf{Y}$ can be obtained by setting the derivative to zero:
\begin{equation}
\label{instance_finalEq}
\textbf{Y}=\textbf{SC}^T(\frac{1}{\lambda}\textbf{L}+\textbf{C}_1)^{-1}.
\end{equation}

Instance-level learning is nonlinear and flexible to deal with various mappings without assuming any parametric model.
It has been shown that there are multiple sources of nonlinearity in HSIs~\cite{bachmann2005exploiting}, which indicates that nonlinear methods might be more competent than linear models in HSI dimension reduction and visualization.
However, the learning result is defined only on the pixels in the input HSI, and is hard to be generalized to visualize new pixels from other HSIs.
\begin{figure}[!tp]
\centering
\subfigure[]
{\includegraphics[angle=90,width=0.9\linewidth]{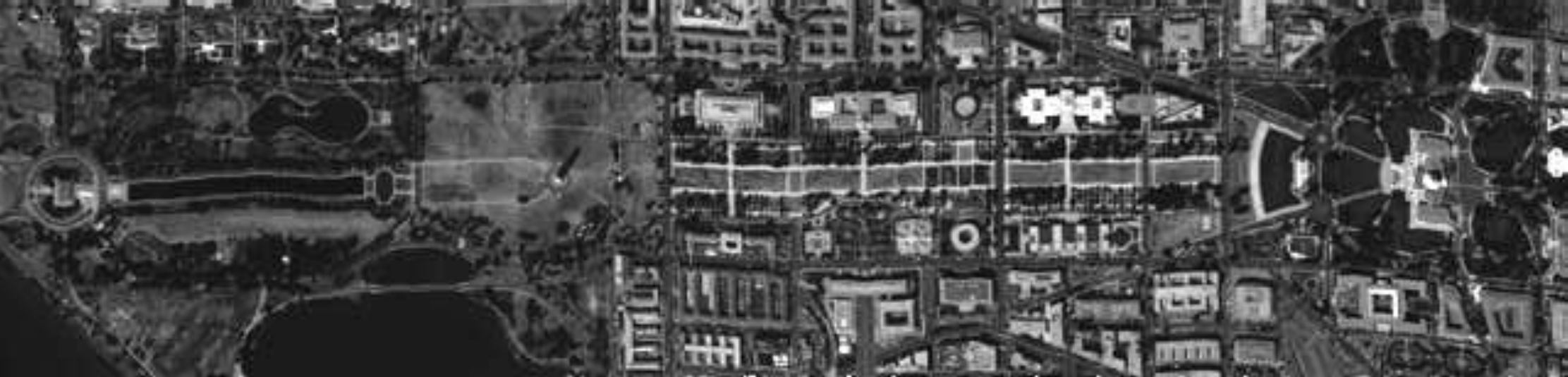}
\label{fig:band50}
}
\subfigure[]
{\includegraphics[angle=90,width=0.9\linewidth]{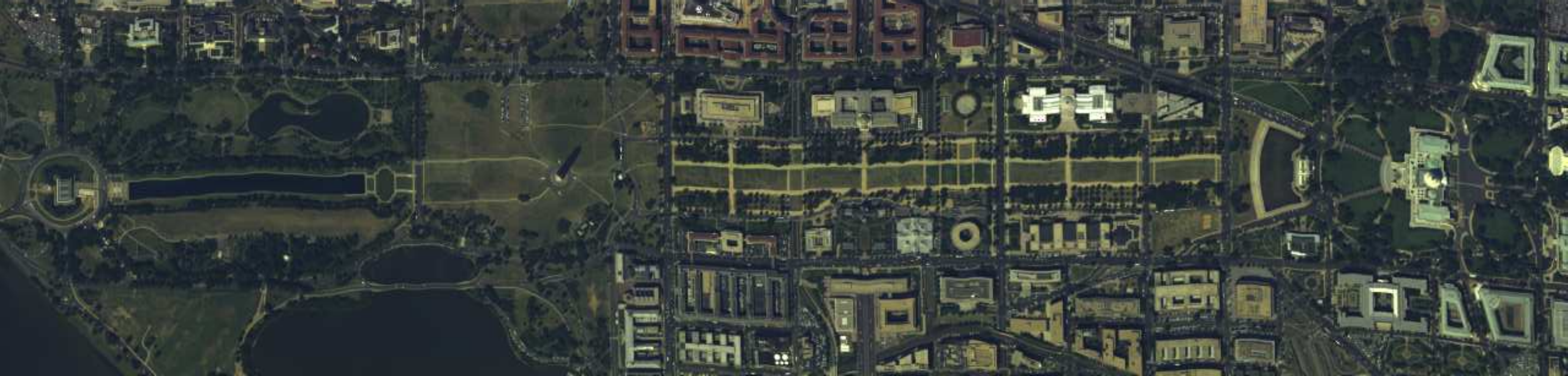}
\label{fig:registered}
}
\subfigure[]
{\includegraphics[width=0.3\linewidth]{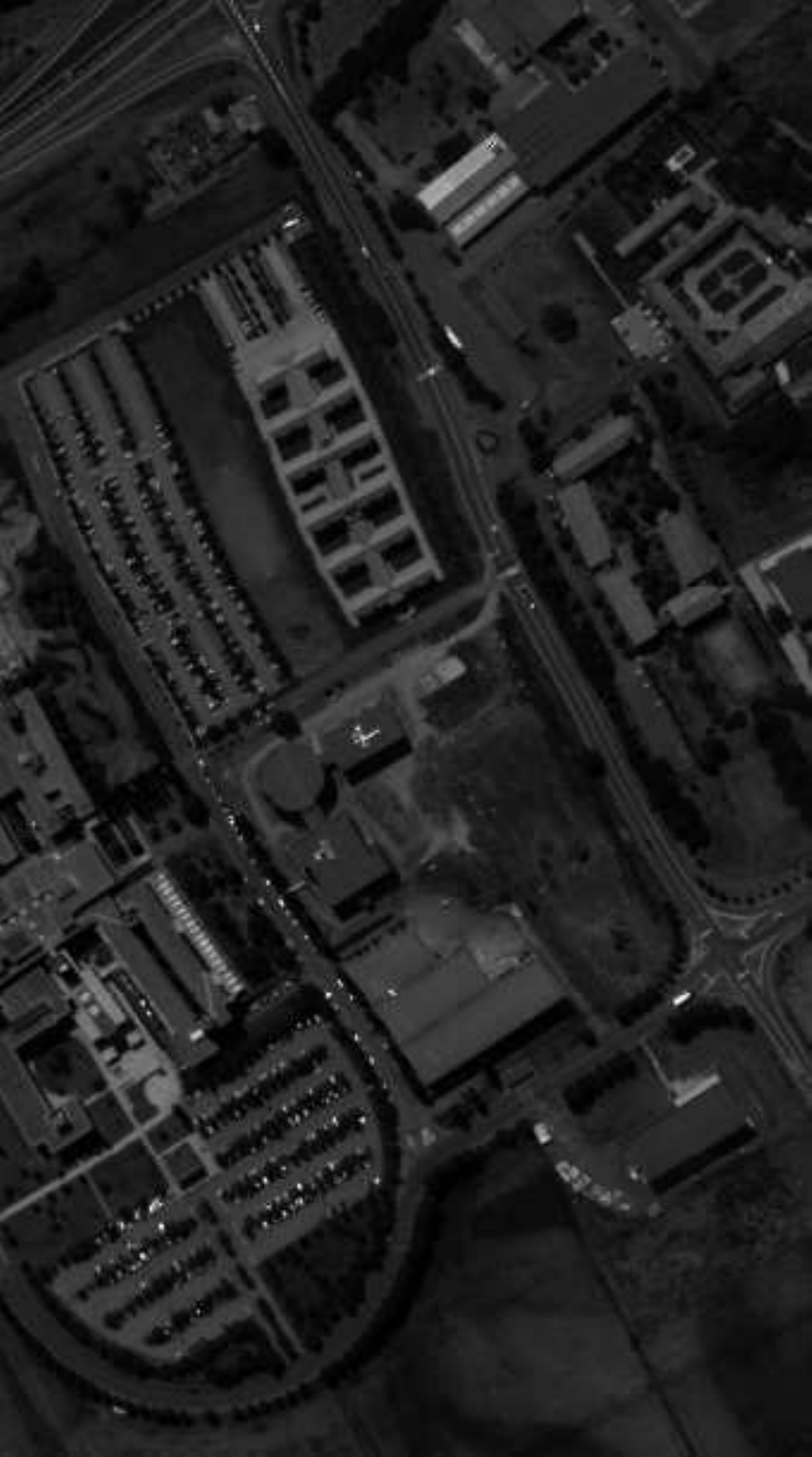}
\label{fig:PaviaUBand50}
}
\subfigure[]{
\includegraphics[width=0.3\linewidth]{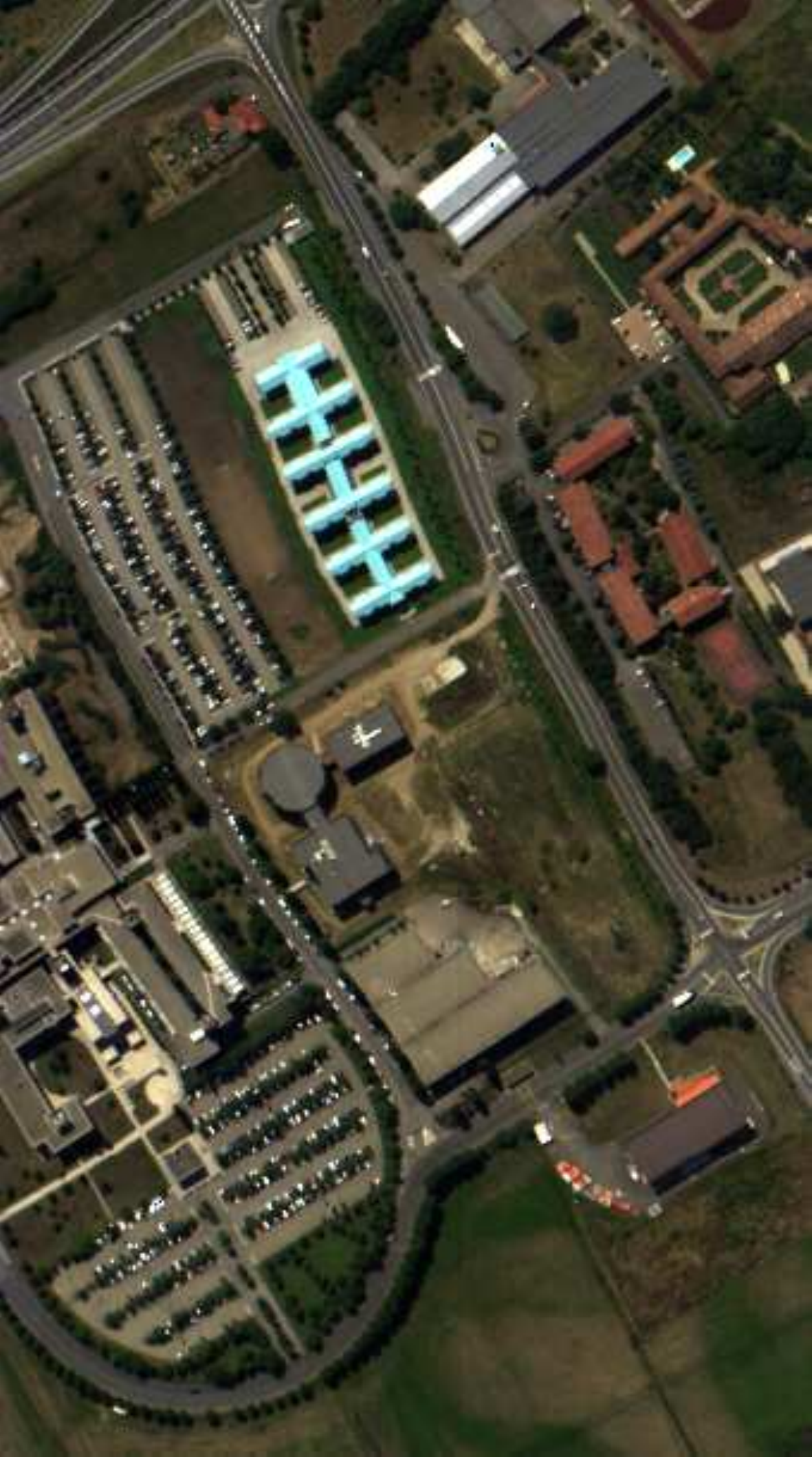}
\label{fig:RGBPaviaU}
}
\subfigure[]
{\includegraphics[width=0.45\linewidth]{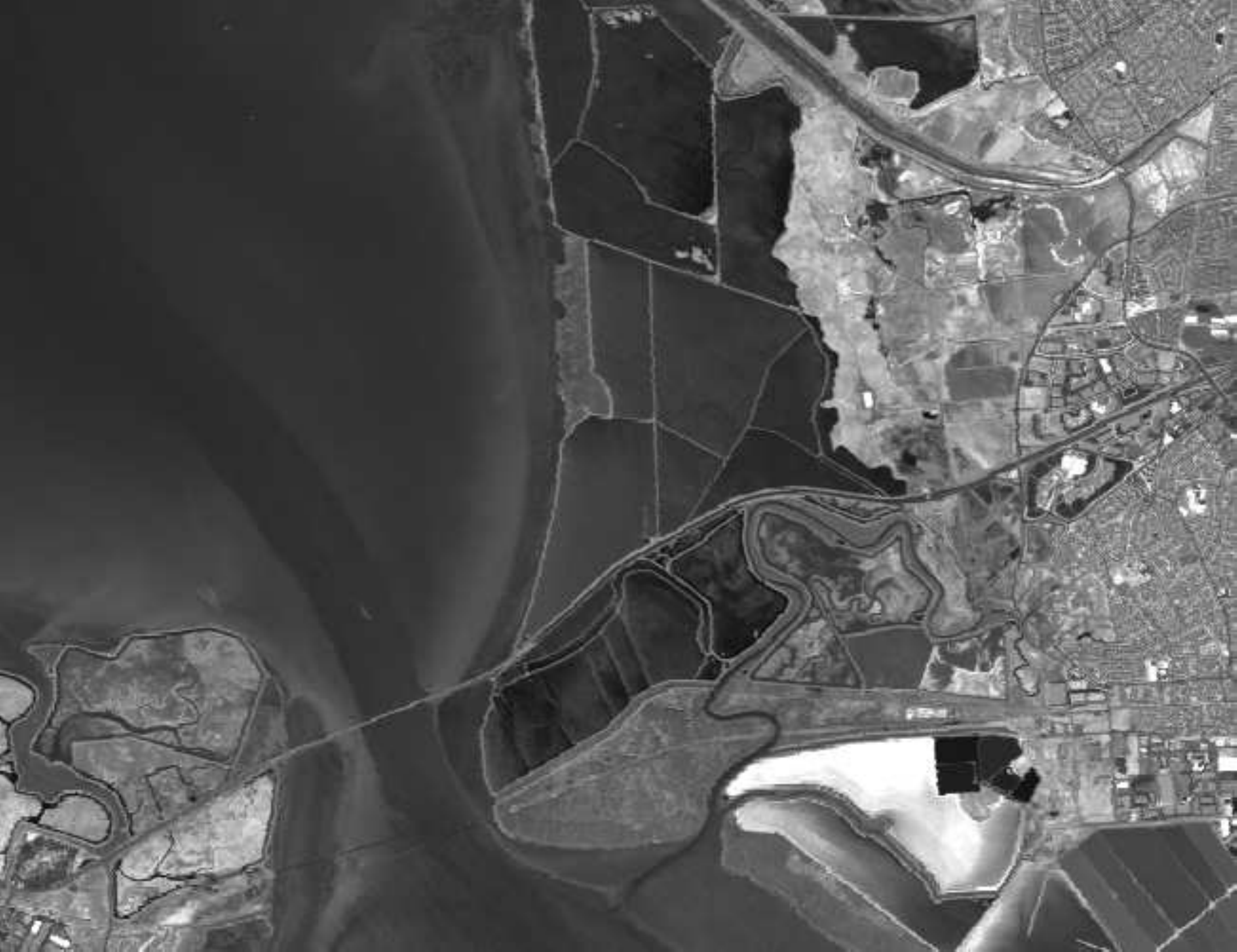}
\label{fig:MoffettBand50}
}
\subfigure[]{
\includegraphics[width=0.45\linewidth]{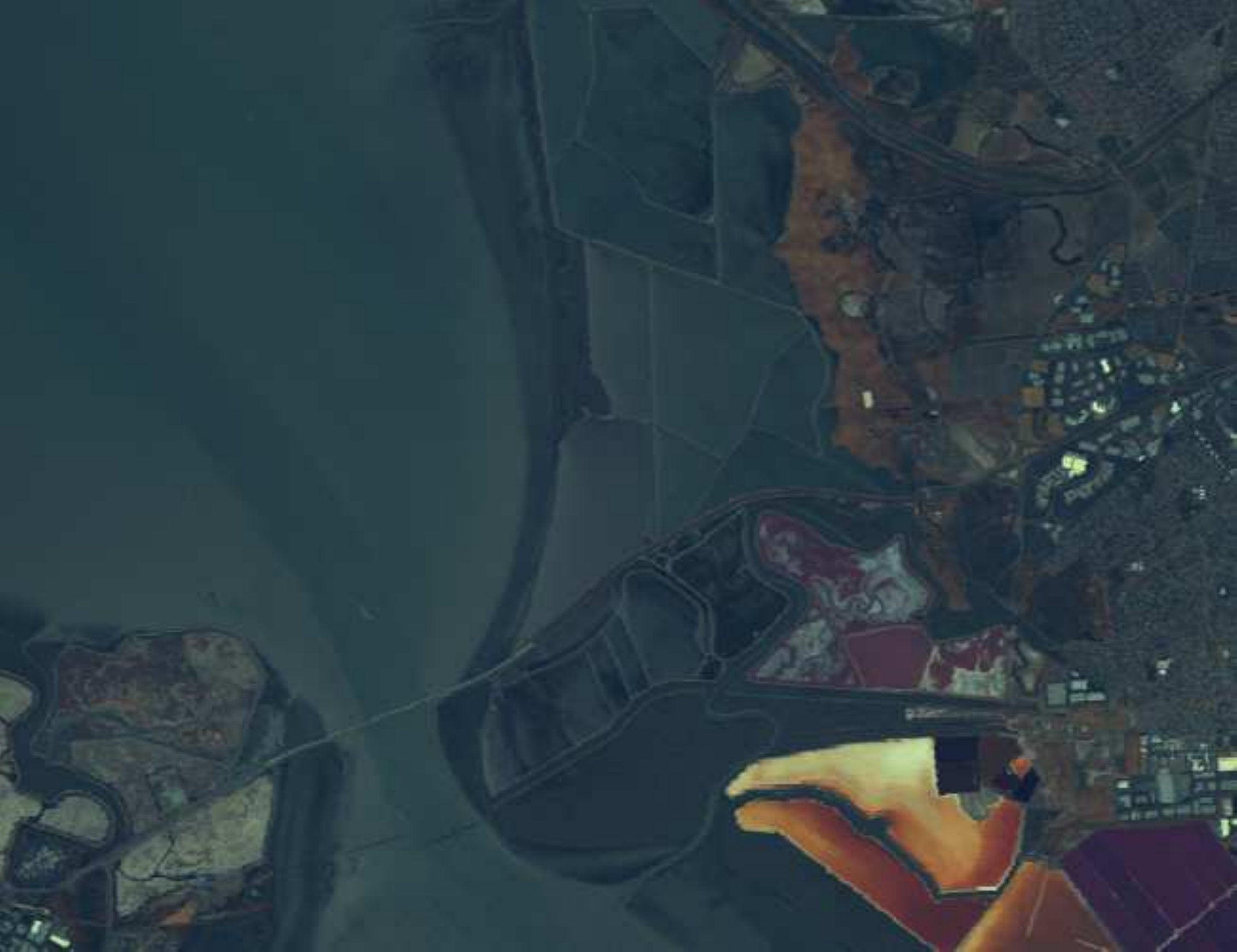}
\label{fig:RGBmoffett}
}
\subfigure[]
{\includegraphics[width=0.45\linewidth]{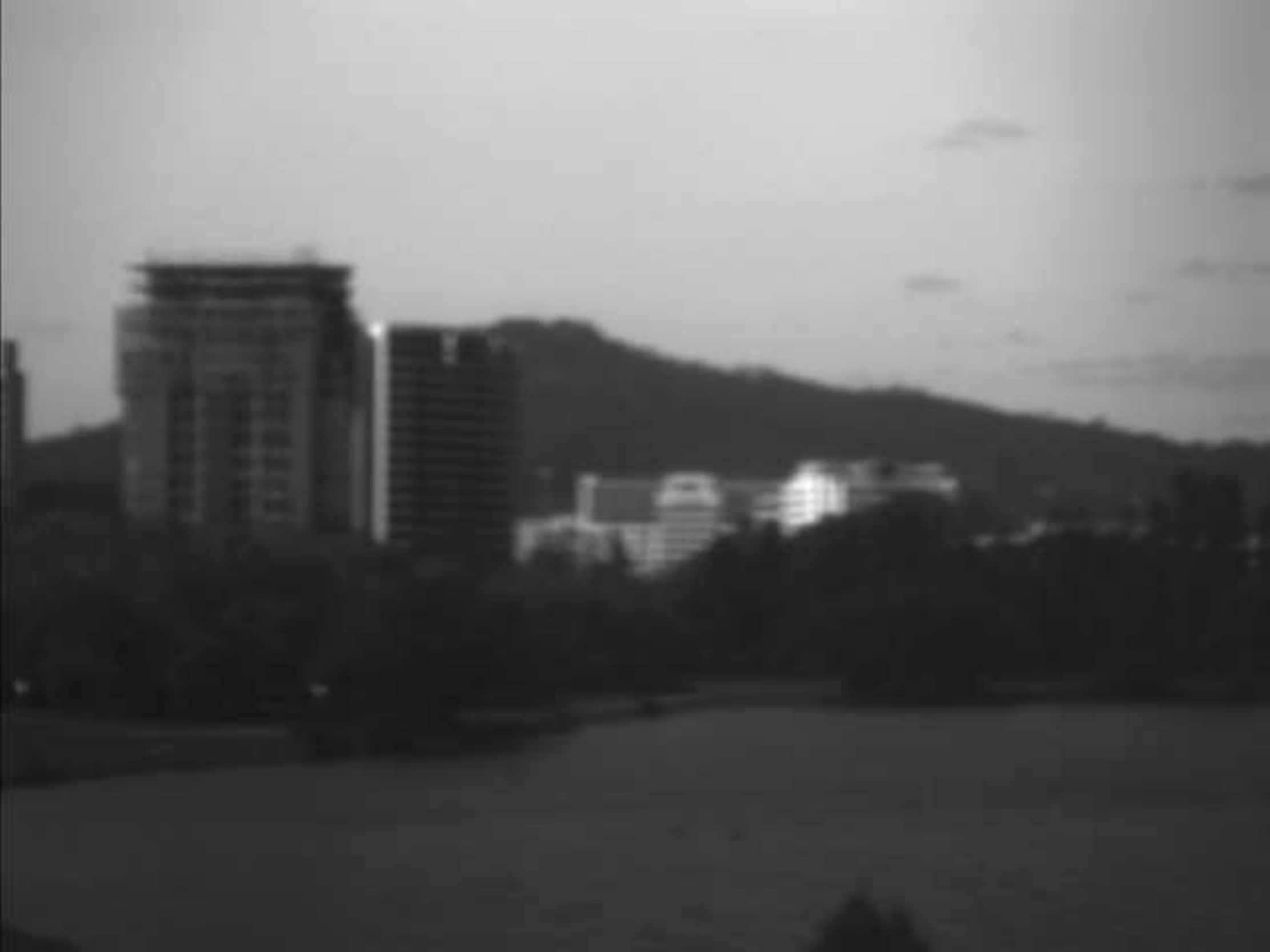}
\label{fig:G03band1}
}
\subfigure[]{
\includegraphics[width=0.45\linewidth]{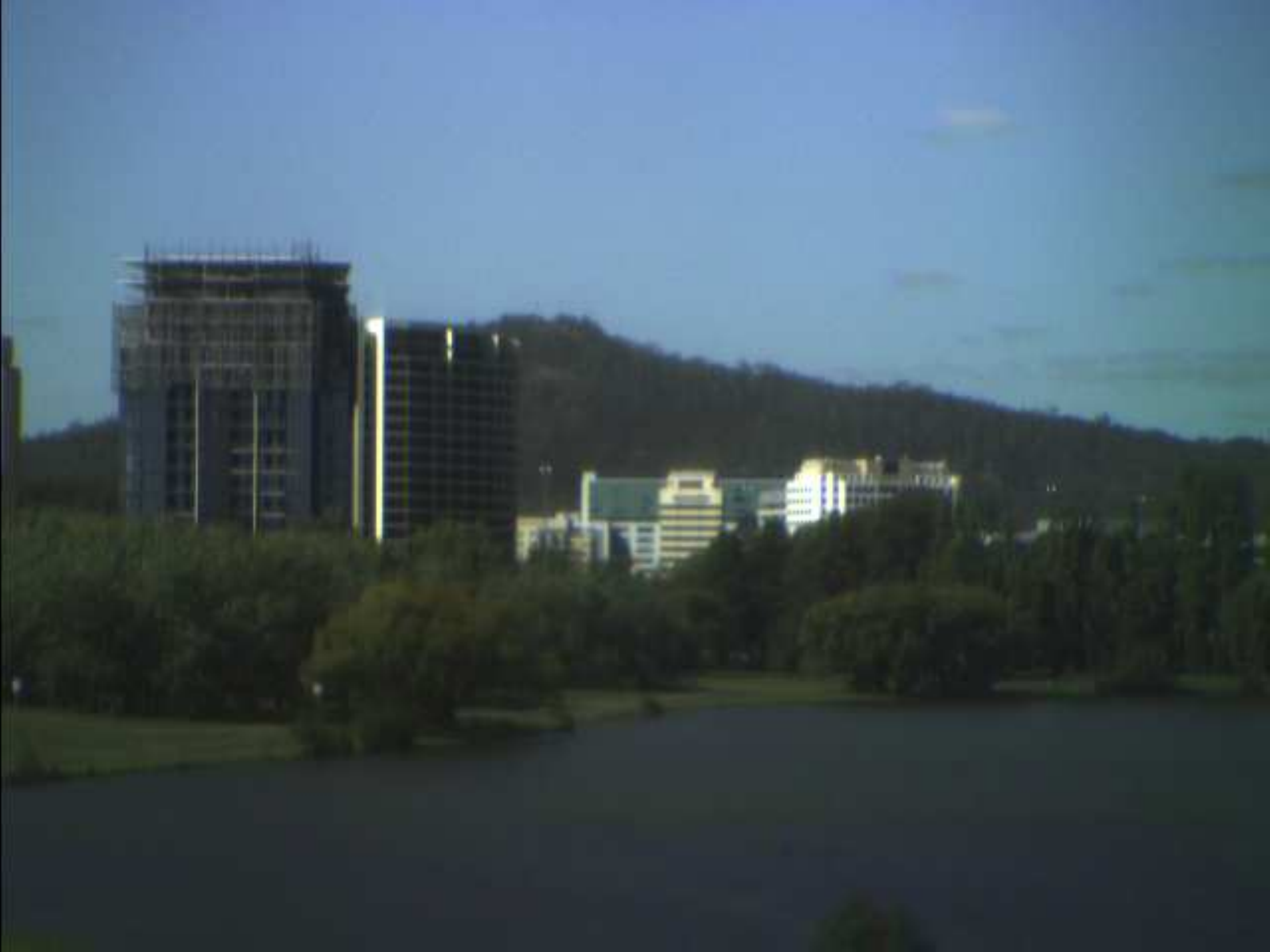}
\label{fig:RGBG03}
}
\subfigure[]
{\includegraphics[width=0.45\linewidth]{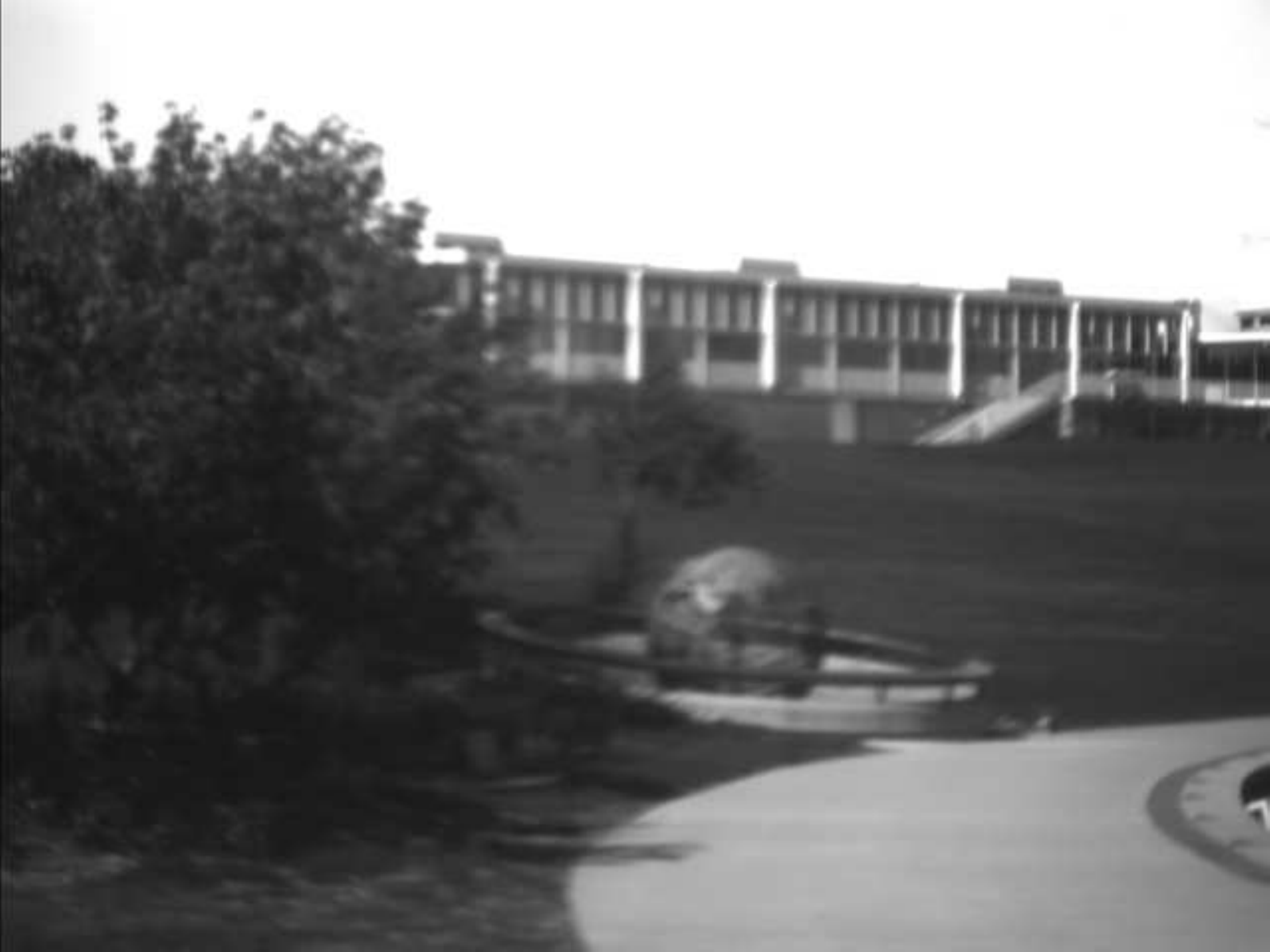}
\label{fig:D04band1}
}
\subfigure[]{
\includegraphics[width=0.45\linewidth]{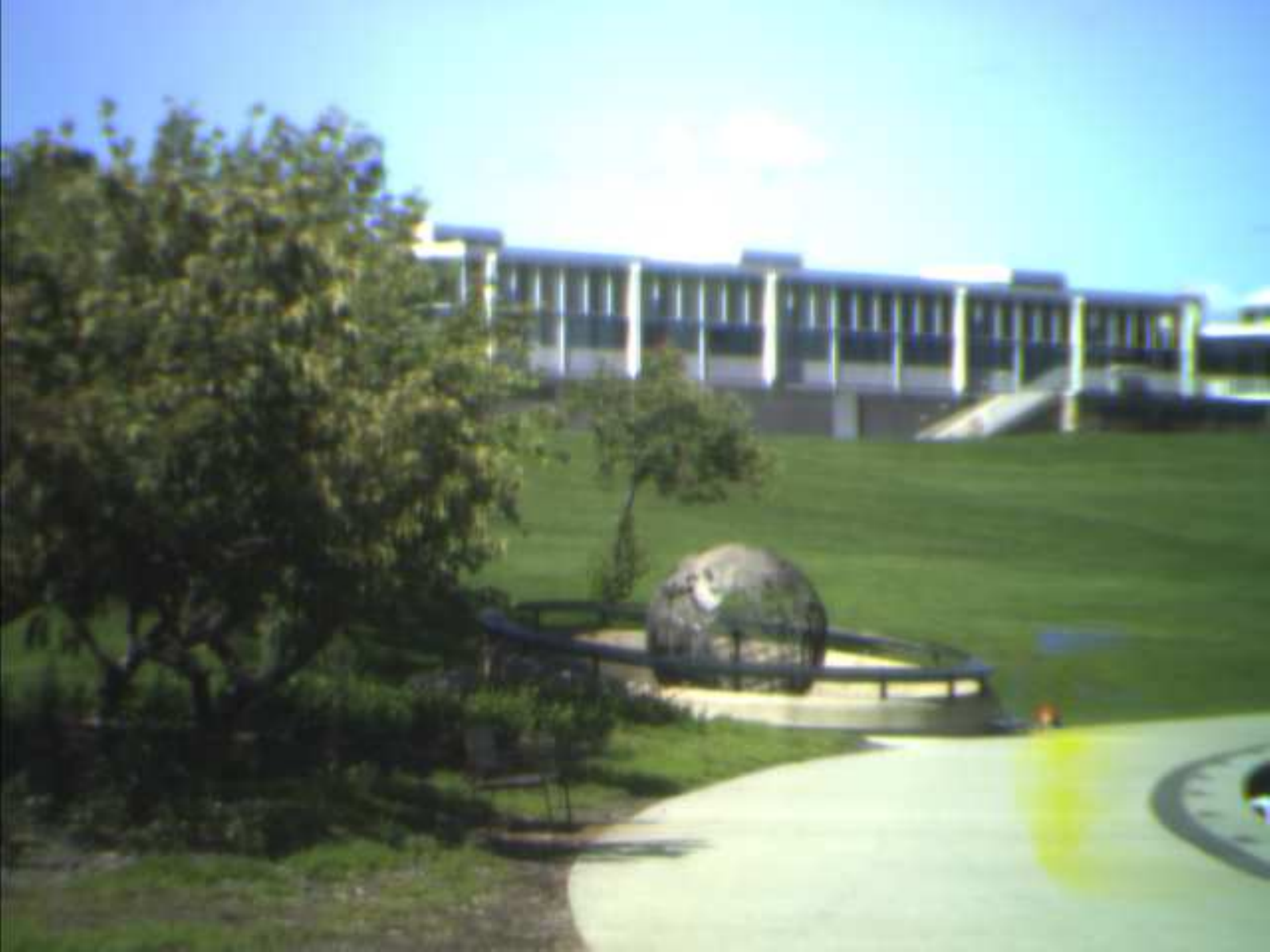}
\label{fig:RGBD04}
}
\caption{HSIs and their corresponding RGB images. (a) The 50th band of the Washiongton D.C. mall (b) The RGB image of the Washiongton D.C. mall.
(c) The 50th band of University of Pavia. (d) The RGB image of Uiversity of Pavia.
(e) The 50th band of the Moffett Field. (f) The RGB image of the Moffett Field.
(g) The first band of G03.
(h) The RGB image of G03. (i) The first band of D04. (j) The RGB image of D04. }
\label{HSIandRGB}
\end{figure}

\begin{figure}[!tp]
\centering
\subfigure[]{
\includegraphics[angle=90,width=0.45\textwidth]{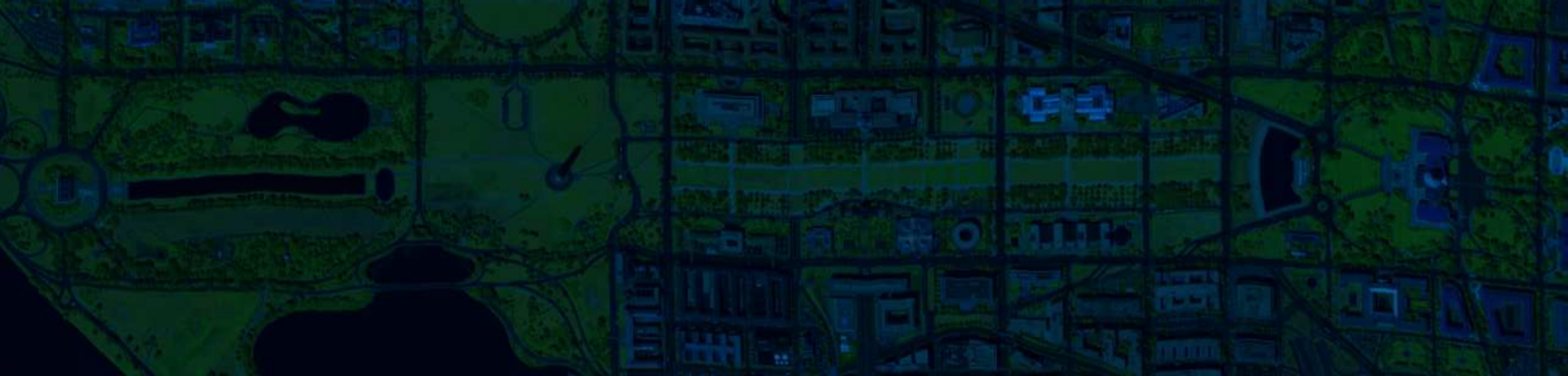}
\label{fig:DCCMF}
}
\subfigure[]{
\includegraphics[angle=90,width=0.45\textwidth]{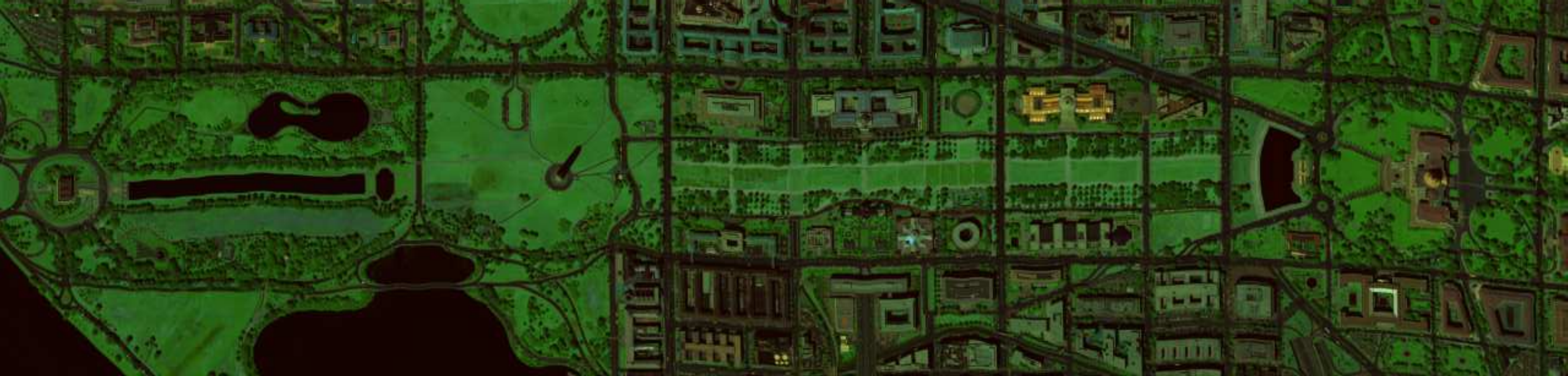}
\label{fig:DCBF}
}
\subfigure[]{
\includegraphics[angle=90,width=0.45\textwidth]{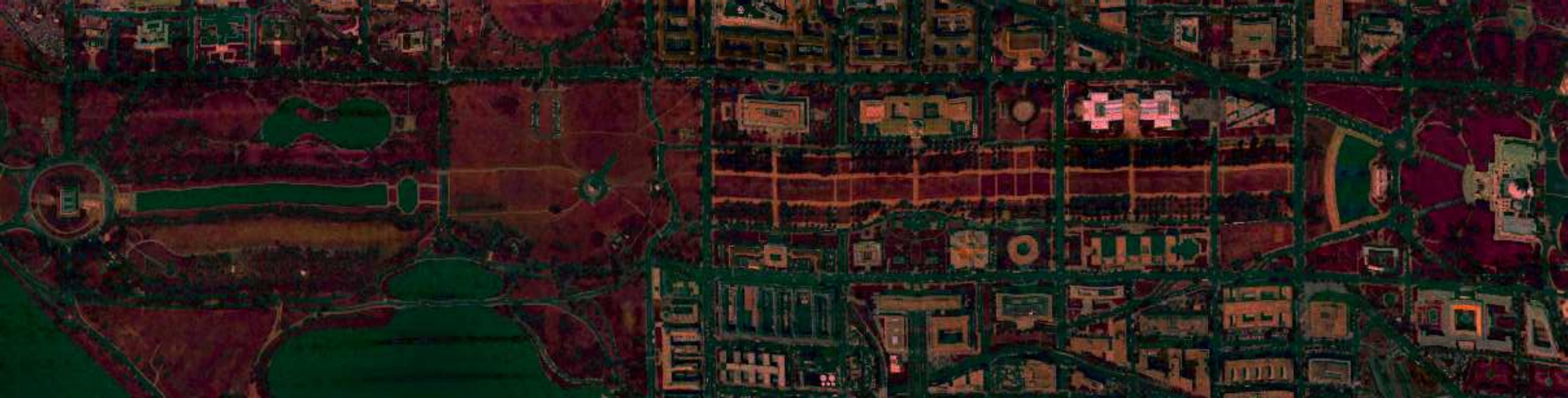}
\label{fig:DCBC}
}
\subfigure[]{
\includegraphics[angle=90,width=0.45\textwidth]{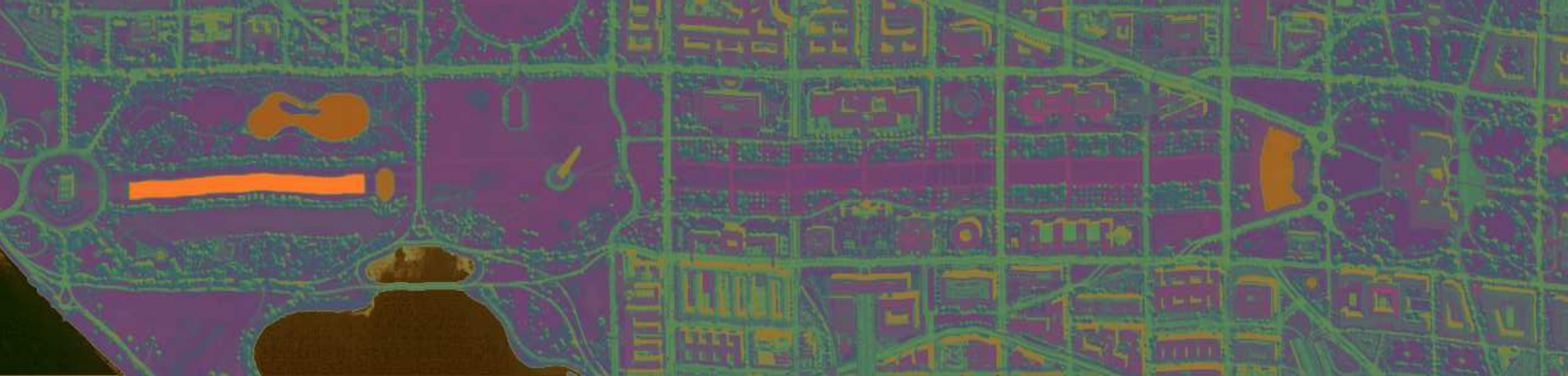}
\label{fig:LE}
}
\subfigure[]{
\includegraphics[angle=90,width=0.45\textwidth]{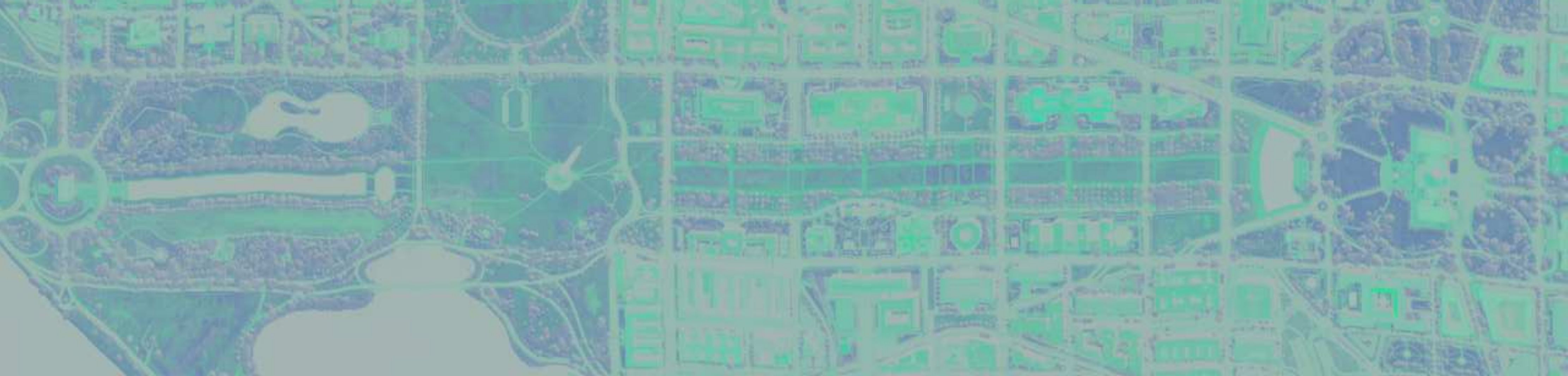}
\label{fig:LPP}
}
\subfigure[]{
\includegraphics[angle=90,width=0.45\textwidth]{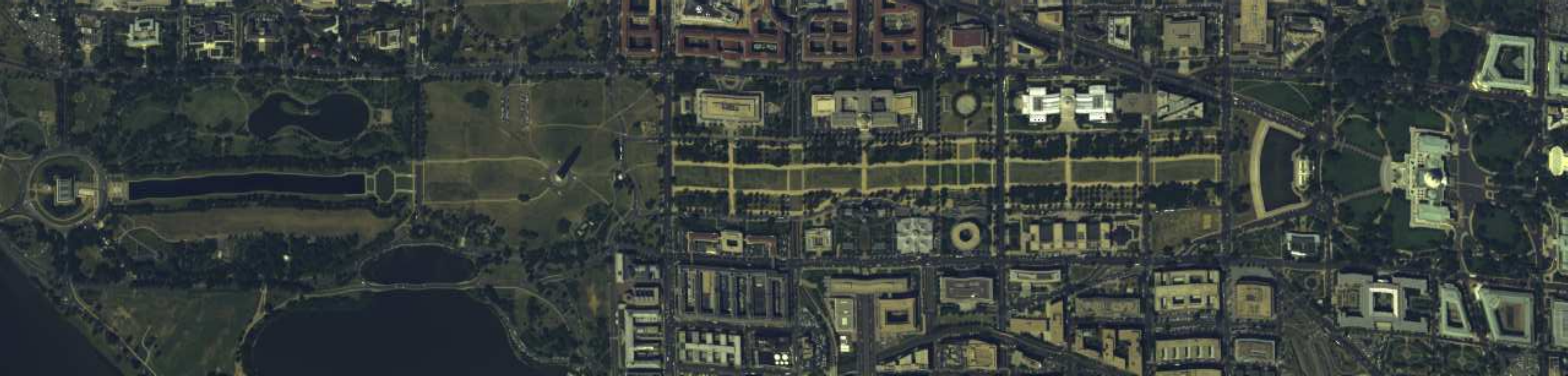}
\label{fig:DCMA}
}
\subfigure[]{
\includegraphics[angle=90,width=0.45\textwidth]{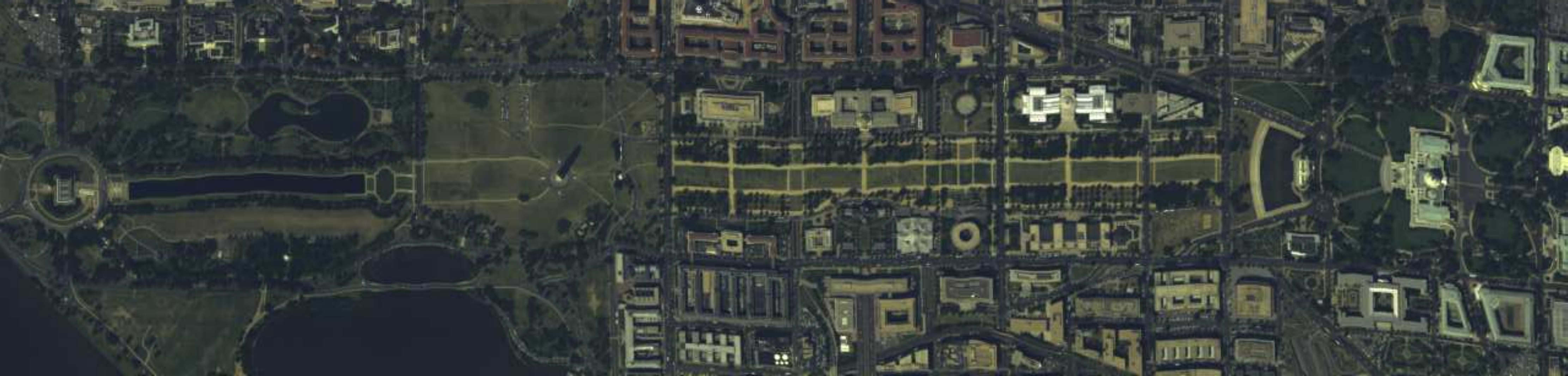}
\label{fig:DCLinear}
}
\subfigure[]{
\includegraphics[angle=90,width=0.45\textwidth]{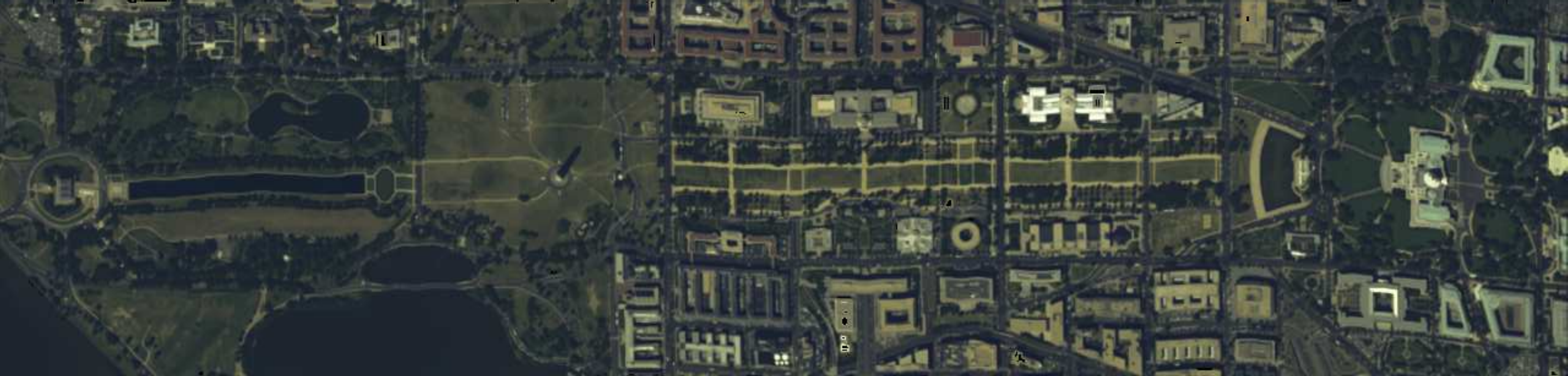}
\label{fig:DCNonlinear}
}
\caption{Visual comparison of different visualization approaches on the Washington DC Mall data set. (a) Stretched CMF. (b) Bilateral filtering. (c) Bicriteria optimization. (d) Laplacian Eigenmaps. (e) LPP. (f) Manifold alignment. (g) The proposed feature-level learning. (h) The proposed instance-level learning.}
\label{DCcomparison}
\end{figure}
\subsection{Feature-level Constrained Manifold Learning}
\label{linearsection}

The proposed feature-level manifold learning learns a linear transformation to find the embedding of the data in a lower dimensional space.
It is based on LPP and further constrains the embedding to be aligned with a referencing data set.
For HSI visualization, it seeks to learn a linear transformation matrix $\textbf{F}$ to map the HSI to the RGB space to generate a natural-looking visualization.
A reasonable criterion for choosing such a transformation is to minimize the following objective function:

\begin{equation}
\label{lossfunction2}
\begin{split}
 G(\textbf{F})&=\frac{1}{2}\sum_{ij} \|\textbf{F}^T\textbf{X}_i-\textbf{F}^T\textbf{X}_j\|^2 \times \textbf{W}_{ij}\\
 &+\lambda \sum_{ik}\|\textbf{F}^T\textbf{X}_i-\textbf{S}_k\|^2 \times \textbf{C}_{ik}
 \end{split}
\end{equation}
where $\textbf{F}^T\textbf{X}_i$ is the representation of $\textbf{X}_i$ in the visualized space.

The first term on the right hand side is the same as LPP, which penalizes the inconsistency between $\textbf{F}^T\textbf{X}_i$ and $\textbf{F}^T\textbf{X}_j$ when $\textbf{X}_i$ and $\textbf{X}_j$ are similar. The second term enforces the colors of pixels in the visualized image to be similar to their matching pixels in the corresponding RGB image .

This objective function is also convex and differentiable. It can be rewritten in matrix form as
\begin{equation}
\label{lossfunctionnew}
\begin{split}
 G(\textbf{F})&=tr(\textbf{F}^T\textbf{XLX}^T\textbf{F})\\
 &+\lambda~tr(\textbf{F}^T\textbf{XC}_1\textbf{X}^T\textbf{F}+
 \textbf{SC}_2\textbf{S}^T-2\textbf{F}^T\textbf{XCS}^T).
 \end{split}
\end{equation}
The derivative of the function with respect to $\textbf{F}$ is
\begin{equation}
\frac{\partial{G}}{\partial{\textbf{F}}}=2\textbf{XLX}^T\textbf{F}+2\lambda \textbf{XC}_1\textbf{X}^T\textbf{F}-2\lambda \textbf{XCS}^T.
\end{equation}
The optimal $\textbf{F}$ minimizing the objective function can be obtained by
\begin{equation}
\label{linear_finalEq}
\textbf{F}=(\textbf{X}(\frac{1}{\lambda}\textbf{L}+\textbf{C}_1)\textbf{X}^T)^{-1}\textbf{XCS}^T.
\end{equation}

The main difference between instance-level and feature-level methods is that the former is nonlinear while the latter is linear. More specifically, instance-level learning directly computes the coordinates of samples in the low dimensional space. It does not assume any explicit transformations between the two spaces.
The feature-level learning builds connections between features rather than instances, thus the learning result can be generalized to new test instances.
In the task of HSI visualization, the projection function learned from a pair of HSI and RGB image can be directly applied to visualize other similar HSIs acquired by the same imaging sensor.
Assume a mapping function $\textbf{F}$ is learnt from an HSI $\textbf{X}$. For another HSI $\textbf{X}'$ captured by the same sensor as $\textbf{X}$, the visualized image $\textbf{Y}'$ of $\textbf{X}'$ can be obtained by $\textbf{Y}'=\textbf{F}^T\textbf{X}'$.
Since the same types of objects shall have similar spectral responses with the same sensor, the same/similar objects in $\textbf{X}$ and $\textbf{X}'$ will be presented by consistent and natural colors.
This scenario is very helpful when users require a quick overview of a batch of HSIs generated by the same imaging sensor,
since the projection function learning only need to be undertaken once.

Another advantage of feature-level learning is that it is faster than instance-level method since it only requires to estimate the $q\times 3$ parameters in the transformation matrix.
More specifically, to solve Equation~(\ref{linear_finalEq}), feature-level learning only needs to compute the inverse of a $p\times p$ matrix, where $p$ is the number of channels in the HSI.  The instance-level learning, however, needs to compute the inverse of a $n\times n$ matrix in Equation~(\ref{instance_finalEq}), where $n$ is the number of pixels in the HSI. The inversion of a matrix of size $n\times n$ has a time complexity of $O(n^3)$. Since the construction of the graph Laplacian has a time complexity of $O(pn\log n)$ where $p<<n$, instance-level learning has a time complexity of $O(n^3)$ and feature-level learning has a time complexity of $O(pn \log n)$.
\begin{figure*}[!tp]
\centering
\subfigure[]{
\includegraphics[width=0.15\linewidth]{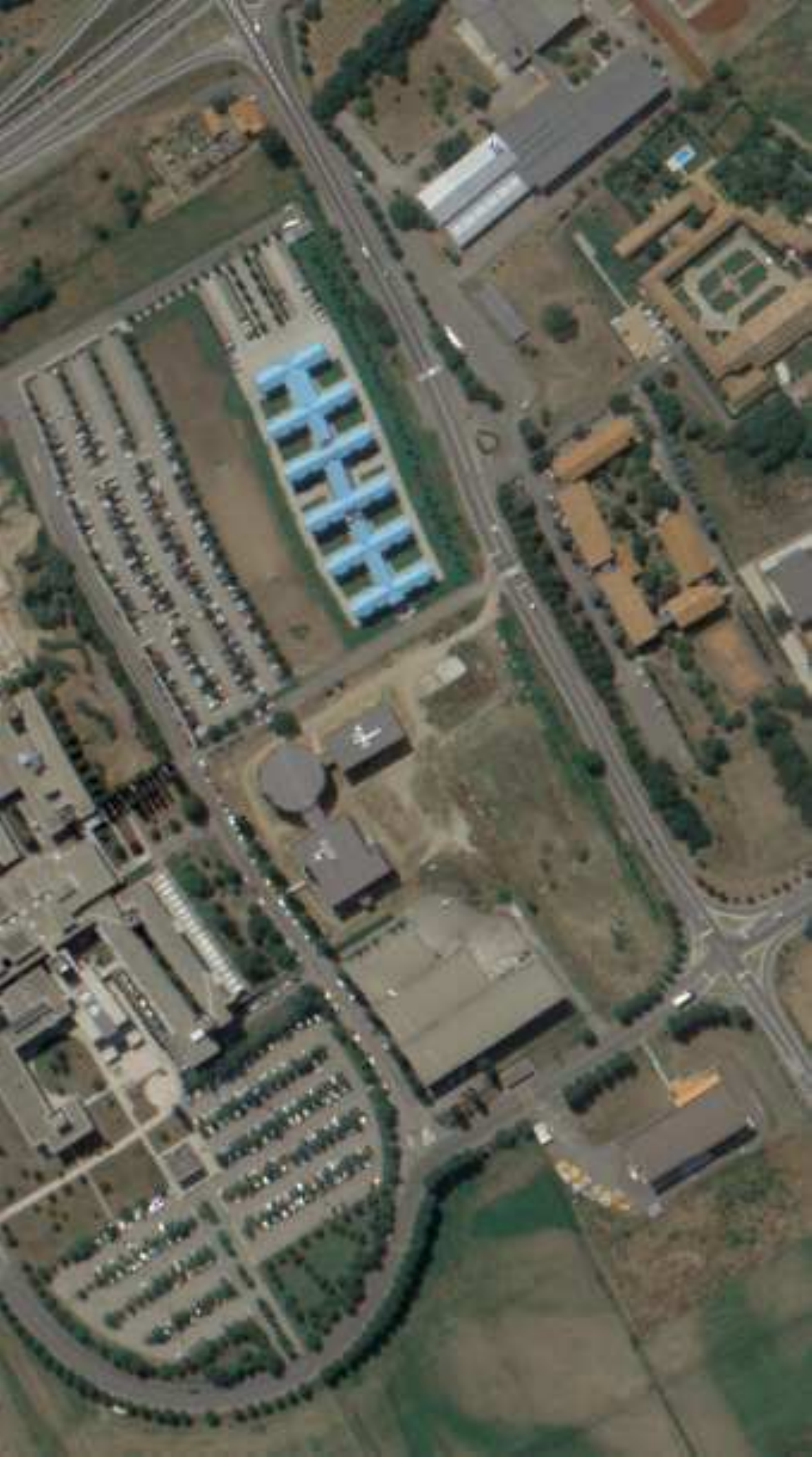}
}
\subfigure[]{
\includegraphics[width=0.15\linewidth]{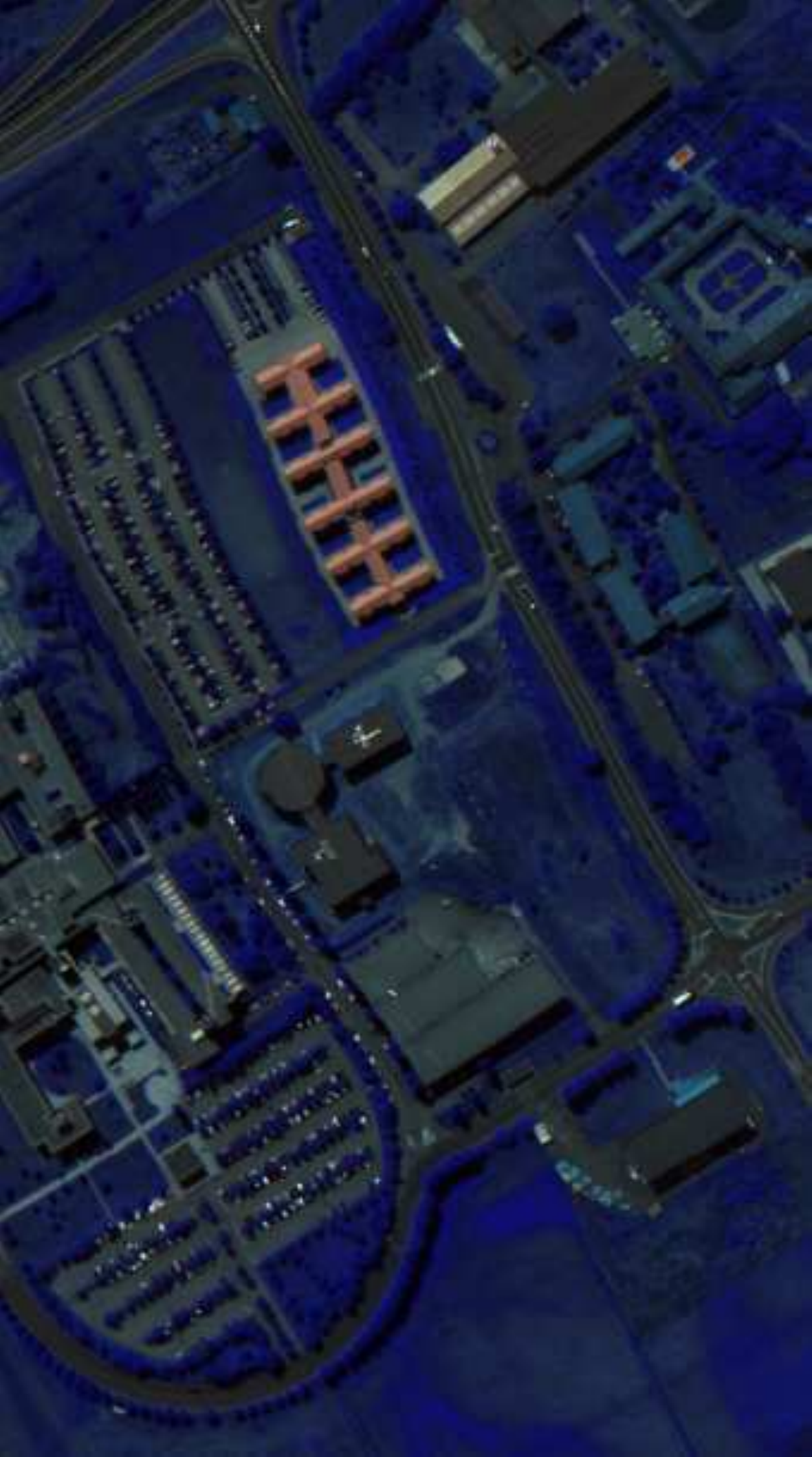}
}
\subfigure[]{
\includegraphics[width=0.15\linewidth]{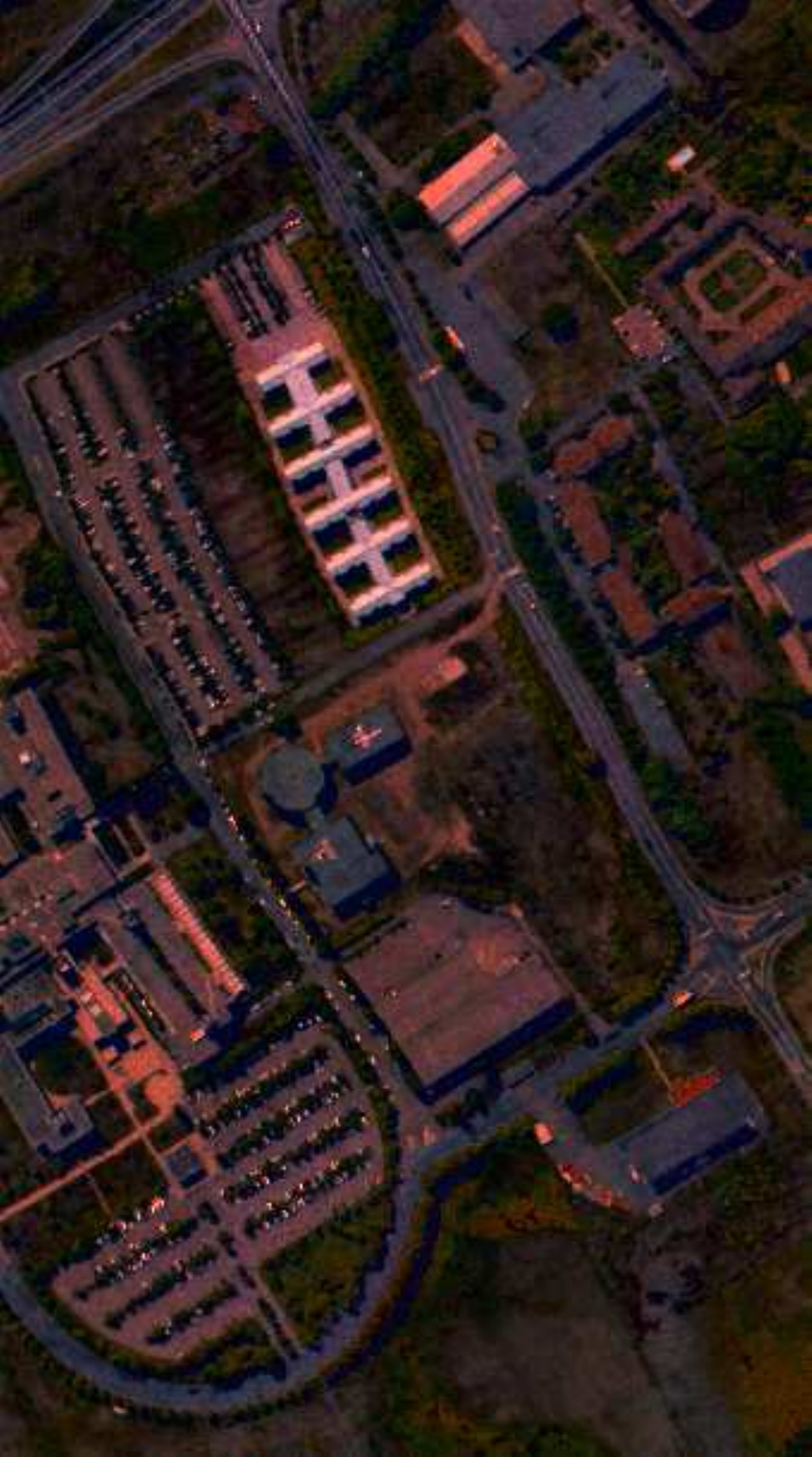}
}
\subfigure[]{
\includegraphics[width=0.15\linewidth]{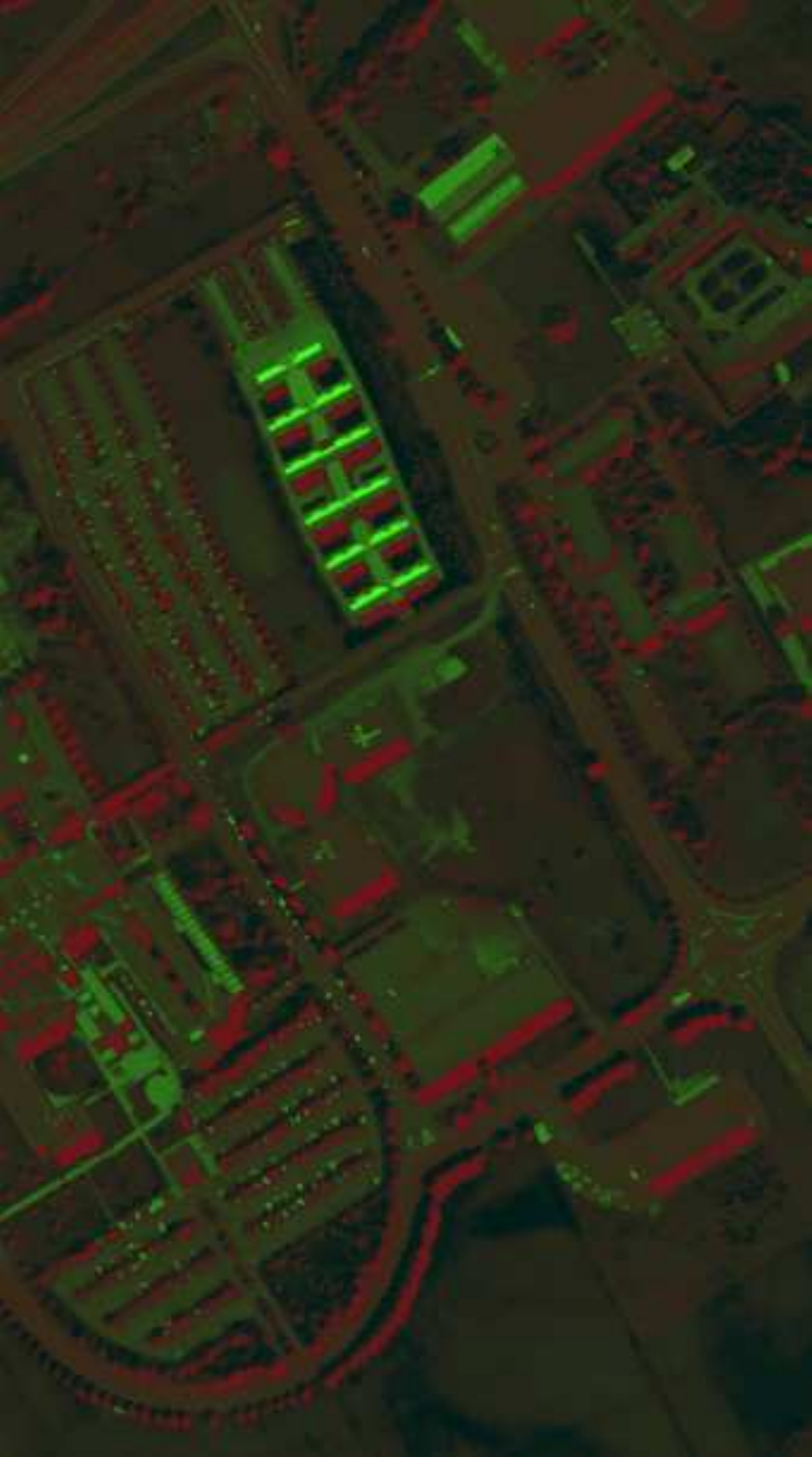}
}\\
\subfigure[]{
\includegraphics[width=0.15\linewidth]{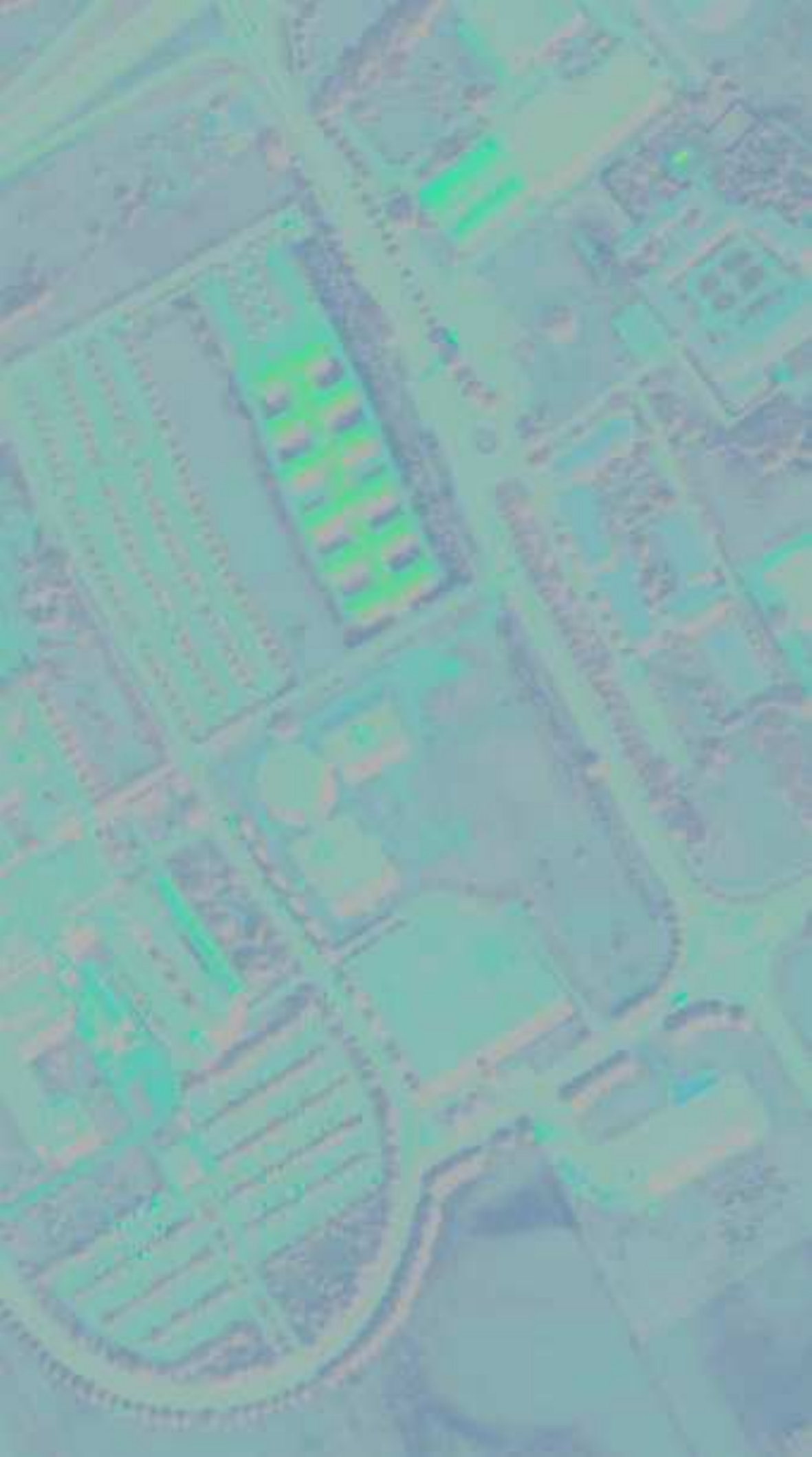}
}
\subfigure[]{
\includegraphics[width=0.15\linewidth]{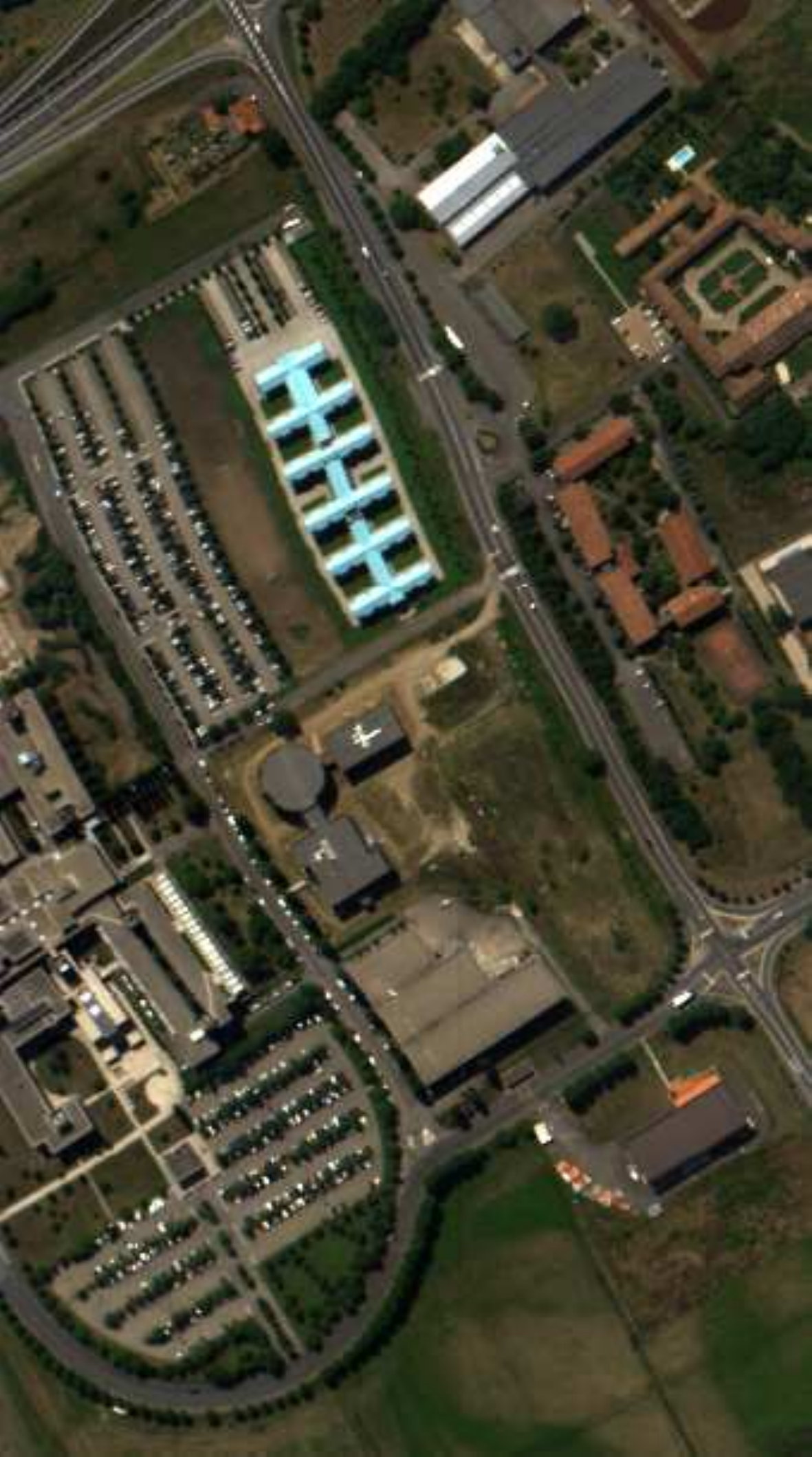}
}
\subfigure[]{
\includegraphics[width=0.15\linewidth]{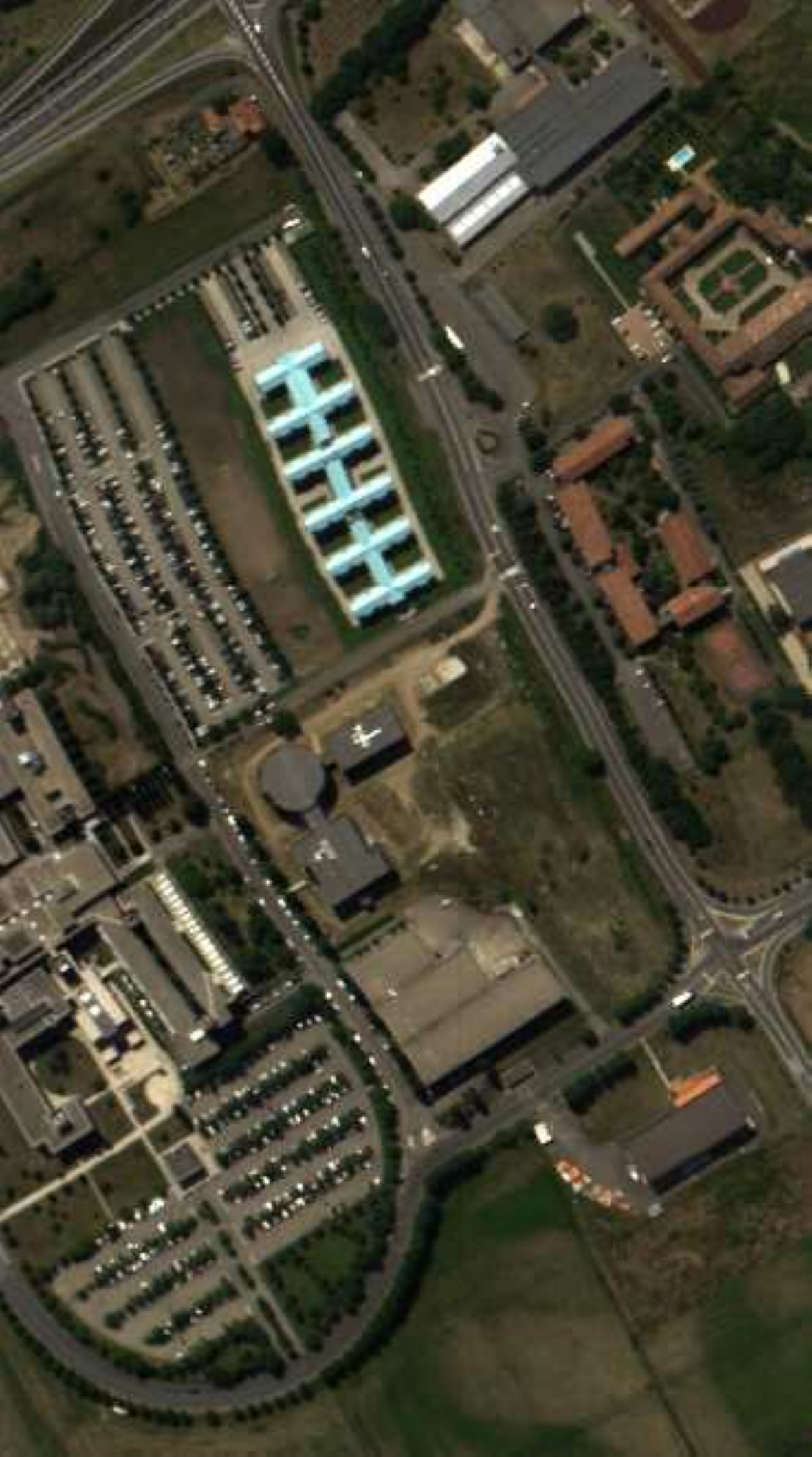}
}
\subfigure[Nonlinear]{
\includegraphics[width=0.15\linewidth]{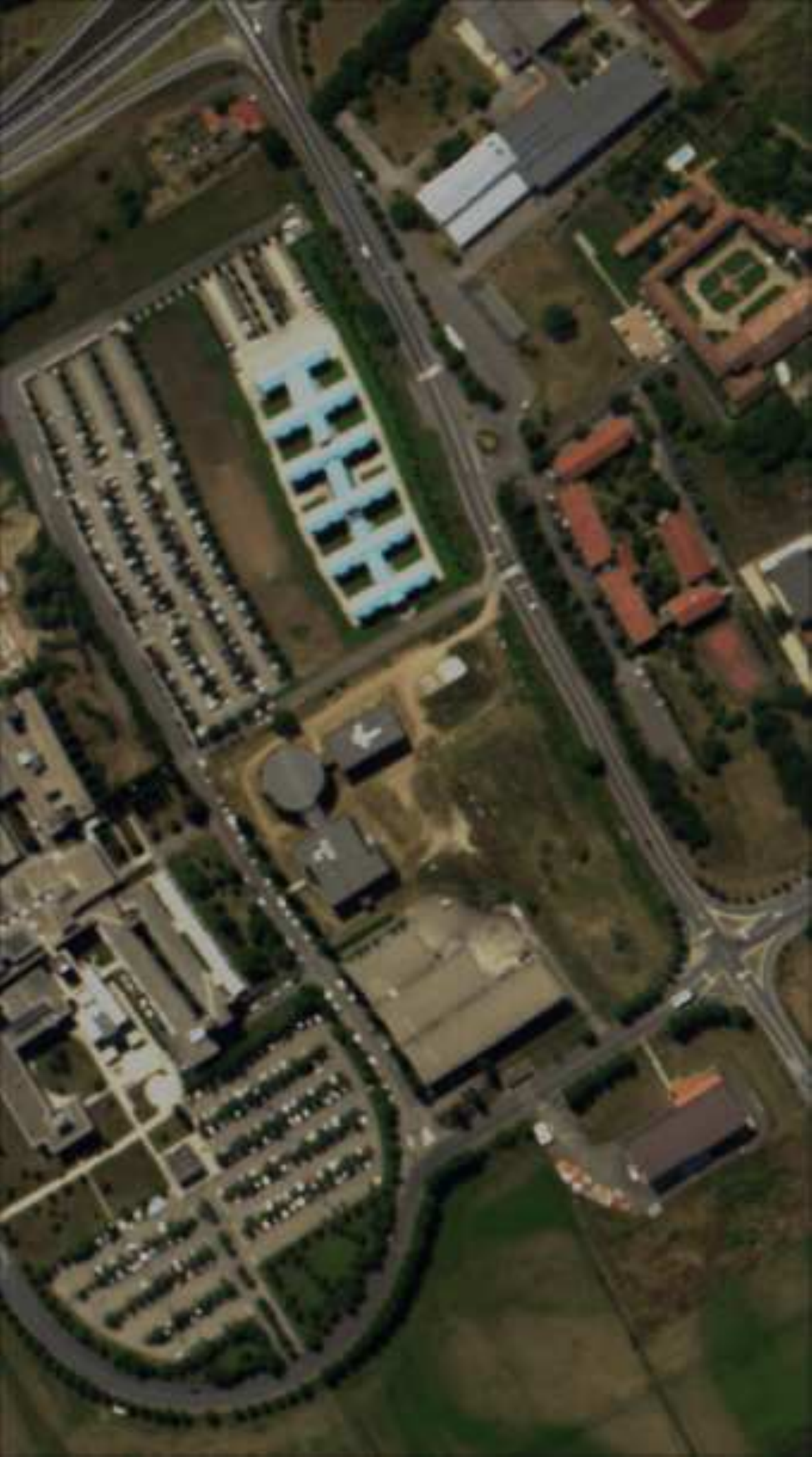}
}
\caption{Visual comparison of different visualization approaches on the University of Pavia data set.
(a) Stretched CMF. (b) Bilateral filtering. (c) Bicriteria optimization. (d) Laplacian Eigenmaps. (e) LPP. (f) Manifold alignment. (g) The proposed feature-level learning. (h) The proposed instance-level learning.}
\label{PaviaExperiments}
\end{figure*}

\begin{figure*}[!tp]
\centering
\subfigure[]{
\includegraphics[width=0.23\linewidth]{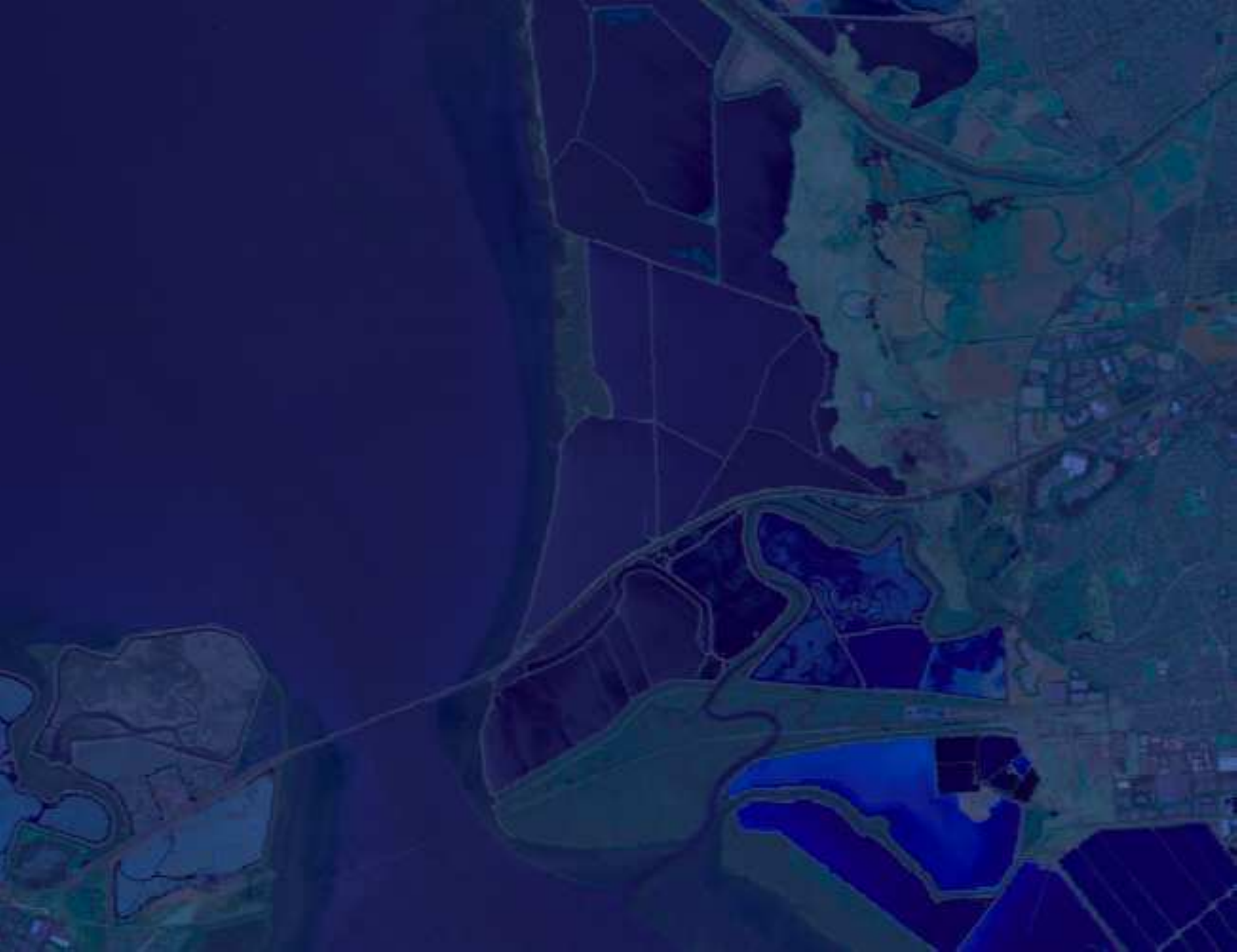}
}
\subfigure[]{
\includegraphics[width=0.23\linewidth]{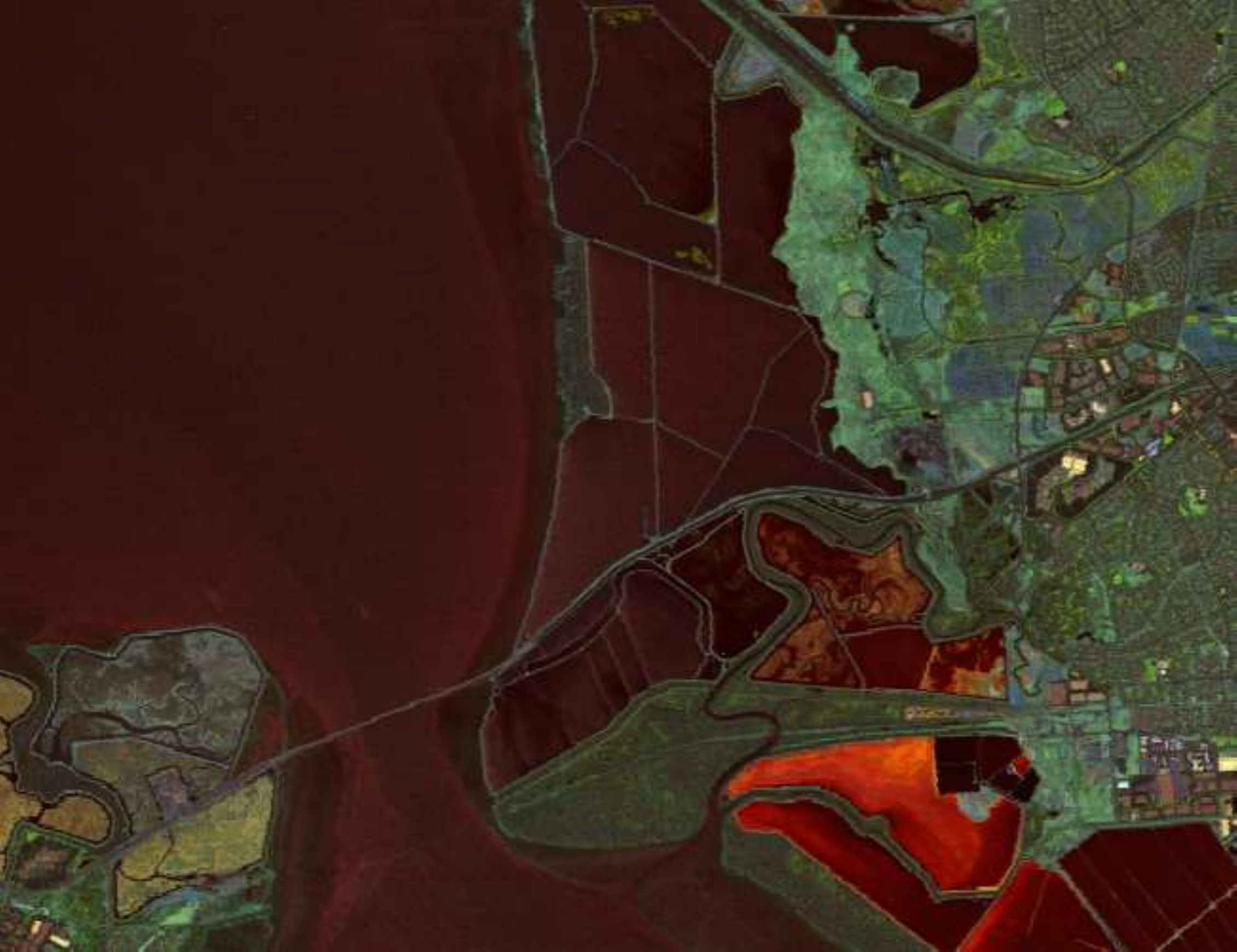}
}
\subfigure[]{
\includegraphics[width=0.23\linewidth]{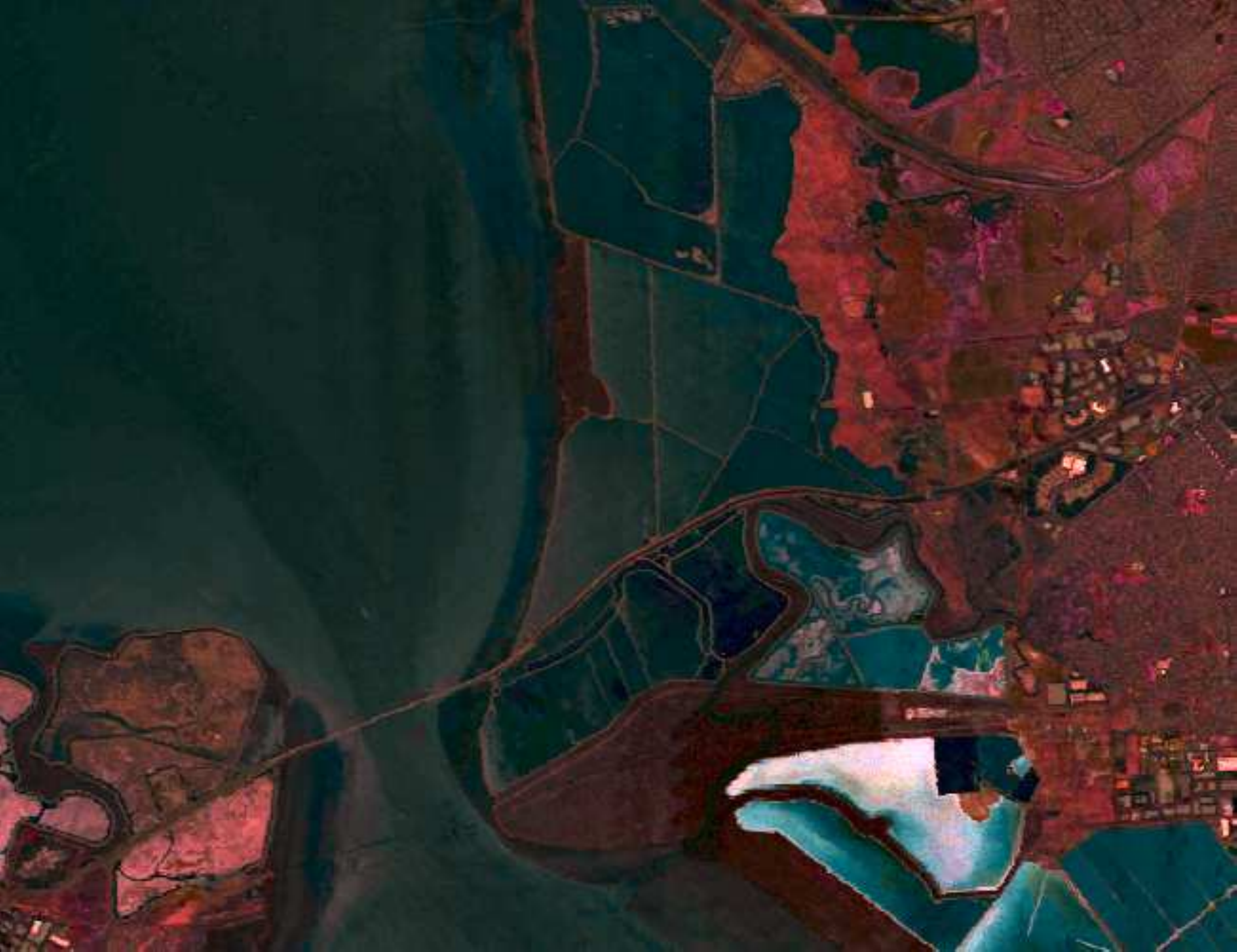}
}
\subfigure[]{
\includegraphics[width=0.23\linewidth]{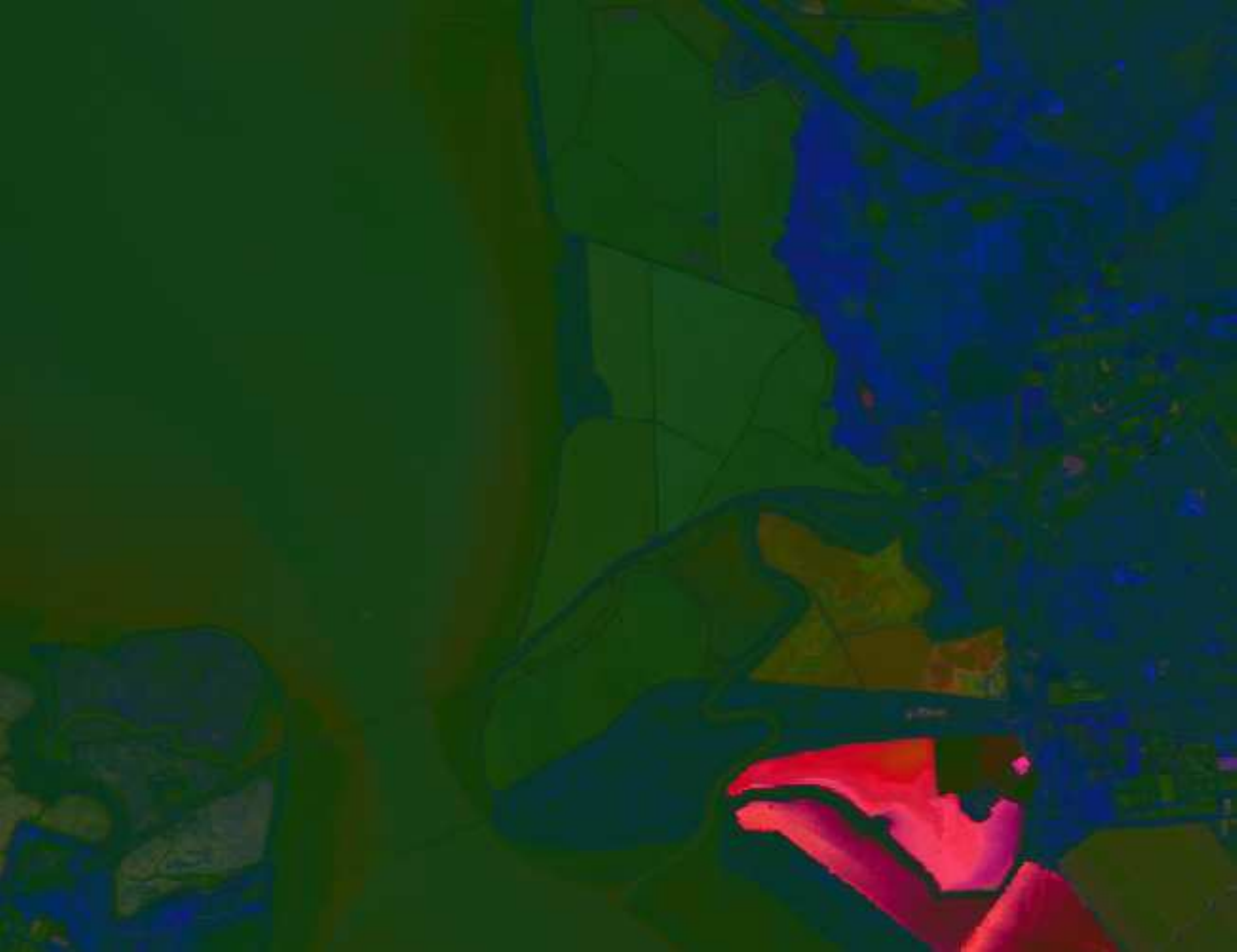}
}
\subfigure[]{
\includegraphics[width=0.23\linewidth]{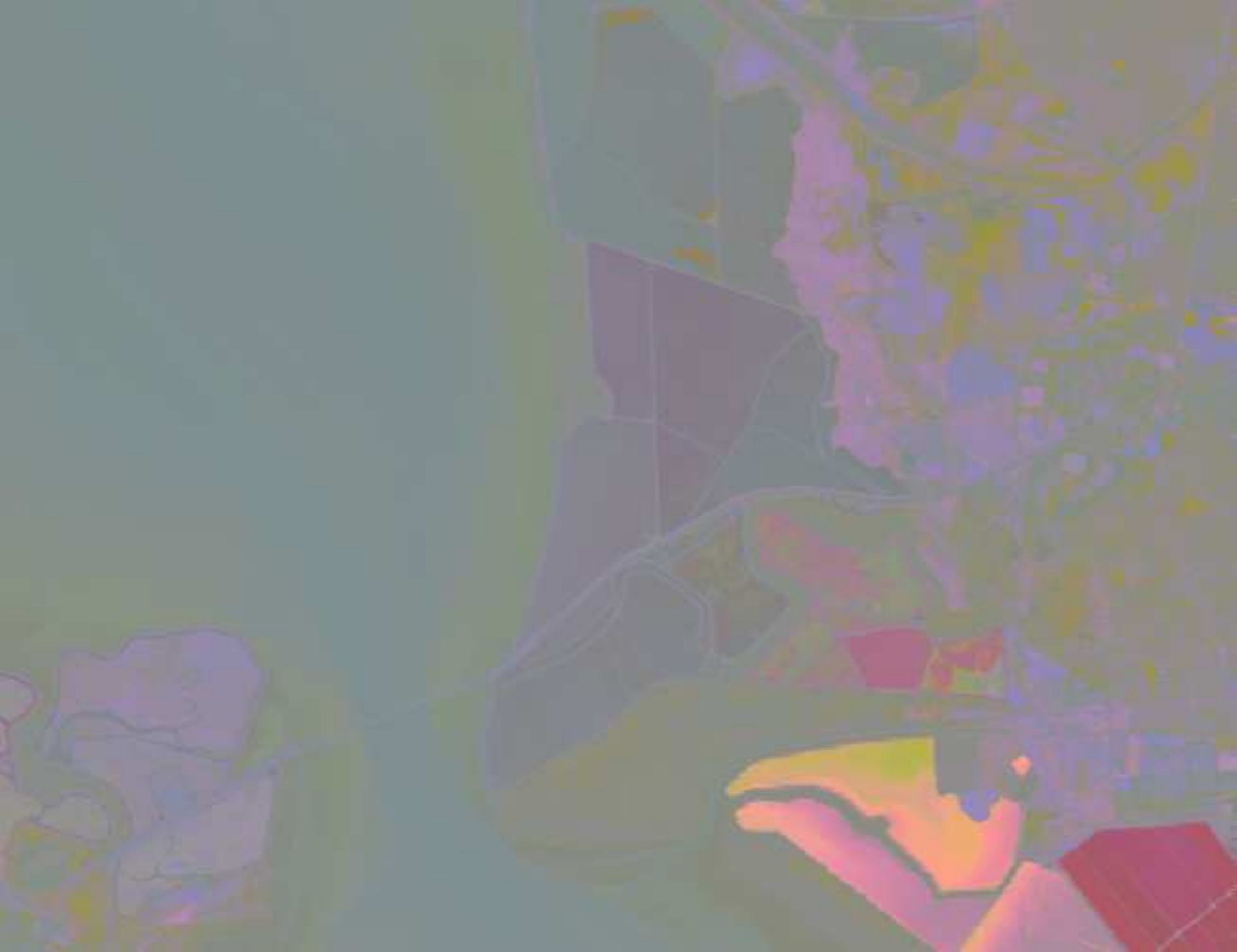}
}
\subfigure[]{
\includegraphics[width=0.23\linewidth]{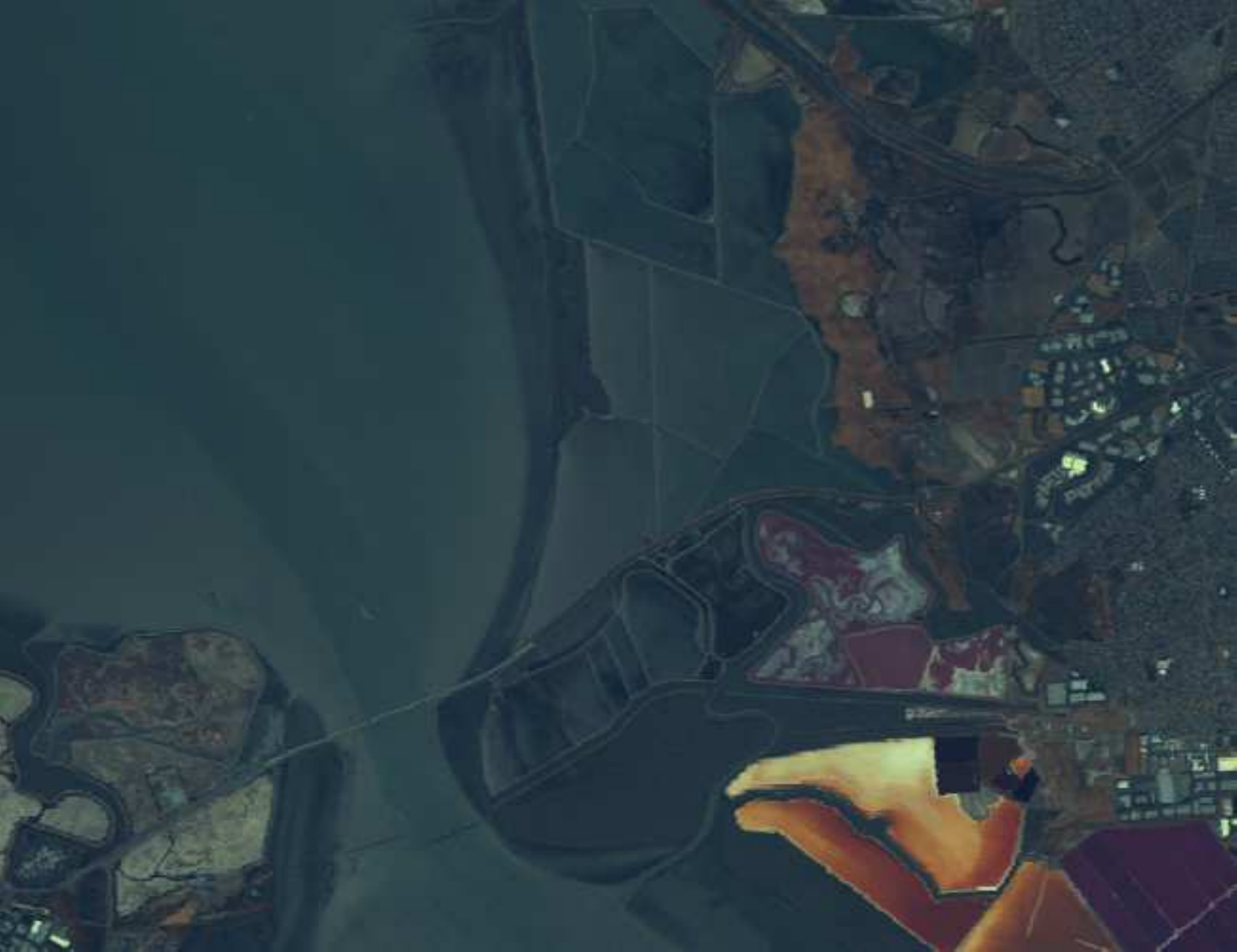}
}
\subfigure[]{
\includegraphics[width=0.23\linewidth]{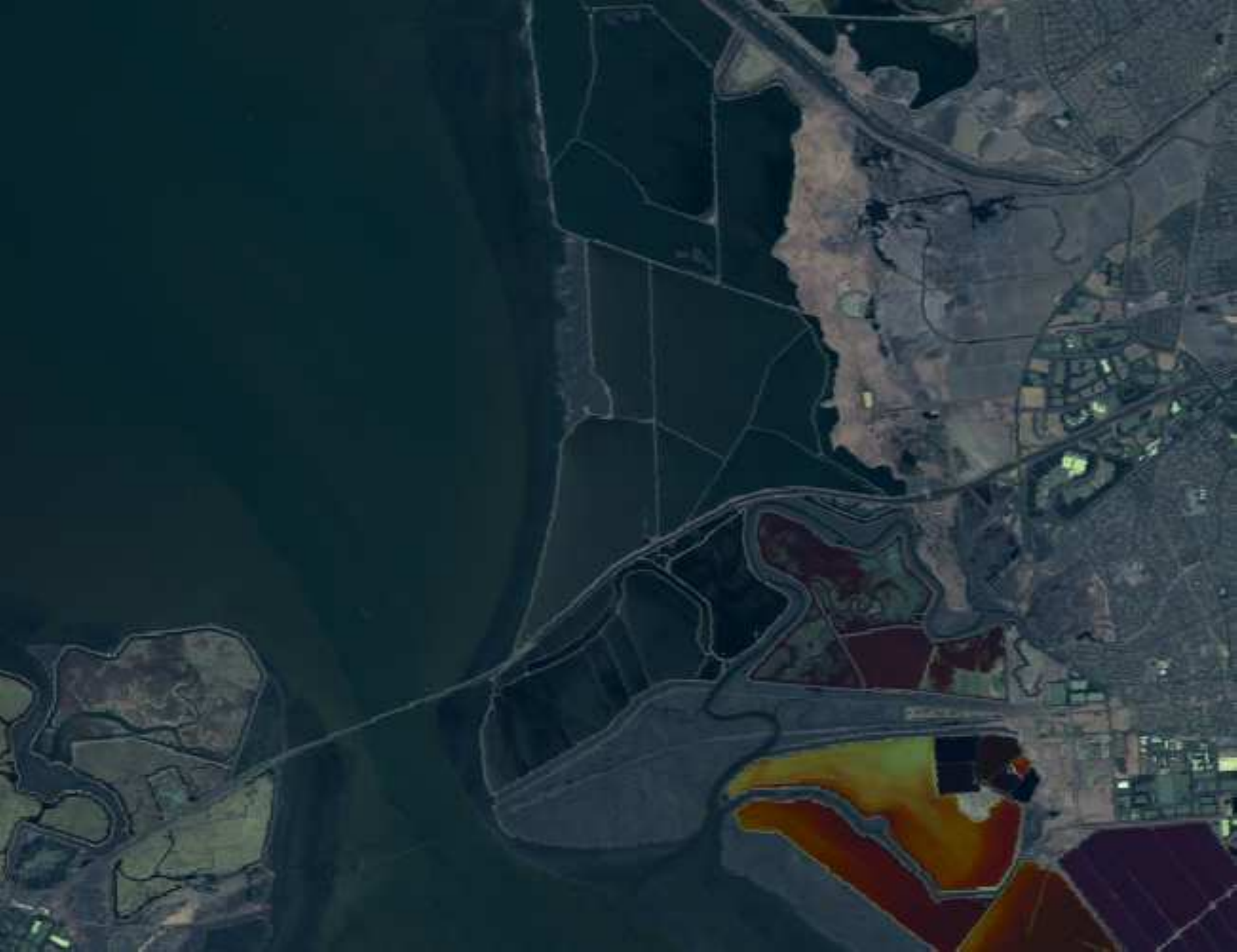}
}
\subfigure[]{
\includegraphics[width=0.23\linewidth]{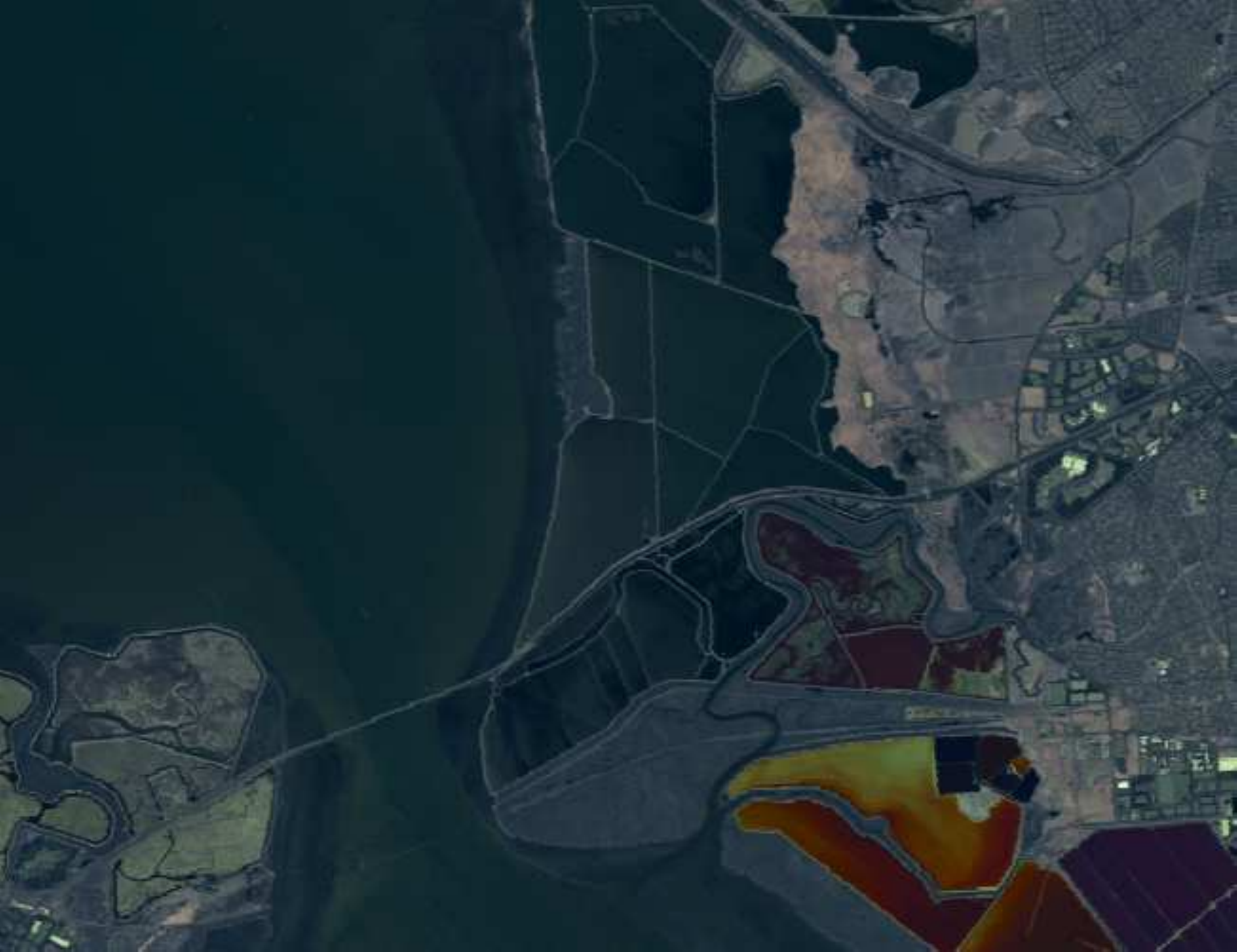}
}
\caption{Visual comparison of different visualization approaches on the Moffett Field data set.
(a) Stretched CMF. (b) Bilateral filtering. (c) Bicriteria optimization. (d) Laplacian Eigenmaps. (e) LPP. (f) Manifold alignment. (g) The proposed feature-level learning. (h) The proposed instance-level learning.}\label{MoffettExperiments}
\end{figure*}

\begin{figure*}[!tp]
\centering
\subfigure[]{
\includegraphics[width=0.23\linewidth]{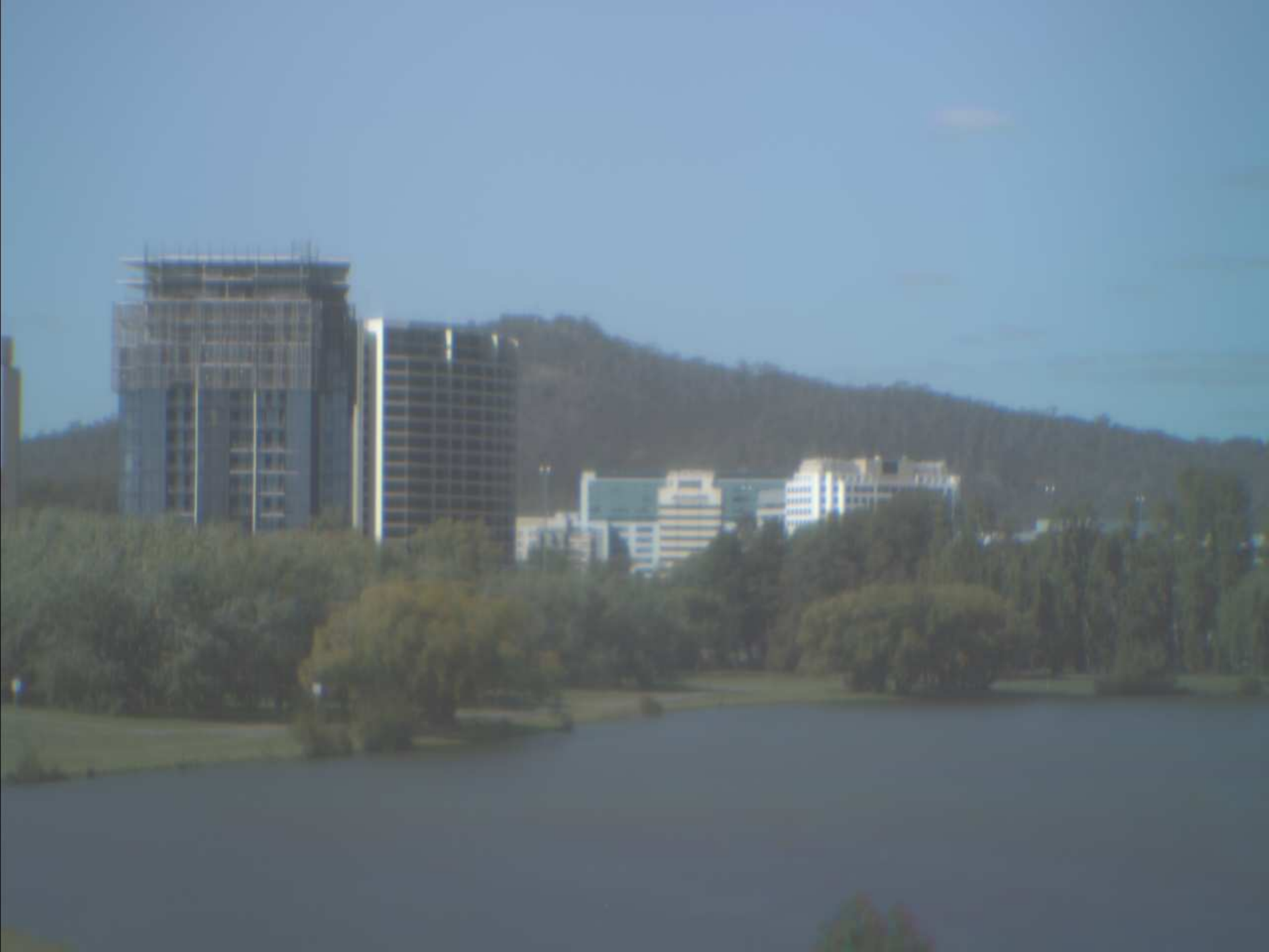}
}
\subfigure[]{
\includegraphics[width=0.23\linewidth]{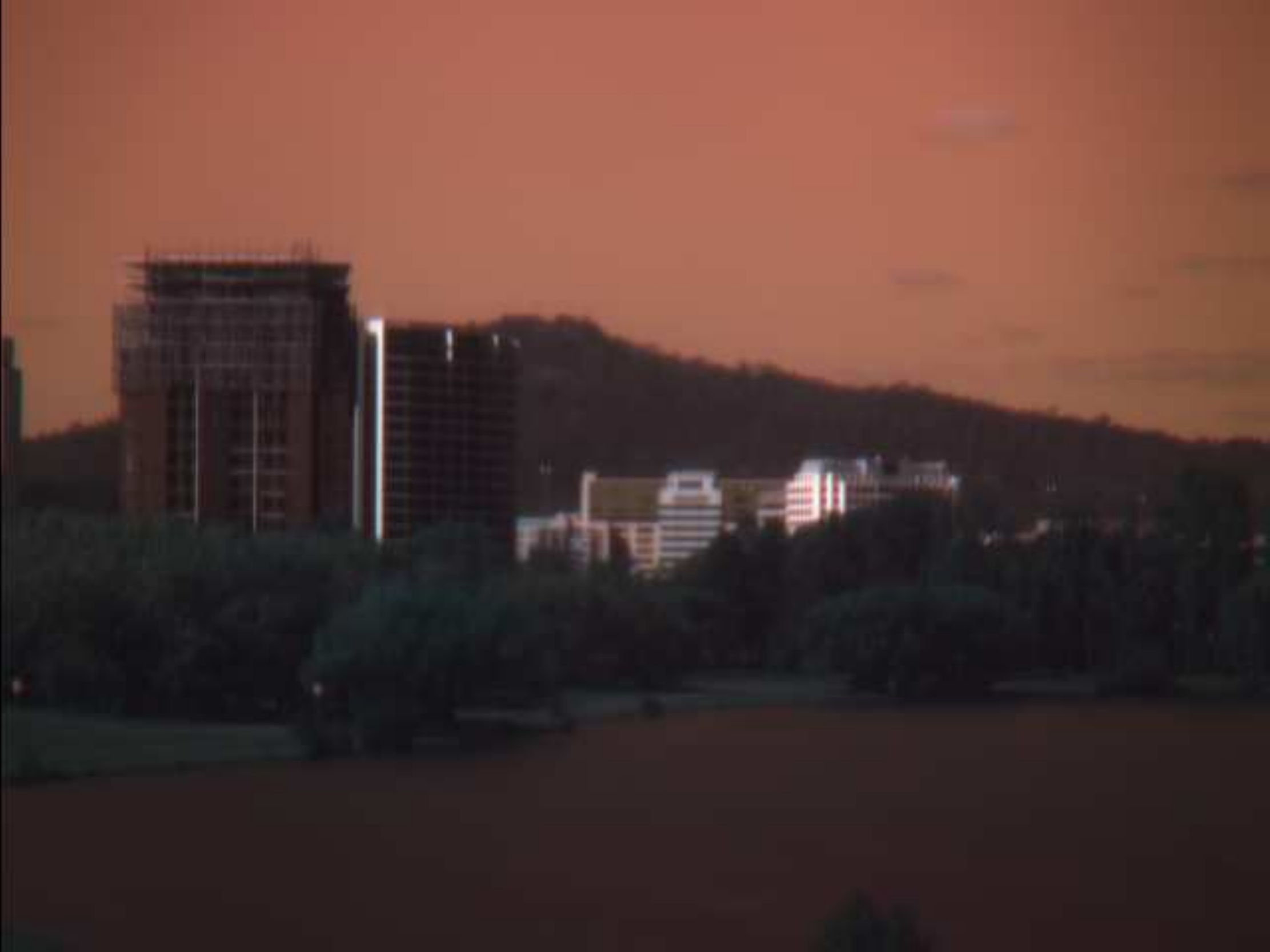}
}
\subfigure[]{
\includegraphics[width=0.23\linewidth]{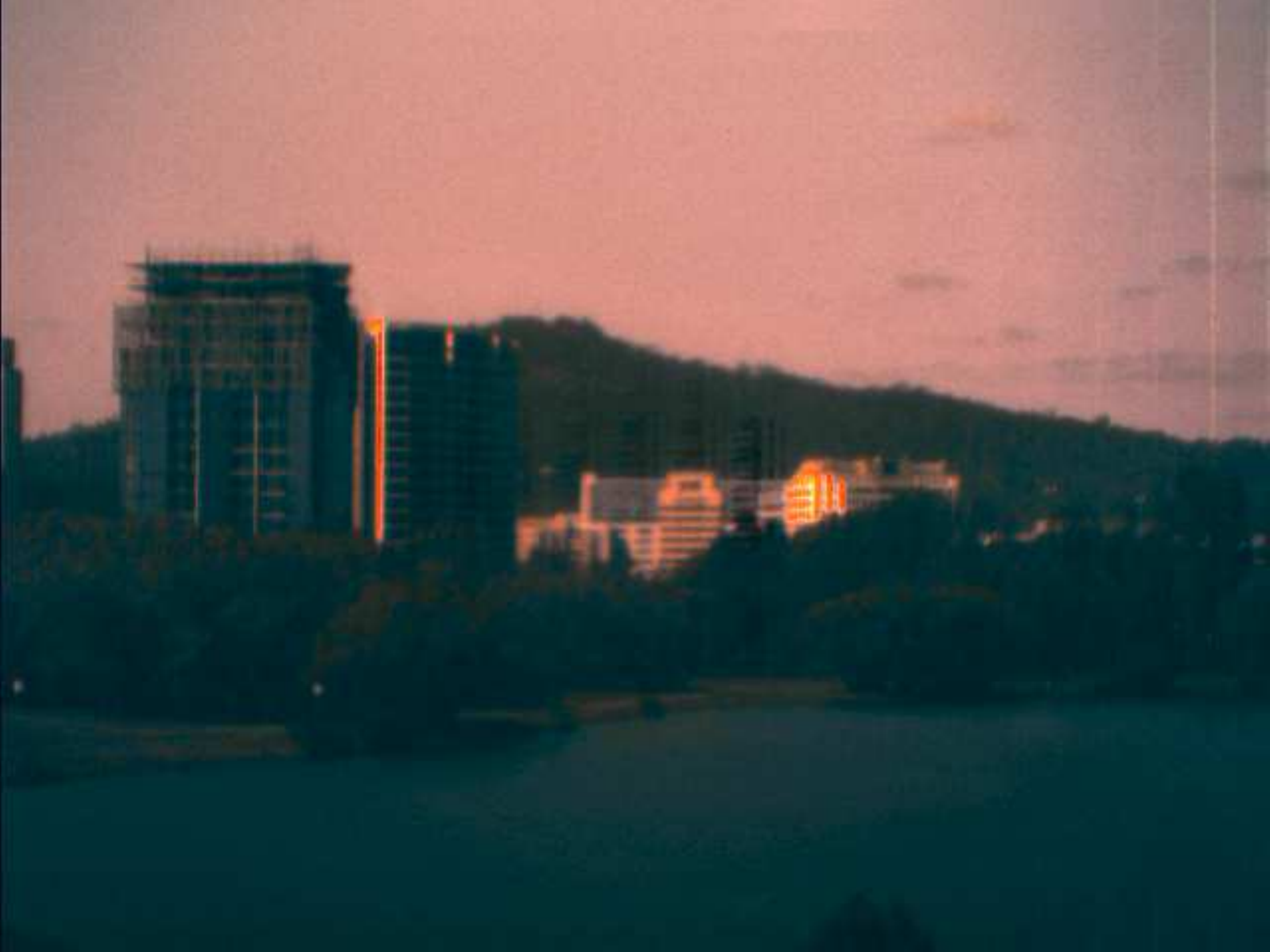}
}
\subfigure[]{
\includegraphics[width=0.23\linewidth]{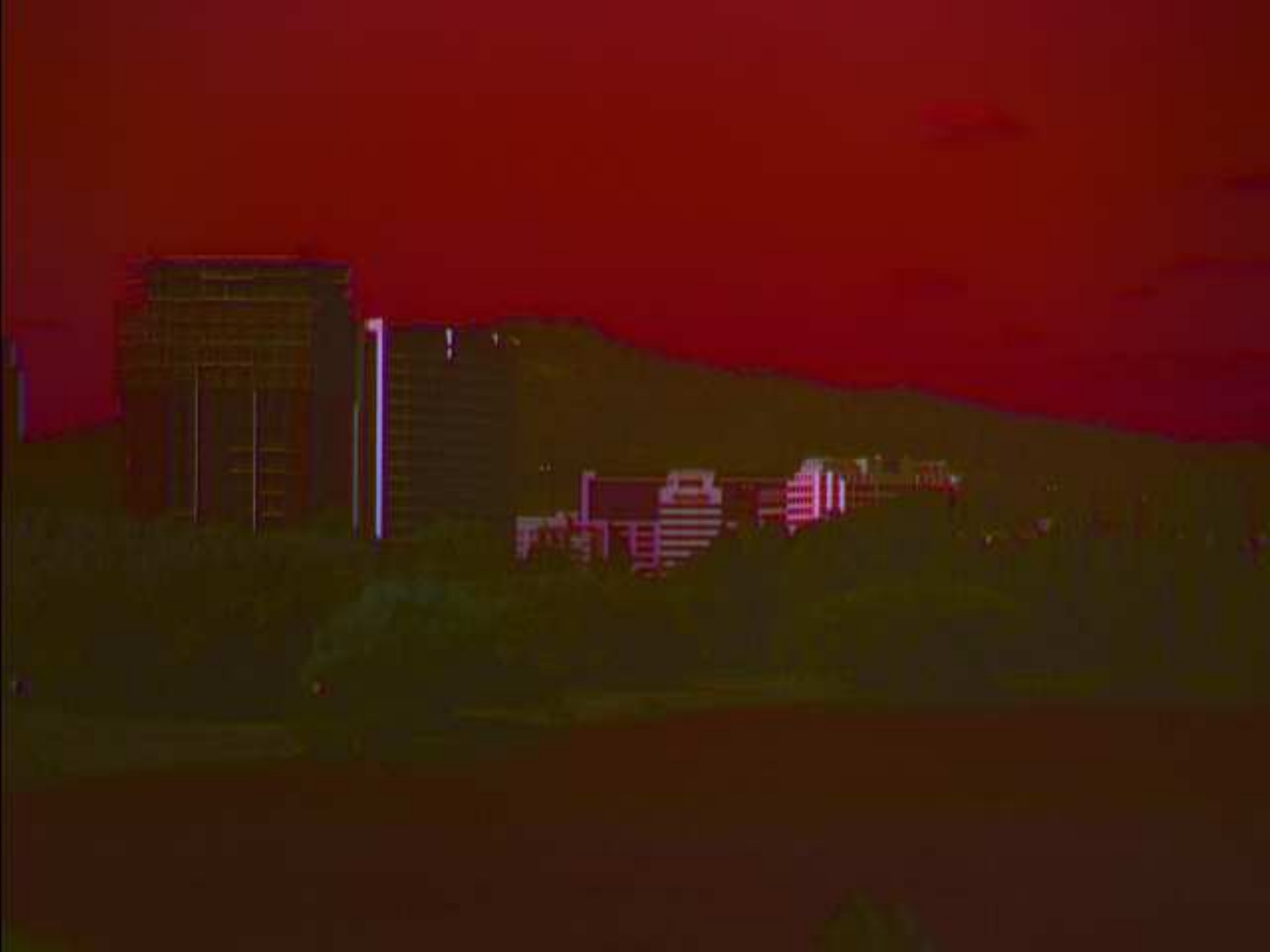}
}
\subfigure[]{
\includegraphics[width=0.23\linewidth]{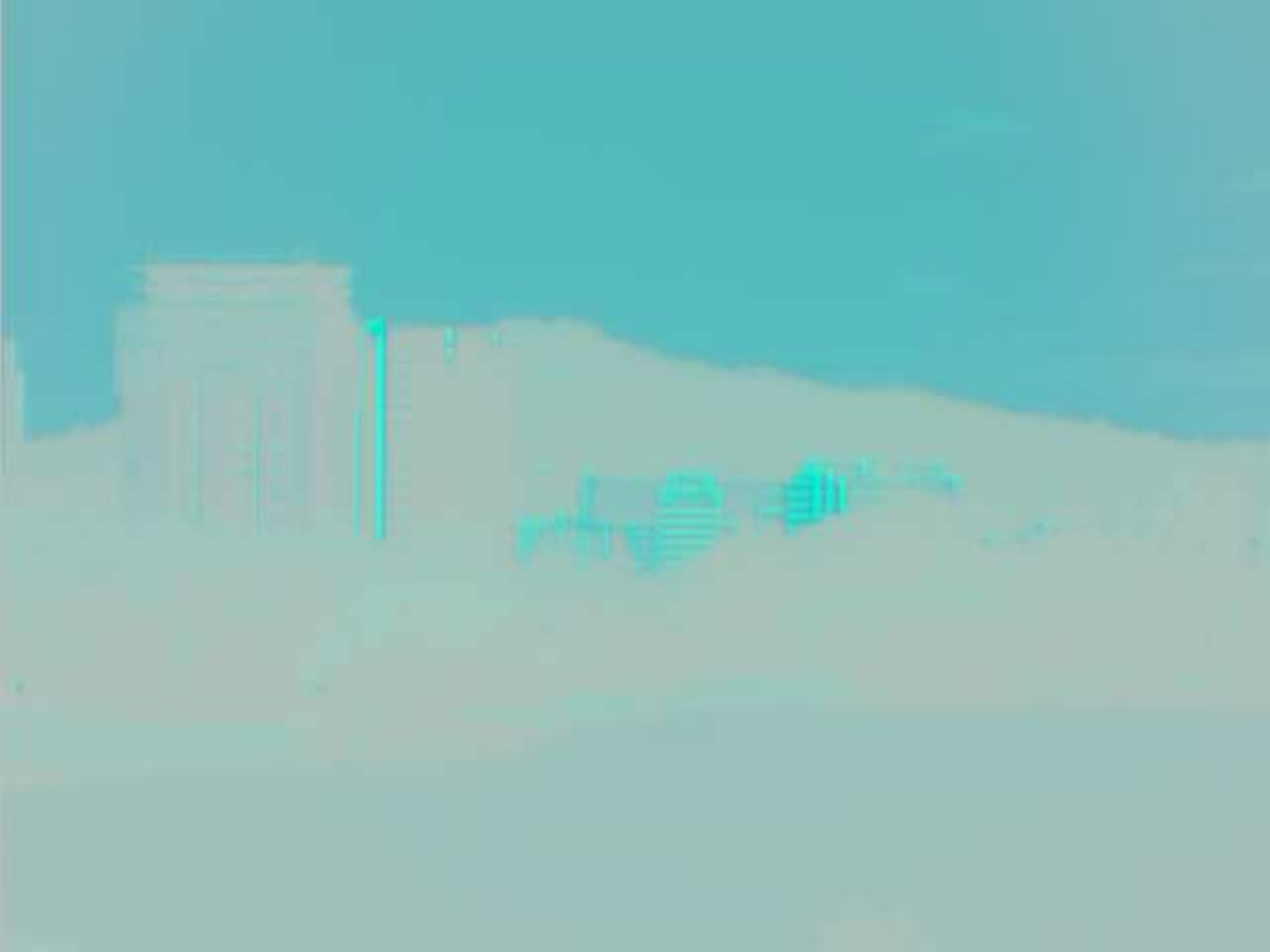}
}
\subfigure[]{
\includegraphics[width=0.23\linewidth]{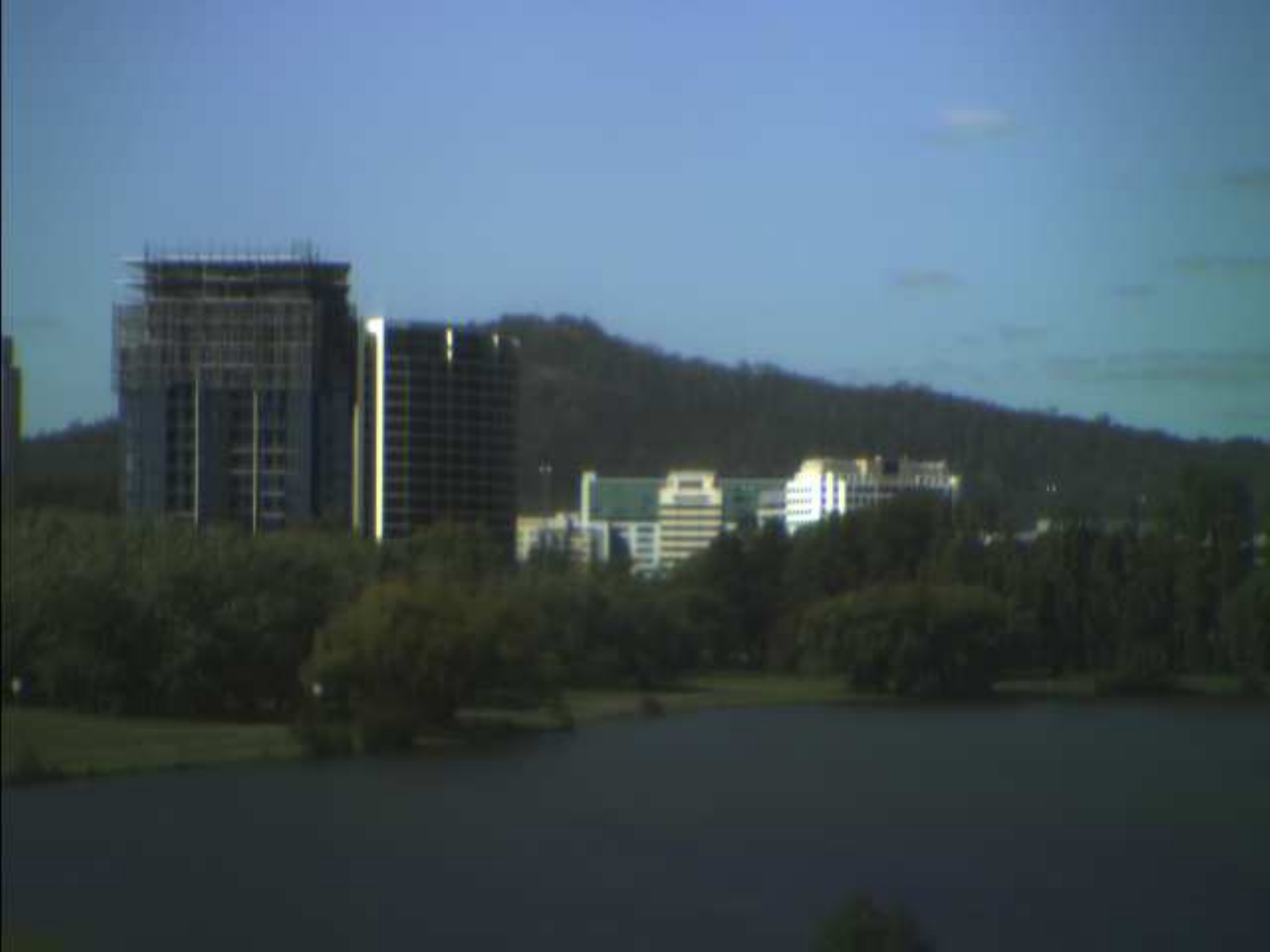}
}
\subfigure[]{
\includegraphics[width=0.23\linewidth]{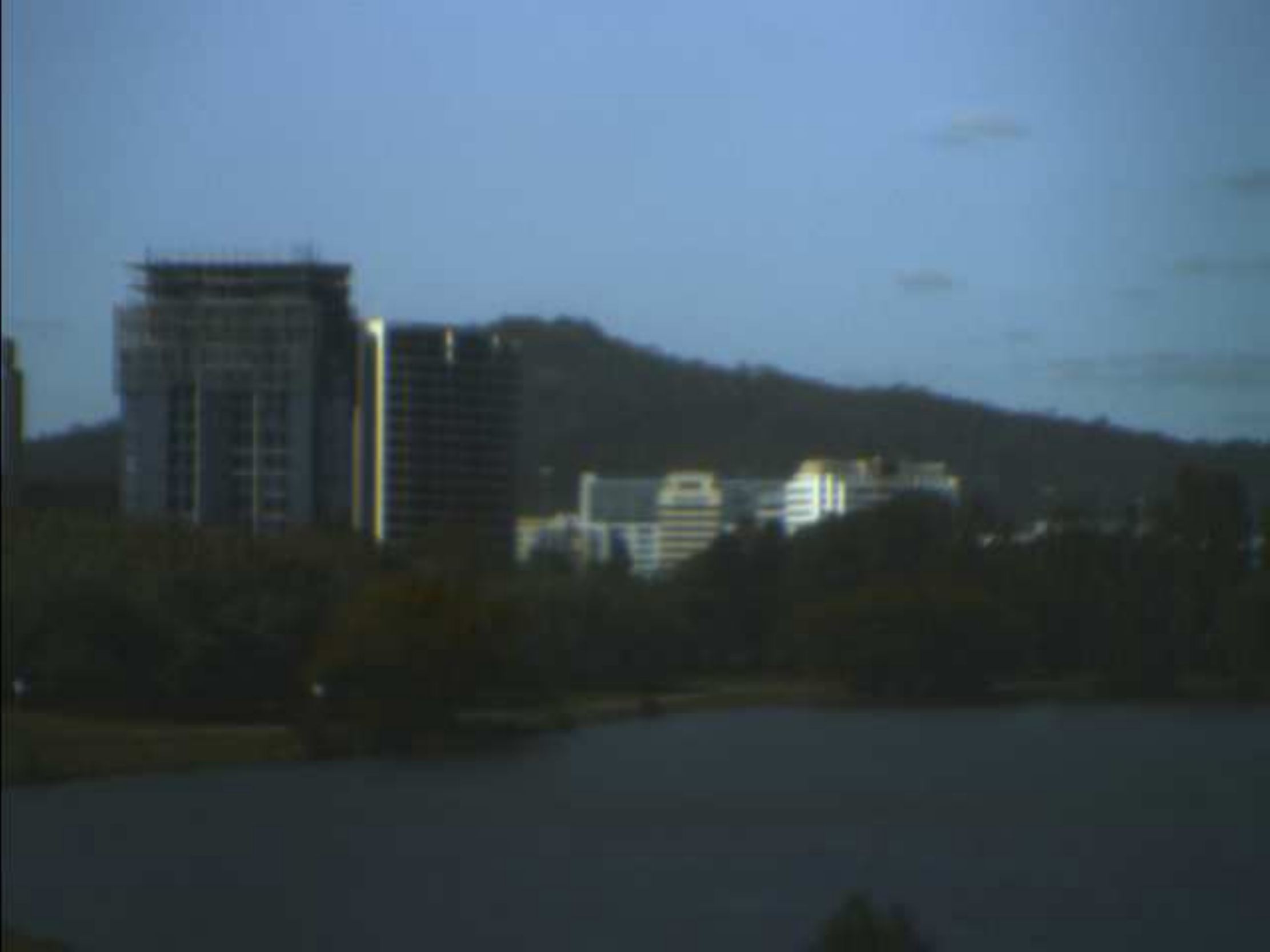}
}
\subfigure[]{
\includegraphics[width=0.23\linewidth]{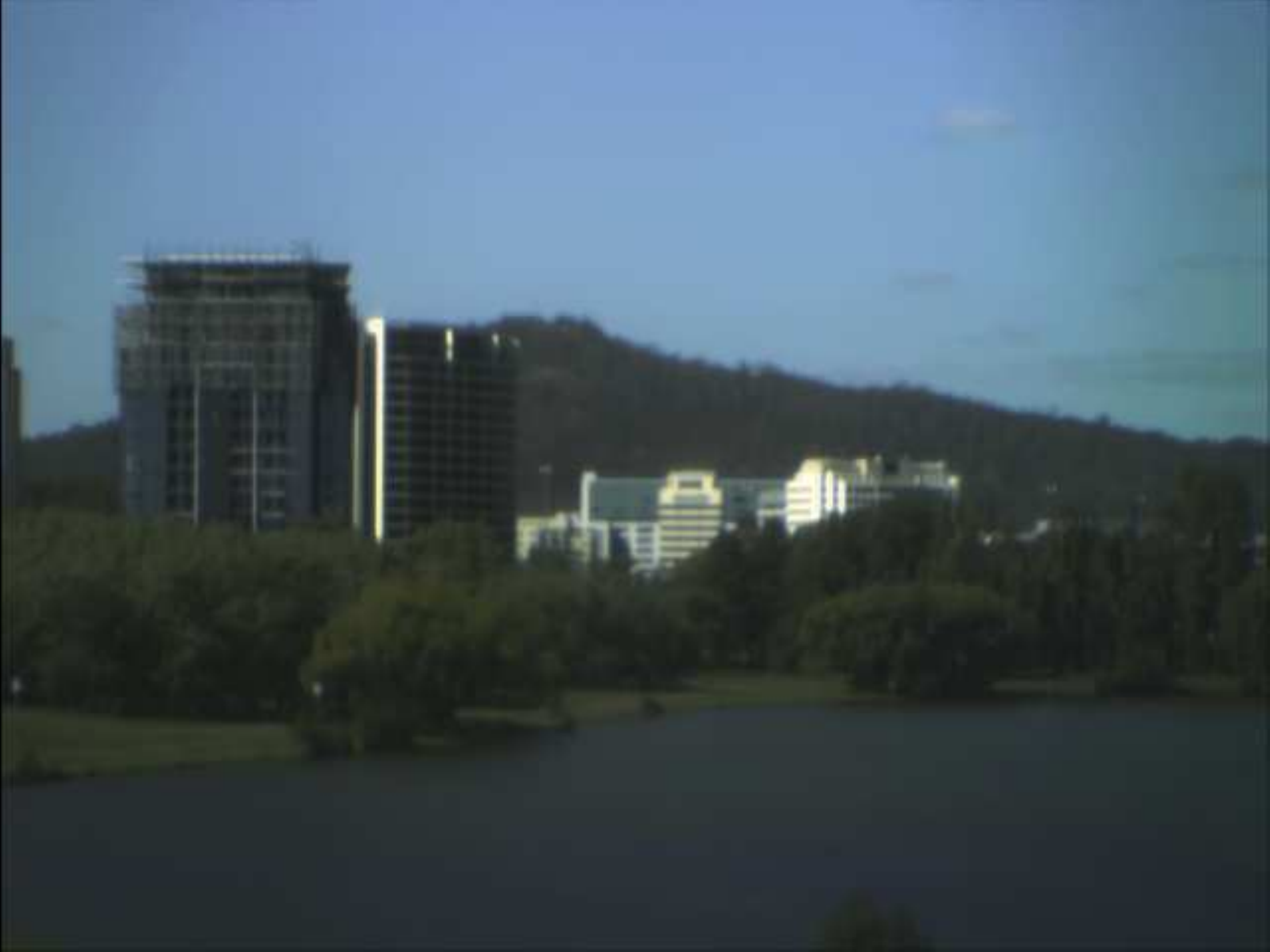}
}
\caption{Visual comparison of different visualization approaches on the G03 data set. (a) Stretched CMF. (b) Bilateral filtering. (c) Bicriteria optimization. (d) Laplacian Eigenmaps. (e) LPP. (f) Manifold alignment. (g) The proposed feature-level learning. (h) The proposed instance-level learning.}
\label{G03comparison}
\end{figure*}

\begin{figure*}[!tp]
\centering
\subfigure[]{
\includegraphics[width=0.23\linewidth]{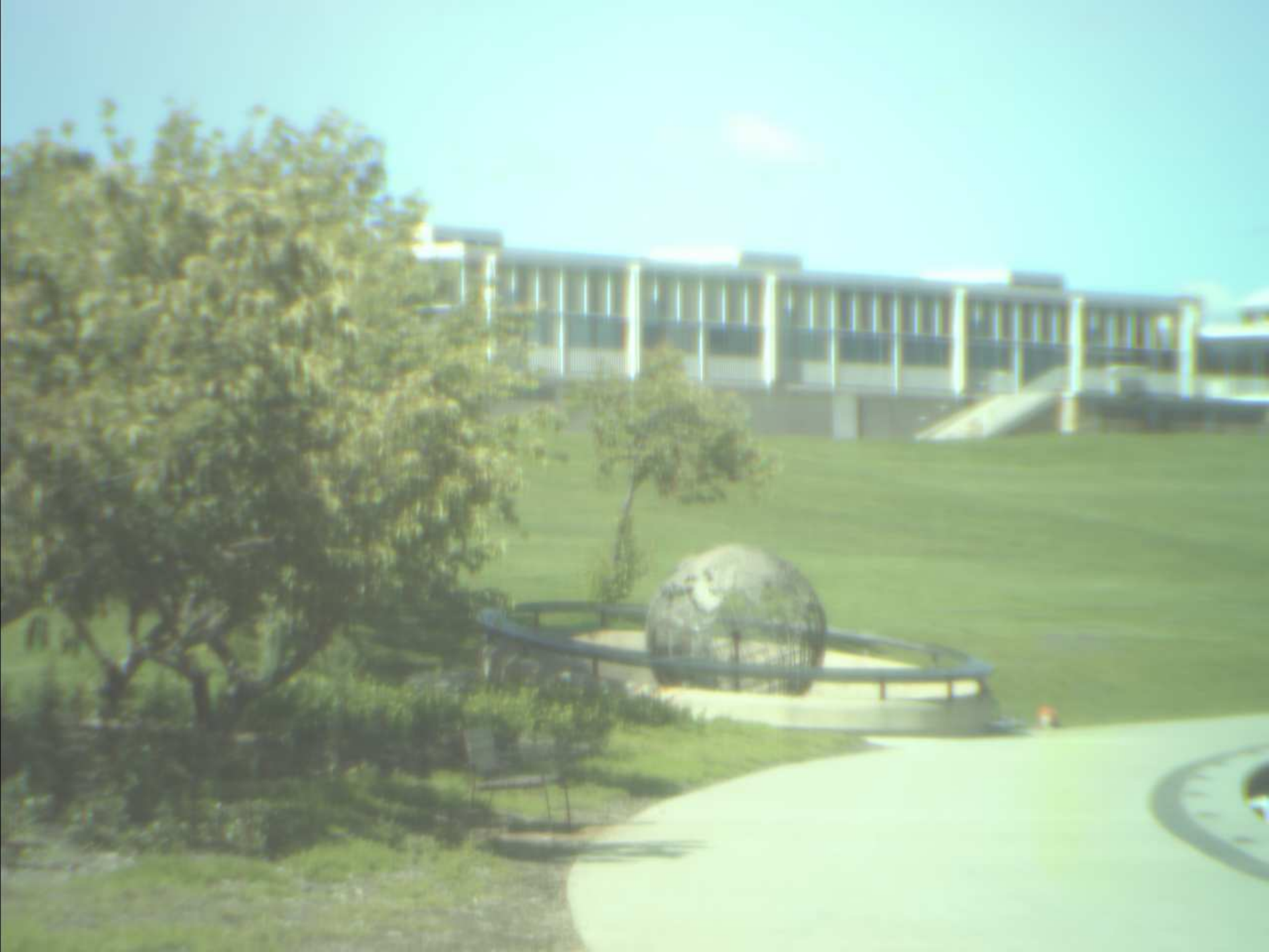}
\label{d04CMF}
}
\subfigure[]{
\includegraphics[width=0.23\linewidth]{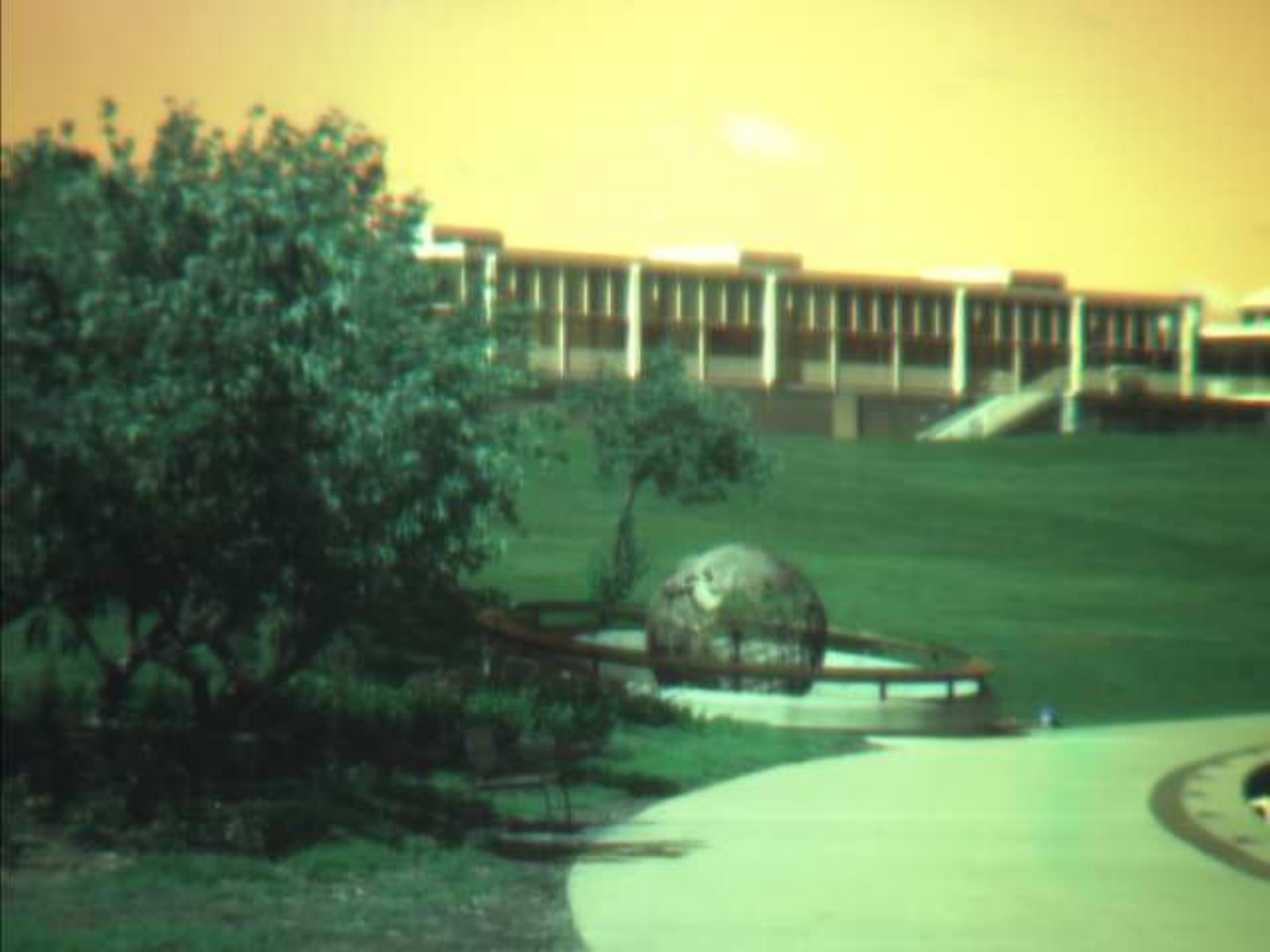}
\label{d04BF}
}
\subfigure[]{
\includegraphics[width=0.23\linewidth]{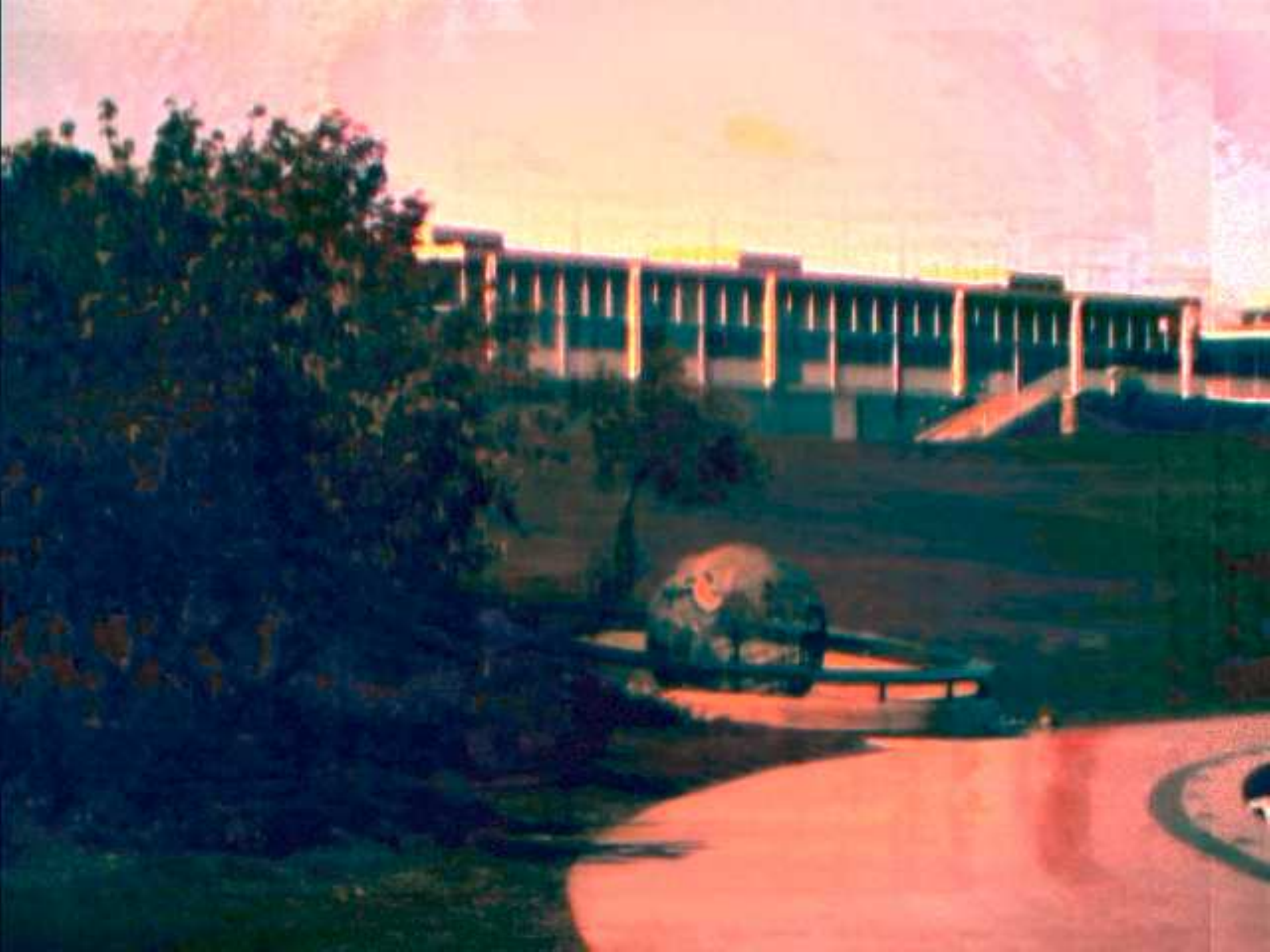}
\label{d04DCOCDM}
}
\subfigure[]{
\includegraphics[width=0.23\linewidth]{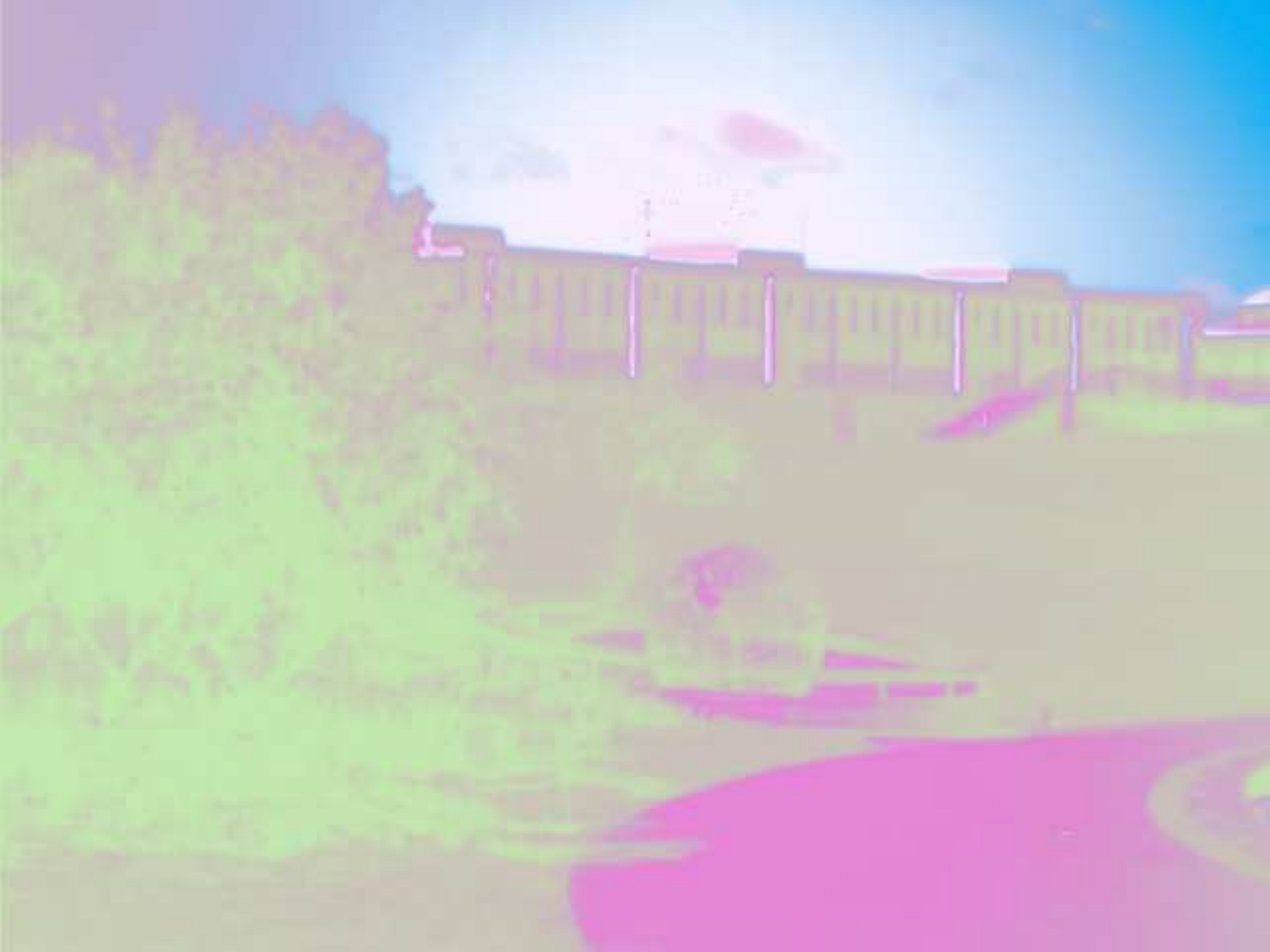}
}
\subfigure[]{
\includegraphics[width=0.23\linewidth]{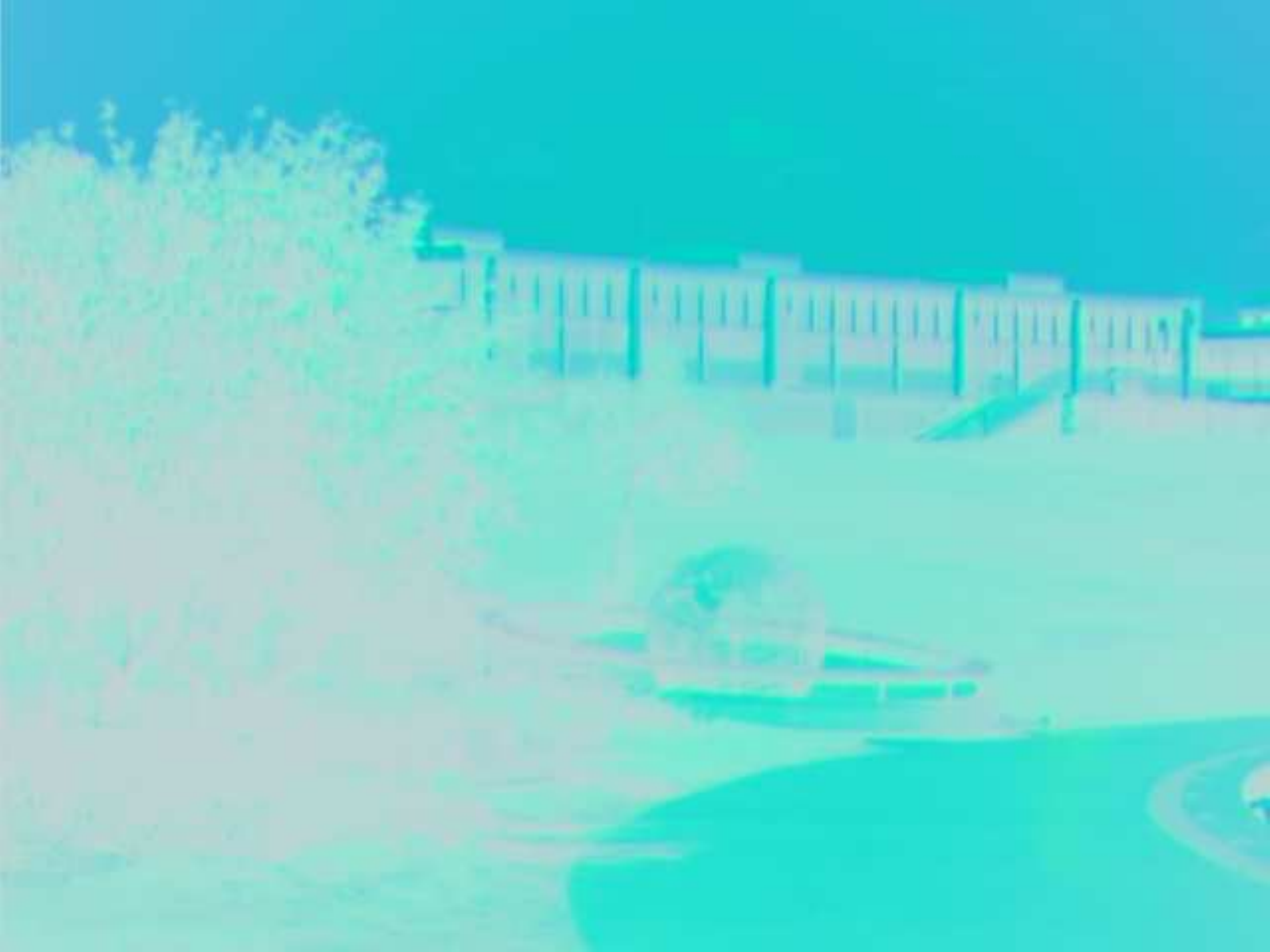}
\label{d04bandselection}
}
\subfigure[]{
\includegraphics[width=0.23\linewidth]{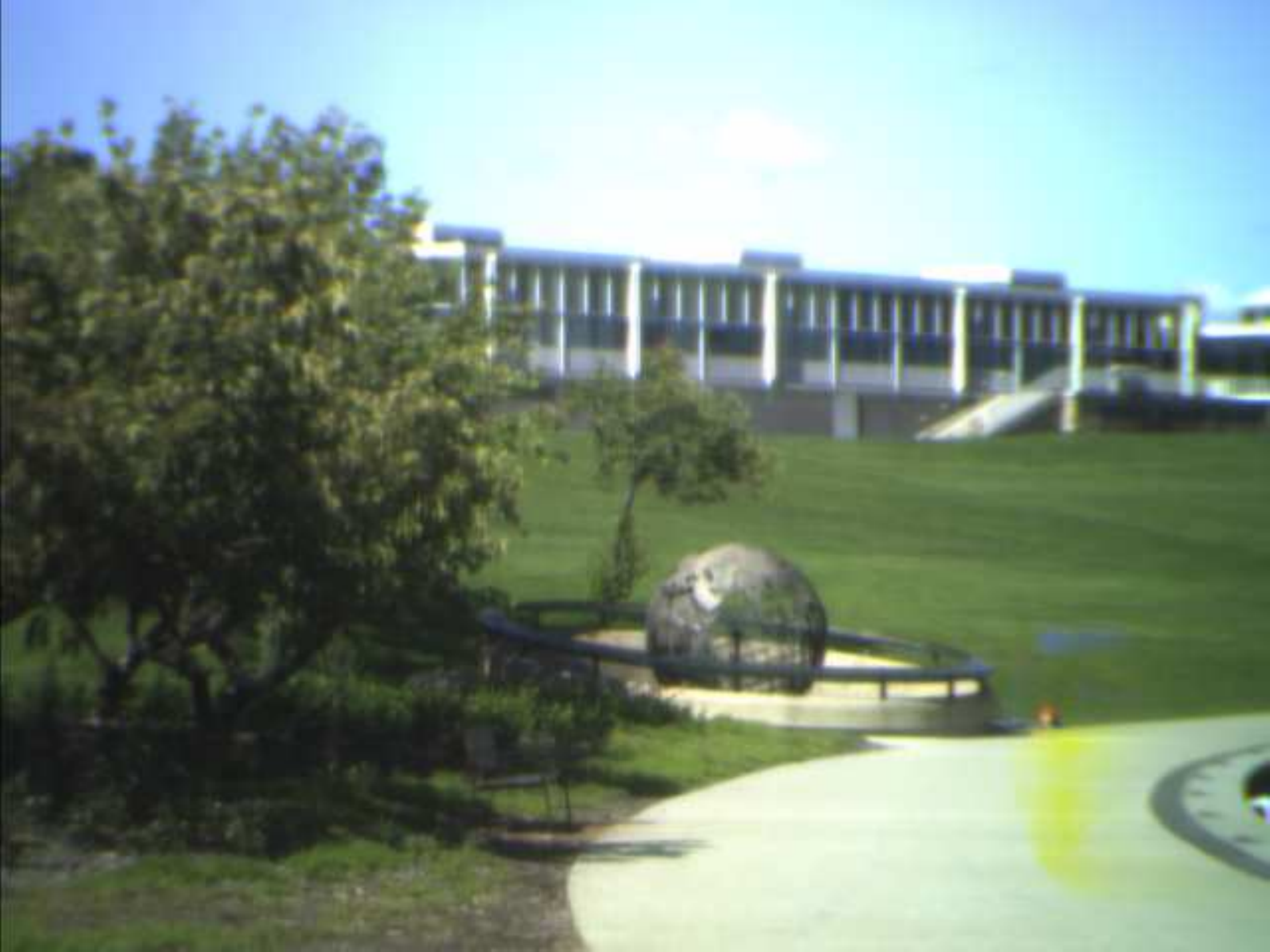}
}
\subfigure[]{
\includegraphics[width=0.23\linewidth]{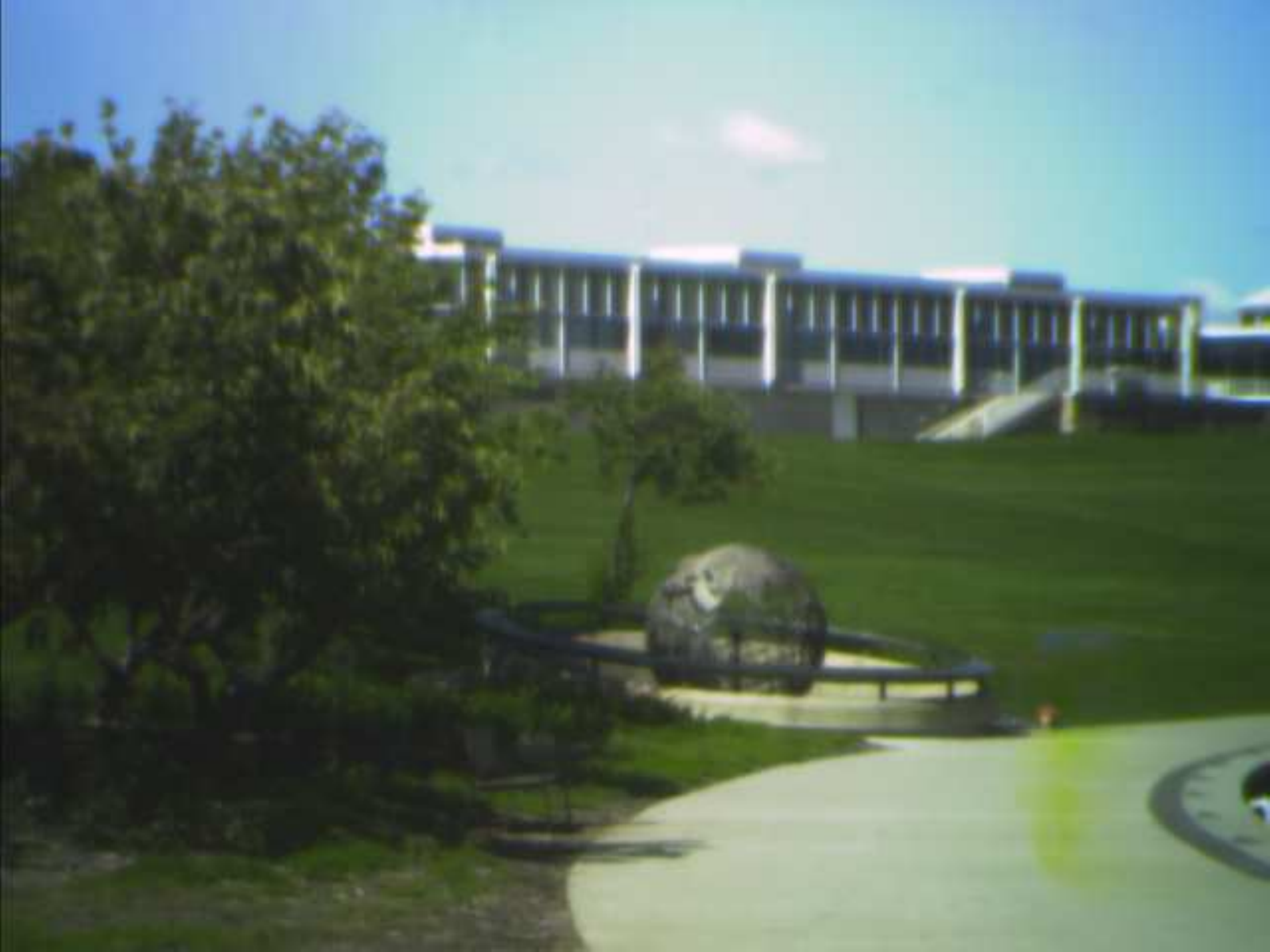}
\label{d04_linear}
}
\subfigure[]{
\includegraphics[width=0.23\linewidth]{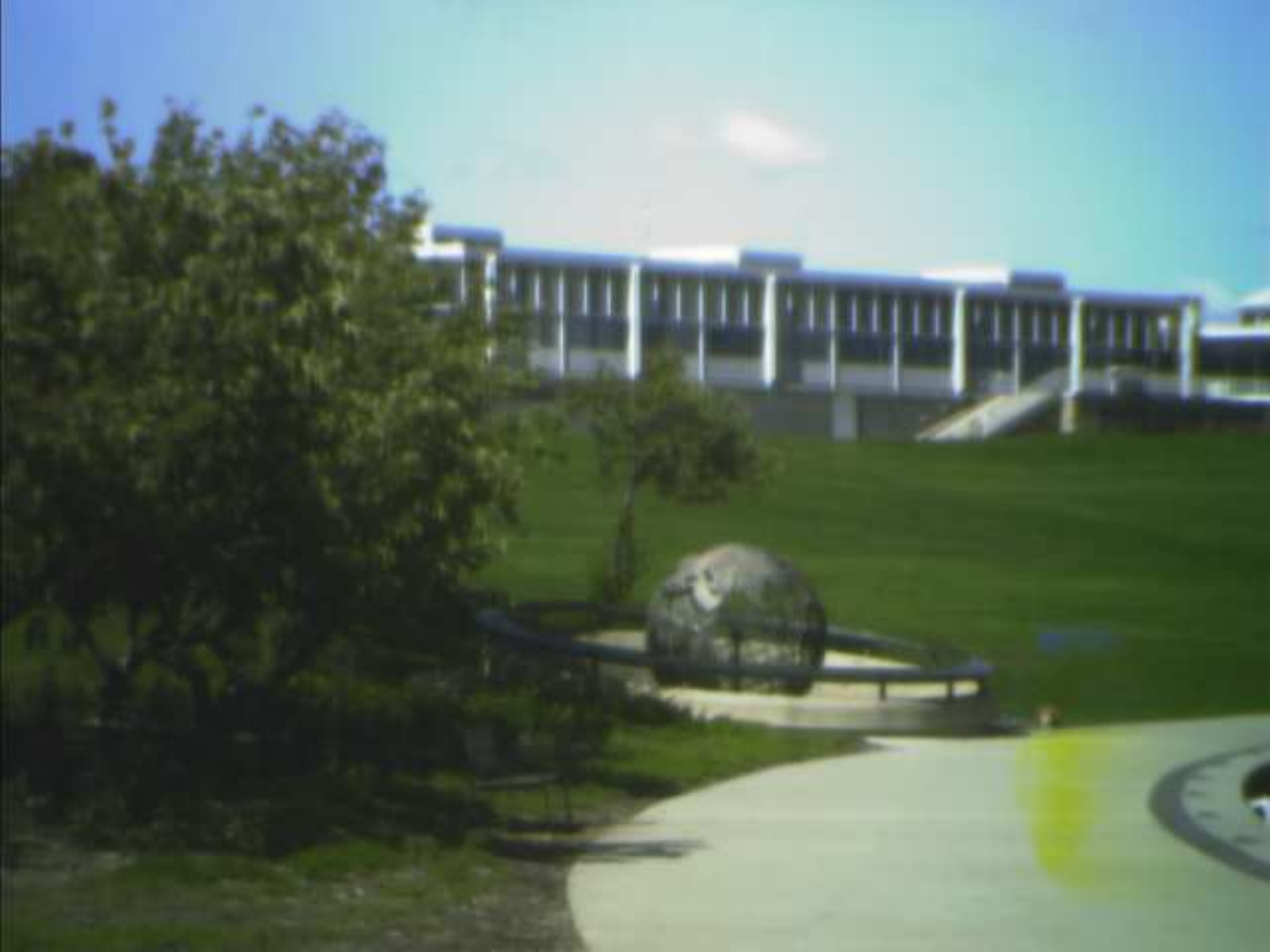}
\label{d04_nonlinear}
}
\caption{Visual comparison of different visualization approaches on the D04 data set. (a) Stretched CMF. (b) Bilateral filtering. (c) Bicriteria optimization. (d) Laplacian Eigenmaps. (e) LPP. (f) Manifold alignment. (g) The proposed feature-level learning. (h) The proposed instance-level learning.}
\label{D04comparison}
\end{figure*}

\section{Experiments}
\label{Experiments}
The experiments are conducted on several HSI data sets captured by aerobat/satellite-based or ground-based
hyperspectral imaging sensors.
Their corresponding RGB images are obtained from their own spectra or from other RGB cameras.
We compare our method against some state-of-the-art HSI visualization approaches including stretched CMF~\cite{jacobson2005design}, LPP~\cite{niyogi2004locality},
bilateral filtering~\cite{kotwal2010visualization}, bicriteria optimization~\cite{mignotte2012bicriteria}, Laplacian Eigenmaps~\cite{belkin2003laplacian} and manifold alignment~\cite{liaomanifold}.
Both subjective visual judgment and objective metric are used to evaluate the visualization results.
\subsection{Experimental Settings}
Five HSI data sets used for experiments are introduced in this section. As their spectral ranges cover most of the visible wavelengths, their corresponding RGB images can be generated by stacking three channels from the visible wavelengths.

The first HSI data set was taken over the Washington D.C. mall by the Hyperspectral Digital Imagery Collection Experiment (HYDICE) sensor.
The data consists of 191 bands after noisy bands removed. The size of each band image is $1208\times 307$.
Fig.~\ref{fig:band50} shows its 50th band image. Its corresponding RGB image was constructed by the 12th, 35th and 41st bands in the visible wavelength ranges, and is shown in Fig.~\ref{fig:registered}.

The second HSI data set was acquired over the University of Pavia, Italy by the
ROSIS-03 (Reflective Optics Systems Imaging Spectrometer) hyperspectral sensor.
The data consists of 103 bands after removing the noisy bands, and the size of band image is $610 \times 340$.
Fig.~\ref{fig:PaviaUBand50} shows its 50th band image. The corresponding RGB image was generated by its 10th, 31st and 46th bands, is shown in Fig.~\ref{fig:RGBPaviaU}.

The third data set was captured by the Airborne Visible/Infrared Imaging Spectrometer (AVIRIS) over Moffett Field, California at the southern end of San Francisco Bay.
The data consists of 224 bands. Each band image has the size of $501\times 651$.
Its 50th band image is shown in Fig.~\ref{fig:MoffettBand50}.
The corresponding RGB image was generated by its 6th, 17th and 36th bands and is shown in Fig.~\ref{fig:RGBmoffett}.

The other two HSI data sets named ``G03'' and ``D04'' were captured by a ground-based OKSI hyperspectral imaging system mounted with a tunable LCTF filter.
Each data set has 18 bands ranging from $460nm$ to $630nm$ at $10nm$ interval.
The size of each band image is $480\times 640$.
Fig.~\ref{fig:G03band1} and Fig.~\ref{fig:D04band1} show the first band images of G03 and D04, respectively.
The corresponding RGB images generated by their 3rd, 9th and 13th bands are shown in Fig.~\ref{fig:RGBG03} and Fig.~\ref{fig:RGBD04}, respectively.

In the experiments, the corresponding RGB images were converted to the uncorrelated $\textit{L}\alpha\beta$ space before they are used for HSI visualization.
$\textit{L}\alpha\beta$ aspires to perceptual uniformity, and its $L$ component closely matches human perception of lightness.
Besides, the $\textit{L}\alpha\beta$ space minimizes the correlation between channels, which makes it more appropriate than the RGB space for color processing.
To obtain a set of matching pairs as the color constraint, we randomly selected $10\%$ of the pixels in the HSI, and then find their  matching pixels in the corresponding RGB images.

\subsection{Visual Comparison}
\label{secComparison}

Fig.~\ref{DCcomparison} shows the visual comparison of different visualization approaches on the Washington D.C. mall data.
It can be seen that both the output images of the proposed instance-level and feature-level manifold learning methods have very natural colors
similar to the corresponding RGB image in Fig.~\ref{fig:registered}, which makes our results much easier to understand.
Also, the details in the HSI are well presented by our method.
Note that the result of manifold alignment (Fig.~\ref{fig:DCMA}) also has very natural colors.
Compared to manifold alignment which fuses two image structures, the proposed  method does not require a corresponding RGB image that is registered precisely, thus it is more flexible in choice of the color constraints.
Besides, the proposed method also saves the time to compute the graph Laplacian for the corresponding RGB image, which is required for manifold alignment.

The comparative results on the University of Pavia, Moffett field, G03 and D04 data sets are given in Fig.~\ref{PaviaExperiments}-\ref{D04comparison}, respectively.
Likewise, in these experiments our approaches not only produce natural-looking visualizations but also preserve the details in the original HSIs.
For example, Fig.~\ref{MoffettExperiments} shows the visualizations of Moffett field, in which the roads in our results are more distinguishable than other manifold learning-based methods such as the Laplacian Eigenmaps, LPP and manifold alignment.
In Fig.~\ref{D04comparison}, the proposed manifold learning approaches can display the clouds clearly, which are not shown or less easy to perceive in some other results such as those of bilateral filtering and bicriteria optimization. It is concluded that the proposed method is able to generate a trichromatic image with natural colors while preserving the spectral information and manifold structure of HSI, with the color information provided by the HSI itself.
\begin{figure}[!tp]
\centering
\subfigure[]{
\includegraphics[width=0.3\linewidth]{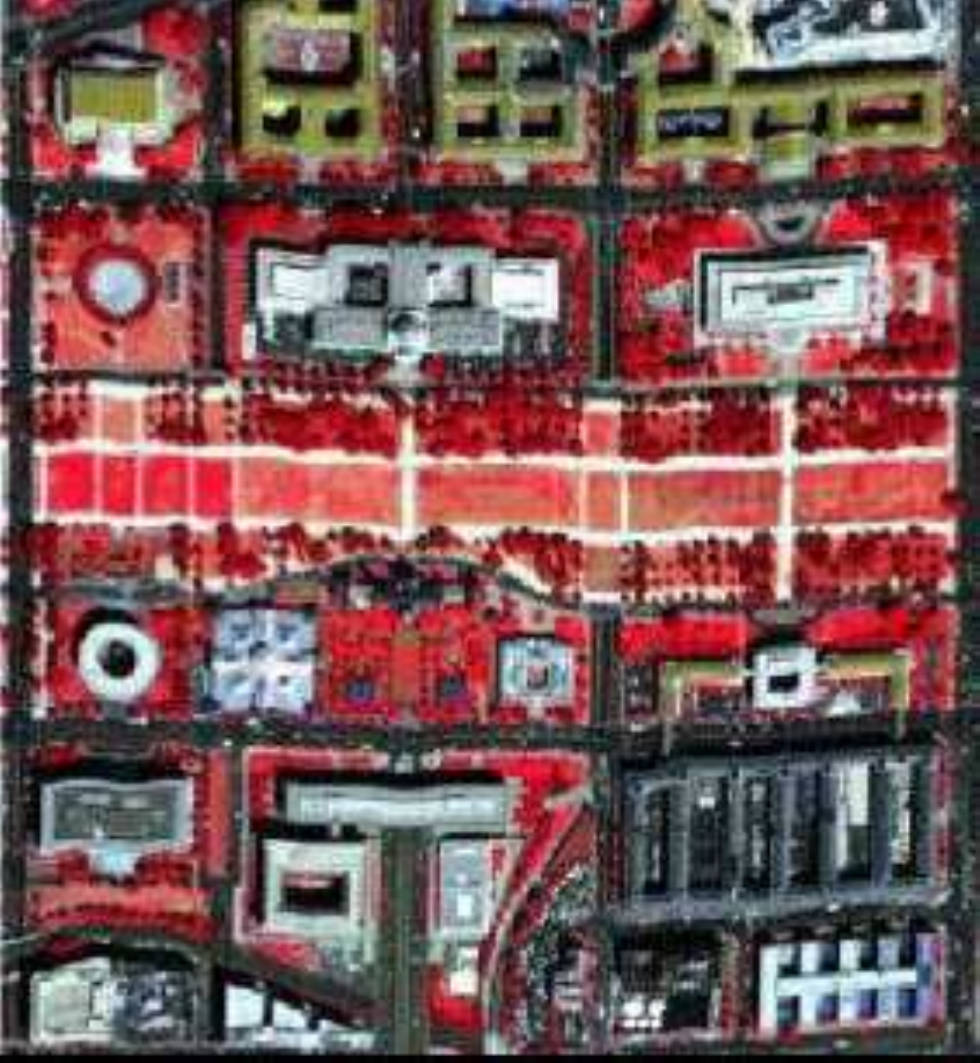}
}
\subfigure[]{
\includegraphics[width=0.3\linewidth]{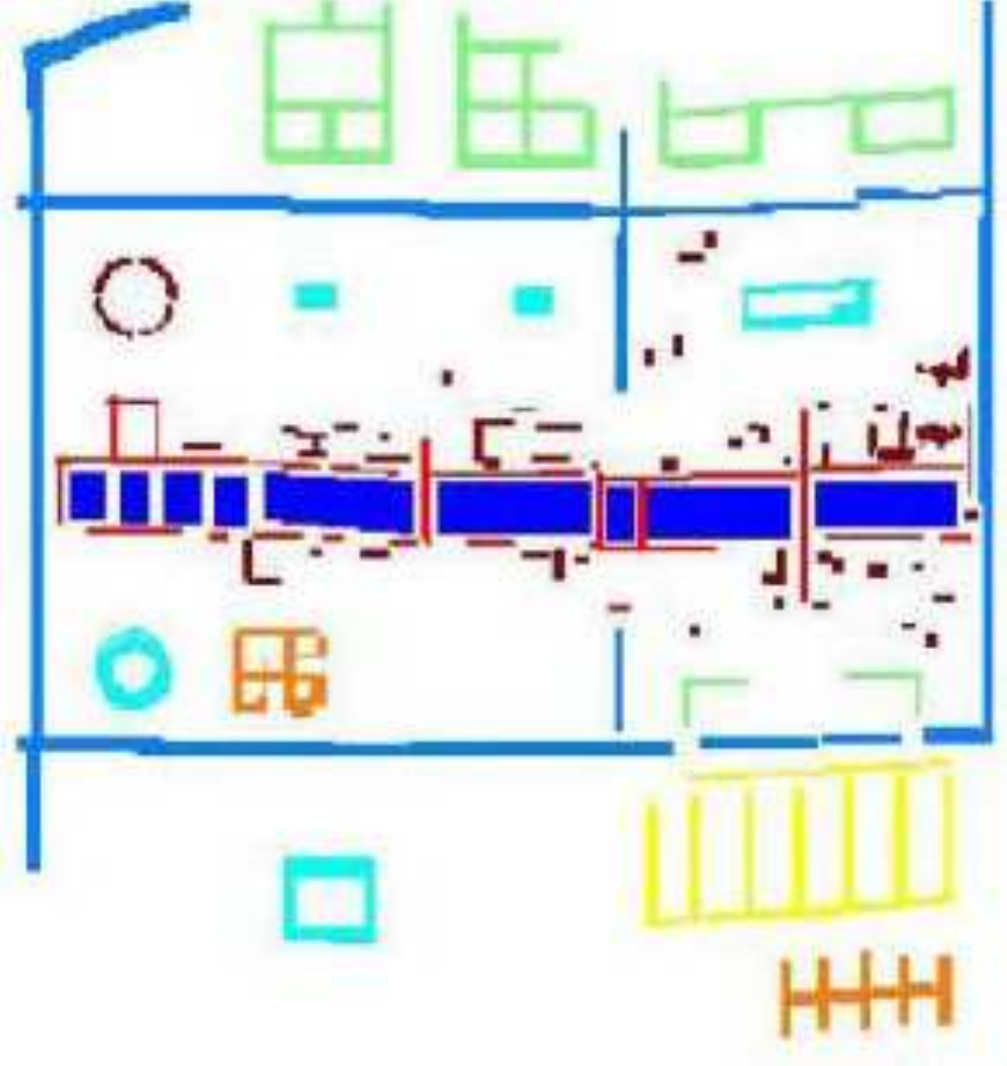}
}
\subfigure[]{
\includegraphics[width=0.17\linewidth]{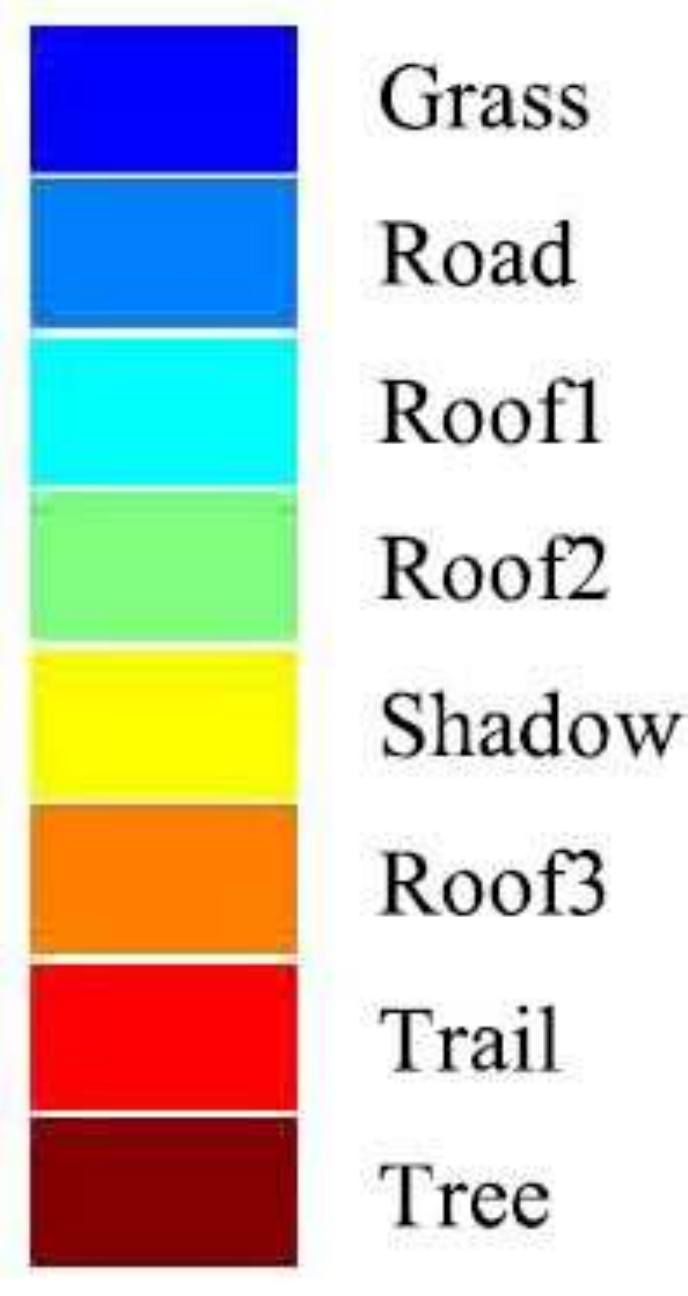}
}
\caption{A sub-image of Washington DC Mall data and the ground truth.}
\label{DCmap}
\end{figure}
\begin{figure}[!tp]
\centering
\subfigure[]{
\includegraphics[width=0.6\linewidth]{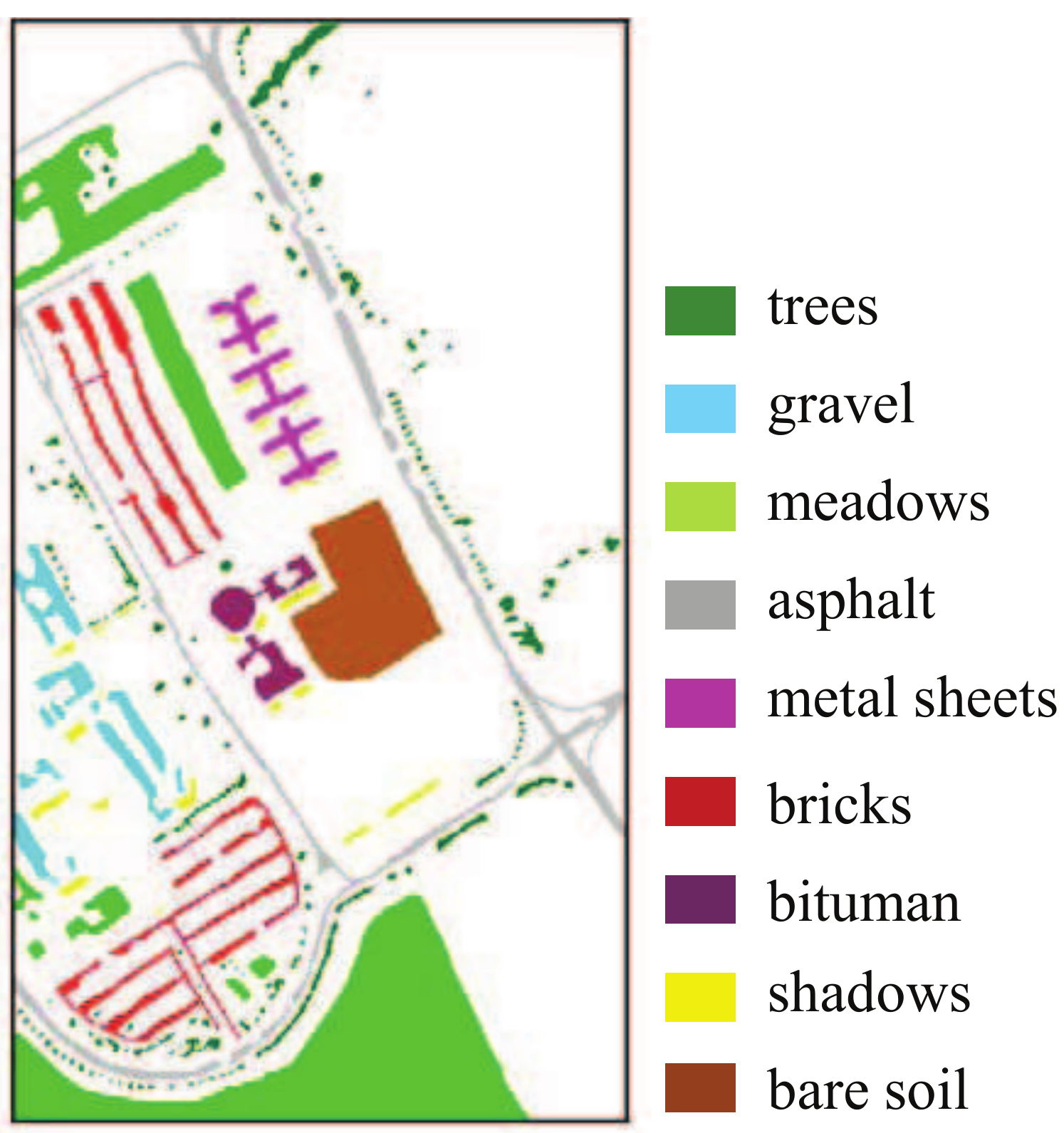}
}
\caption{The ground truth of University of Pavia data set.}
\label{Paviamap}
\end{figure}

\subsection{Quantitative Comparison via Classification}

There is no universally accepted standard for quantitative assessment of HSI visualization.
In our experiments, we used classification as a quantitative measurement for visualization, based on the assumption that if the visualized image with reduced spectral dimensions has higher classification accuracy, it preserves more information of the original HSI.
We performed classification on the visualized images using support vector machines (SVMs) equipped with the RBF kernel on the Washington DC Mall data set and the University of Pavia data set. The parameters of SVM were tuned by 5-fold cross validation.
The classification accuracy is assessed with overall accuracy (OA), average accuracy (AA), and kappa coefficient of agreement ($\kappa$).

The experimental settings for each data set are described as follows:
\begin{itemize}
  \item \emph{Washington DC Mall Data Set:} we used a sub-image with the size of $305 \times 280$ to perform the experiment. The sub-image and the ground truth map are shown in Fig.~\ref{DCmap}, which contains 14266 labeled samples. The number of labeled samples in each class is shown in Table~\ref{numberofsamplesDC}. We randomly chose $10,30,70$ labeled samples per class as the training set, and used all the remaining samples as the test set.

  \item \emph{University of Pavia  Data Set:} The ground truth map is shown in Fig.~\ref{Paviamap} where the training set contains 3921 samples and the test set contains 40002 samples. The number of training and testing samples in each class is shown in Table~\ref{numberofsamplesPavia}. We randomly selected $1 \%$, $10 \%$, and $100 \%$ samples from the training set to train the classifier.
\end{itemize}
\begin{table}[!tp]
\centering
 \caption{Number of labeled samples for each class in the Washington DC Mall data set.}
 \label{numberofsamplesDC}
  \begin{tabular}{c|c}
    \hline
    Class& Labeled Samples \\
    \hline
    Grass&3133\\
    Road&4146\\
    Roof1&1067\\
    Roof2& 1928\\
    Shadow&1013\\
    Roof3&818\\
    Trail&1096\\
    Tree&1065\\
    \hline
  \end{tabular}
\end{table}

\begin{table}[!tp]
\centering
 \caption{Number of labeled samples for each class in the University of Pavia data set.}
  \label{numberofsamplesPavia}
  \begin{tabular}{c|c}
    \hline
    Class&Training / Testing \\
    \hline
    Trees&524 / 3064\\
    Gravel&392 / 2099\\
    Meadows&540 / 18649\\
    Asphalt& 548 / 6631\\
    Metal sheets&265 / 1345\\
    Brciks&514 / 3682\\
    Bituman&375 / 1330\\
    Shadows&231 / 947\\
    Bare soil&532 / 5029\\
    \hline
  \end{tabular}
\end{table}
As one of our baseline, we also performed classification on the corresponding RGB images, which were generated by band selection from the visible wavelength range of the HSIs.
The classification results of compared visualization methods on two data sets are reported in Table~\ref{TableDC}  and Table~\ref{TablePavia}, respectively.
All the experimental results are obtained by randomly selecting  training samples 10 times and averaging the testing results.
In most cases both the proposed feature-level and instance-level manifold learning methods have better classification performance than other approaches, indicating that the composite kernel applied in manifold learning is very helpful in information preservation.
It can also be seen that instance-level learning method produces slightly higher accuracies than feature-level learning in most cases, which indicates the nonlinear structure in the HSIs is well preserved by instance-level method.

It should be noted that although our visualized images are visually similar to their corresponding RGB images generated by band selection from the visible wavelength range, the classification accuracies of the former ones are greatly higher than the latter ones.
This indicates that our method have fused the information from all the spectral bands rather than just three selected bands in the visible wavelength range.

\begin{table*}[!tp]
\centering
 \caption{Classification results on the Washington DC Mall data set.}
  \label{TableDC}
  \begin{tabular}{c|c|c|c|c|c|c|c|c|c|c}
    \hline
    ~&~ &RGB band&CMF &Bilateral &Bicriteria &Laplacian&LPP&Manifold&Feature-level  & Instance-level \\
    ~&~ &selection&~&filter&optimization&Eigenmaps&~&alignment&learning&learning \\
    \hline
    10 training samples&OA($\%$)&86.78&88.42&88.66&79.90&81.13&87.50&87.43&89.09&\textbf{90.50}\\
per class&AA($\%$)&84.49&86.53&84.66&73.48&76.80&83.35&85.47&87.39&\textbf{89.27}\\
~&$\kappa$&84.09&86.01&86.25&75.69&77.27&84.83&84.88&86.84&\textbf{88.55}\\
\hline
30 training samples&OA($\%$)&90.78&92.46&91.19&82.92&84.07&90.98&91.49&92.52&\textbf{93.64}\\
per class&AA($\%$)&89.88&92.17&88.92&76.94&80.80&89.34&90.65&92.06&\textbf{93.17}\\
~&$\kappa$&88.89&90.90&89.36&79.29&80.80&89.09&89.75&90.97&\textbf{92.32}\\
\hline
70 training samples&OA($\%$)&91.29&93.51&92.84&84.12&85.46&91.66&92.42&93.40&\textbf{94.55}\\
per class&AA($\%$)&90.64&93.15&90.82&79.03&83.33&90.14&91.86&93.12&\textbf{94.38}\\
~&$\kappa$&89.50&92.15&91.34&80.77&82.48&89.90&90.86&92.03&\textbf{93.42}\\
    \hline
  \end{tabular}
\end{table*}

\begin{table*}[!tp]
\centering
 \caption{Classification results on the University of Pavia data set.}
  \label{TablePavia}
  \begin{tabular}{c|c|c|c|c|c|c|c|c|c|c}
    \hline
    ~&~ &RGB band&CMF &Bilateral &Bicriteria &Laplacian&LPP&Manifold&Feature-level  & Instance-level \\
    ~&~ &selection&~&filter&optimization&Eigenmaps&~&alignment&learning&learning \\
    \hline
    $1\%$&OA($\%$)&60.54&61.15&63.30&53.31&60.33&62.97&62.94&\textbf{66.09}&61.03\\
    training samples&AA($\%$)&63.05&66.04&\textbf{68.73}&57.52&64.78&68.62&66.83&68.25&67.41\\
    ~&$\kappa$&49.51&50.53&53.18&41.81&49.93&52.83&52.36&\textbf{55.95}&50.74\\
    \hline
    $10\%$&OA($\%$)  &67.46&70.79&70.43&58.79&68.06&69.20&70.97&72.10&\textbf{72.60}\\
   training samples& AA($\%$)  &69.27&72.99&73.21&66.02&68.67&71.76&72.81&72.85&\textbf{74.58}\\
    ~&$\kappa$&57.00&61.15&60.99&48.17&57.90&59.51&61.44&62.57&\textbf{63.71}\\
    \hline
    $100\%$&OA($\%$)&70.56&\textbf{74.33}&70.33&65.97&68.78&70.08&72.10&73.08&74.22\\
    training samples&AA($\%$)&71.66&75.15&74.89&70.05&71.93&74.34&74.52&73.91&\textbf{76.17}\\
    ~&$\kappa$&60.68&65.41&61.08&55.78&58.90&60.59&63.00&63.96&\textbf{65.69}\\
    \hline
  \end{tabular}
\end{table*}
\begin{table*}[!tp]
\centering
 \caption{Preservation of distance.}
  \label{Preservationofdistance}
  \begin{tabular}{c|c|c|c|c|c|c|c|c}
    \hline
   ~ &CMF &Bilateral &Bicriteria &Laplacian&LPP&Manifold&Feature-level  & Instance-level \\
   ~ &~&filter&optimization&Eigenmaps&~&alignment&learning&learning \\
    \hline
    Washington DC Mall&0.6339& \textbf{0.8895} &0.8439 &0.7324 &0.6264 &0.5397&0.6342& 0.6330 \\
    Pavia University&0.8101& 0.8362 &0.9030 &0.8815 &\textbf{0.9211} &0.8948&0.9052& 0.8810 \\
    Moffett Field &0.7523& 0.9043 &0.8685 &0.6884 &0.9020 &0.5021&\textbf{0.9041}& 0.8992\\
    G03&0.9819& 0.9884 &0.9855 &0.9557 &0.9841 &0.9870& \textbf{0.9902} &0.9620\\
    D04&0.9077& \textbf{0.9754 }&0.9402 &0.6672 &0.8248 &0.9294&0.9417& 0.9404 \\
 \hline
  \end{tabular}
\end{table*}
\begin{table*}
\centering
 \caption{Running time (in seconds).}
  \label{runningTime}
  \begin{tabular}{c|c|c|c|c|c|c|c|c}
    \hline
   ~ &CMF &Bilateral &Bicriteria &Laplacian&LPP&Manifold&Feature-level  & Instance-level \\
   ~ &~&filter&optimization&Eigenmaps&~&alignment&learning&learning \\
    \hline
    Washington DC Mall&2.28& 557.80&46.42&2444.41&1921.45&2086.49&1755.08&1901.49\\
    Pavia University&1.47&176.67&23.39&658.293&260.77&302.75&262.49&263.20\\
    Moffett Field &2.13&497.69&37.87&1576.63& 1257.05&1341.81&1222.79&1310.71\\
    G03&1.13& 45.08&30.50&1135.51&105.46&126.23&101.98&118.65\\
    D04&1.09&45.24&30.36&590.63&101.89&123.71&102.82&130.41\\
    \hline
    Average& 1.63& 264.50&33.69&1281.10&729.32&795.61&689.03&744.89\\
 \hline
  \end{tabular}
\end{table*}
\subsection{Quantitative Comparison in terms of preservation of distance}

As pointed out in~\cite{jacobson2005design}, one of the design goals for HSI visualization is ``smallest effective differences'', which means visual distinctions are no larger than needed to effectively show relative differences.
We apply the metric of Preservation of Distances proposed in ~\cite{cui2009interactive} to evaluate the visualization results, which
measures the consistency of the perceptual color distances in the visualized image and the Euclidean distances in the original HSI.
Let $\textbf{x}$ be the vector of all pairwise Euclidean distances of the pixels in the high-dimensional spectral space, and let vector $\textbf{y}$ be the corresponding pairwise Euclidean distances of the pixels in the visualized images in $\textit{L}\alpha\beta$ space.
The preservation of distance is defined as a correlation-based metric $\gamma$, which is defined as the following formula:
\begin{equation}
\label{preservationofdistance}
\gamma=\frac{\textbf{x}^T\textbf{y}/|\textbf{x}|-\bar{\textbf{x}}\bar{\textbf{y}}}{\emph{std}(\textbf{x})\cdot \emph{std}(\textbf{y})}
\end{equation}
where $|\textbf{x}|$ denotes the number of elements in $\textbf{x}$, and $\bar{\textbf{x}}$ and $\emph{std}(\textbf{x})$
denote the mean and standard deviation, respectively. In the ideal case, the normalized correlation equals 1, and the closer
the correlation is to 1, the better the distance in the high dimensional space is preserved in the perceptual color space.

Table~\ref{Preservationofdistance} shows the results of the compared methods in terms of preservation of distance.
Compared with CMF and manifold alignment which also aim to generate natural-looking visualizations, the proposed method has higher score in preserving the spectral distance of pixels in the perceptual color space.
Our method also has competitive performance compared with other methods such as Bilateral filter, Laplacian Eigenmaps, and LPP.

\begin{figure}[!tp]
\centering
\subfigure[RGB image of Washington DC mall from Google Earth.]{
\includegraphics[angle=90,width=0.45\textwidth]{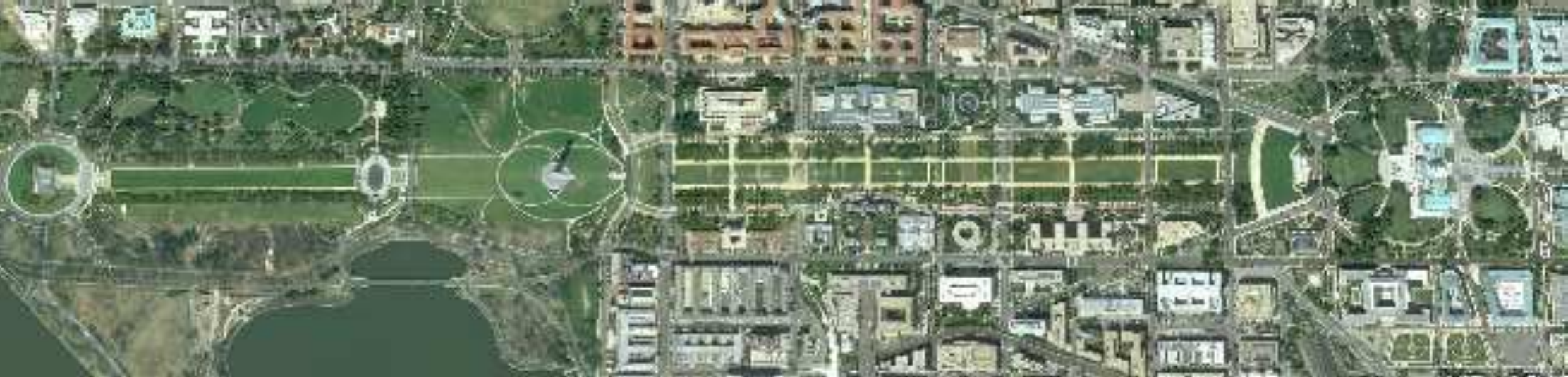}
\label{DCgoogle}
}
\subfigure[Visualization by the proposed instance-level learning.]
{
\includegraphics[angle=90,width=0.45\textwidth]{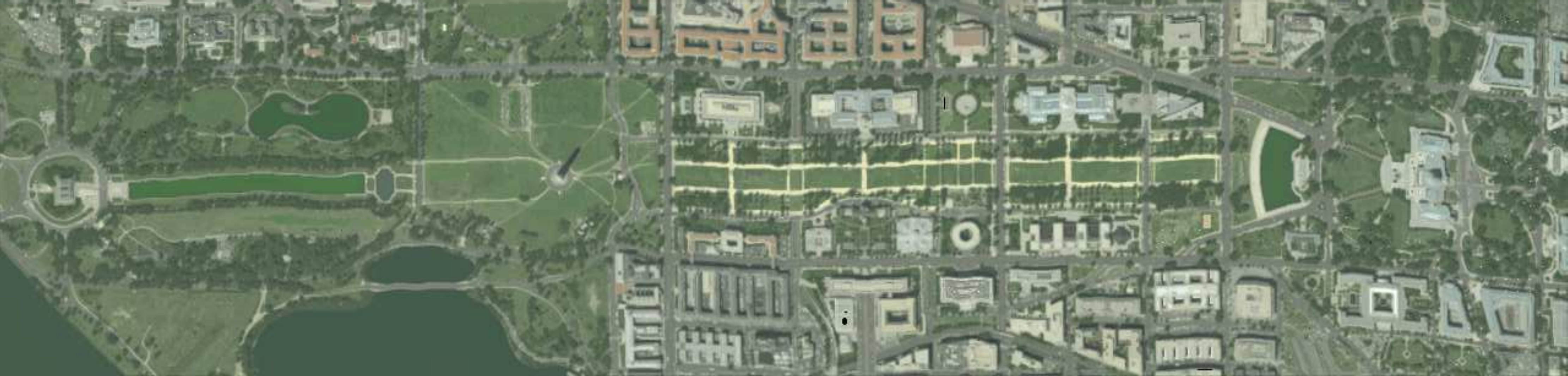}
\label{DCgoogleresult}
}
\subfigure[RGB image of G03 captured by an SLR digital camera.]
{
\includegraphics[width=0.22\textwidth]{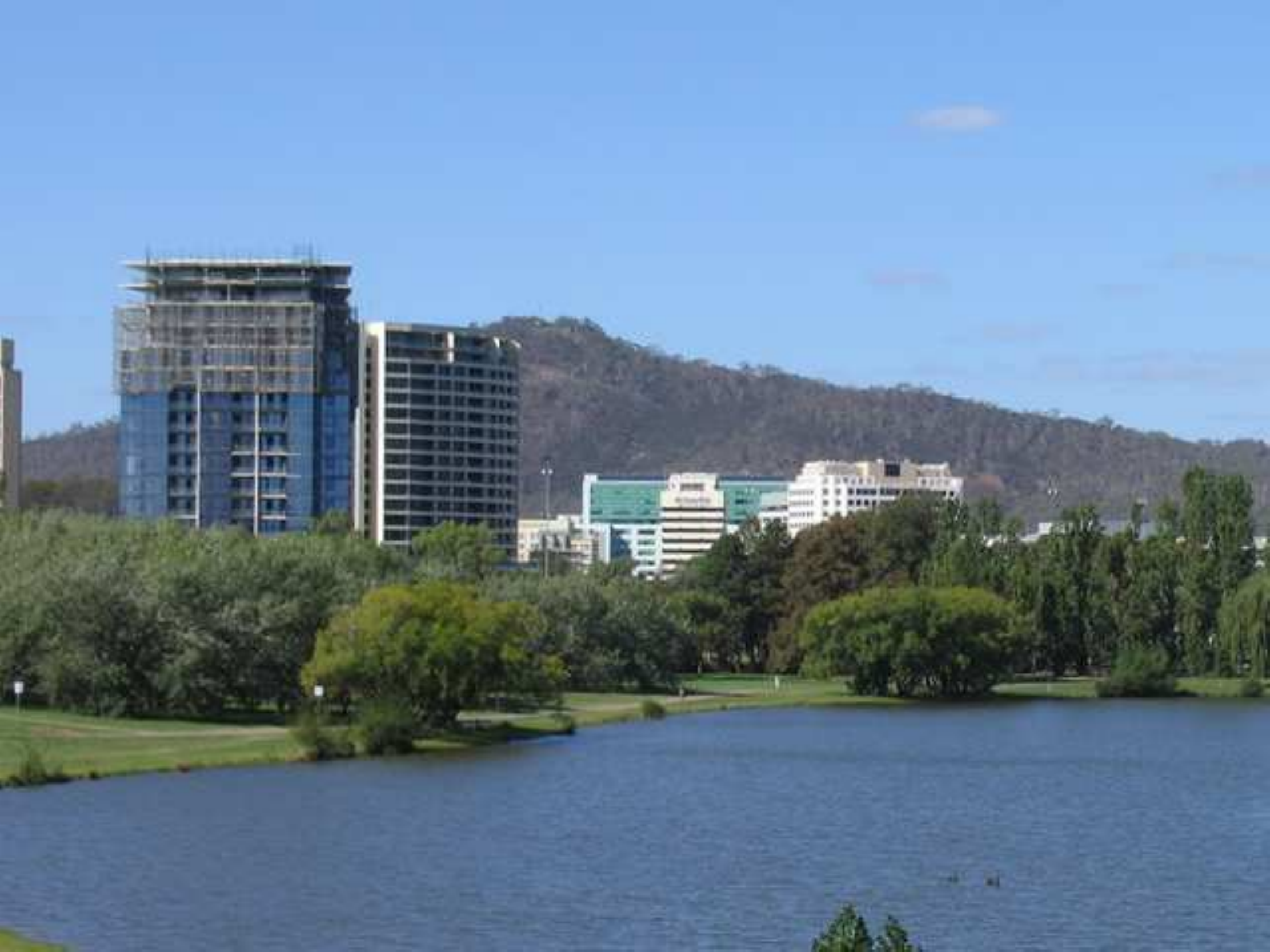}
\label{G03RGB}
}
\subfigure[Visualization by the proposed instance-level learning.]
{
\includegraphics[width=0.22\textwidth]{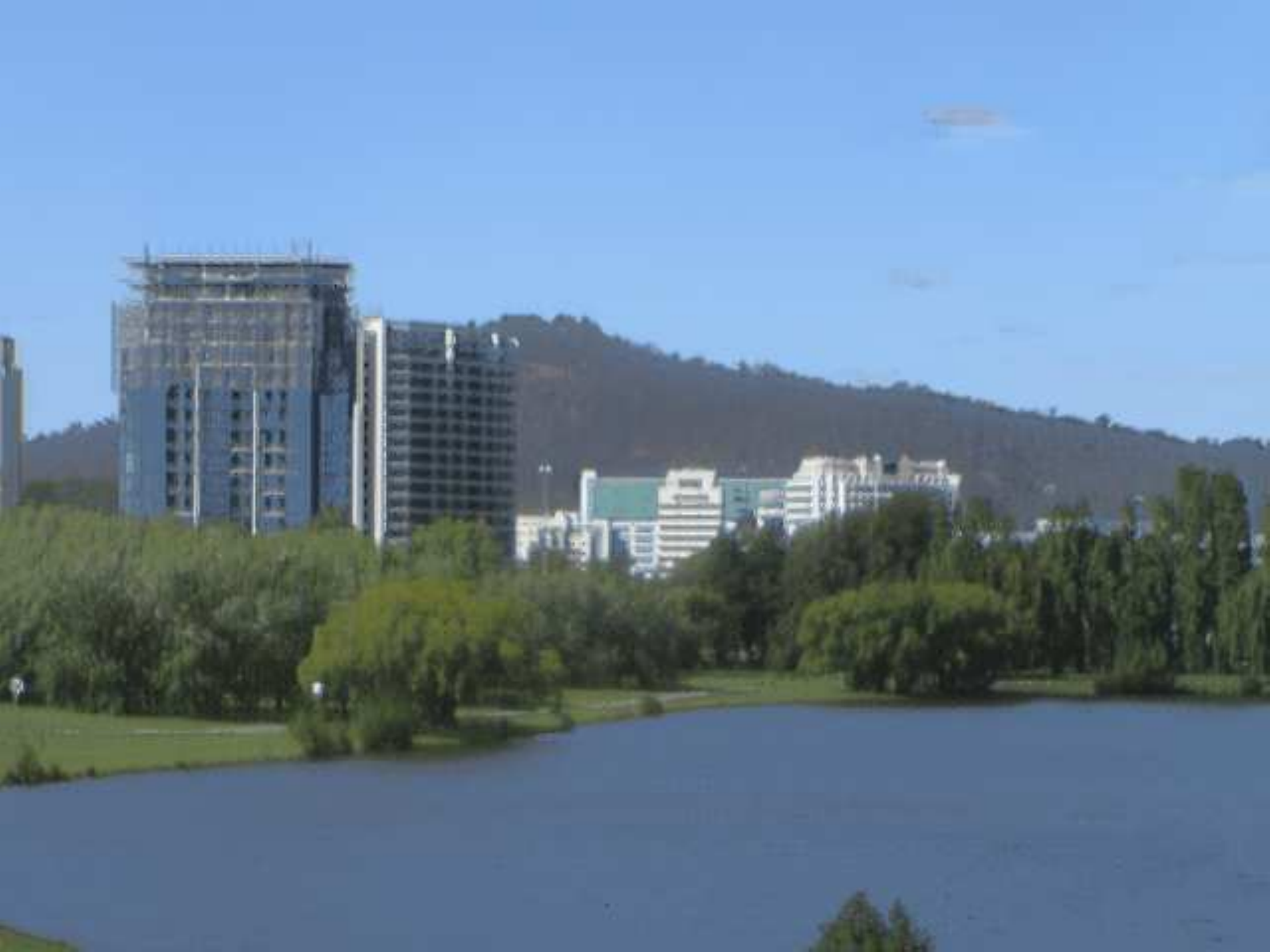}
\label{G03RGBresults}
}
\caption{Visualization with a RGB image captured by a different camera.}\label{fig:differentcameras}
\end{figure}

\begin{figure}
\centering
\subfigure[]{
\includegraphics[width=0.45\linewidth]{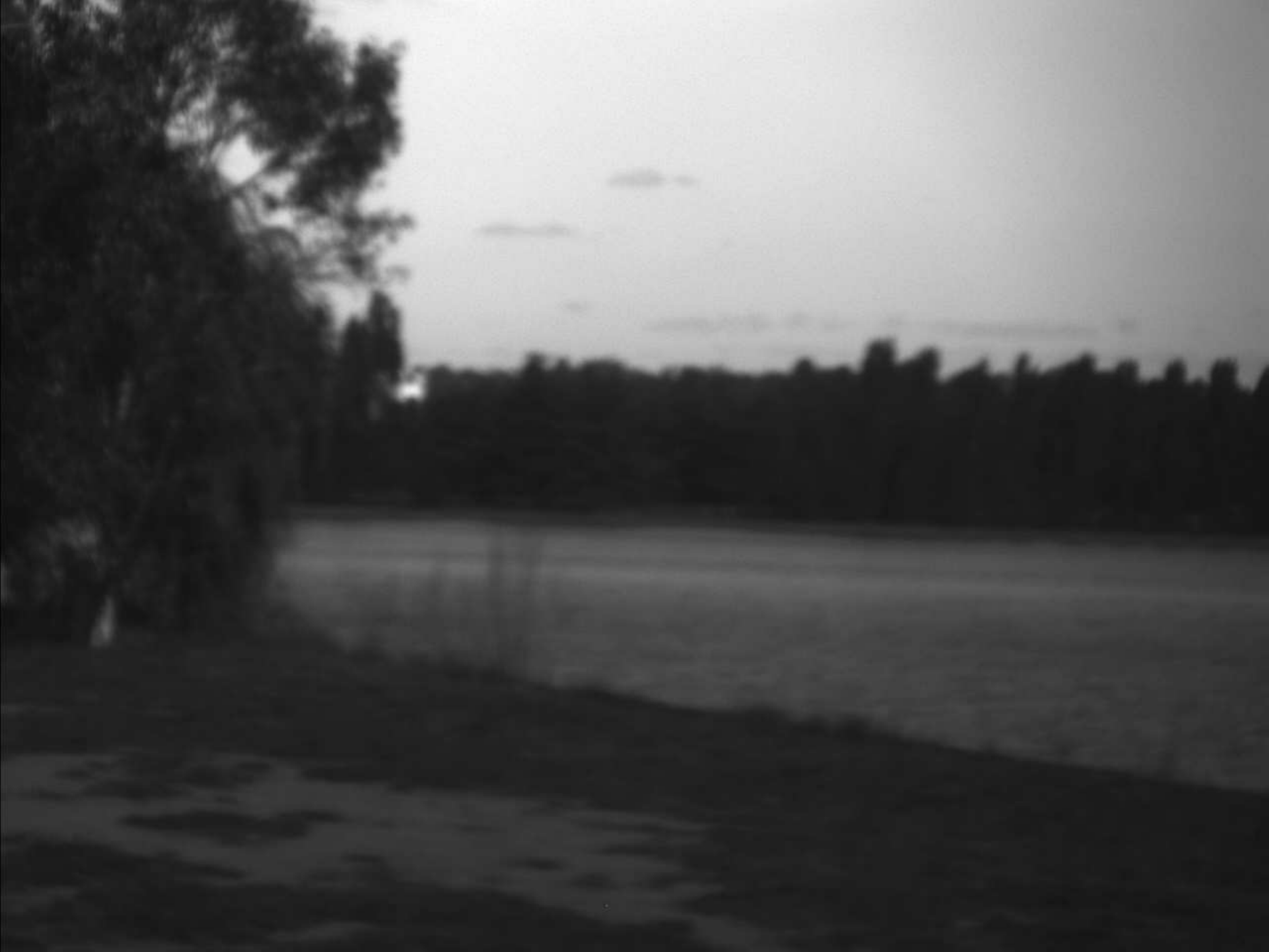}
}
\subfigure[]{
\includegraphics[width=0.45\linewidth]{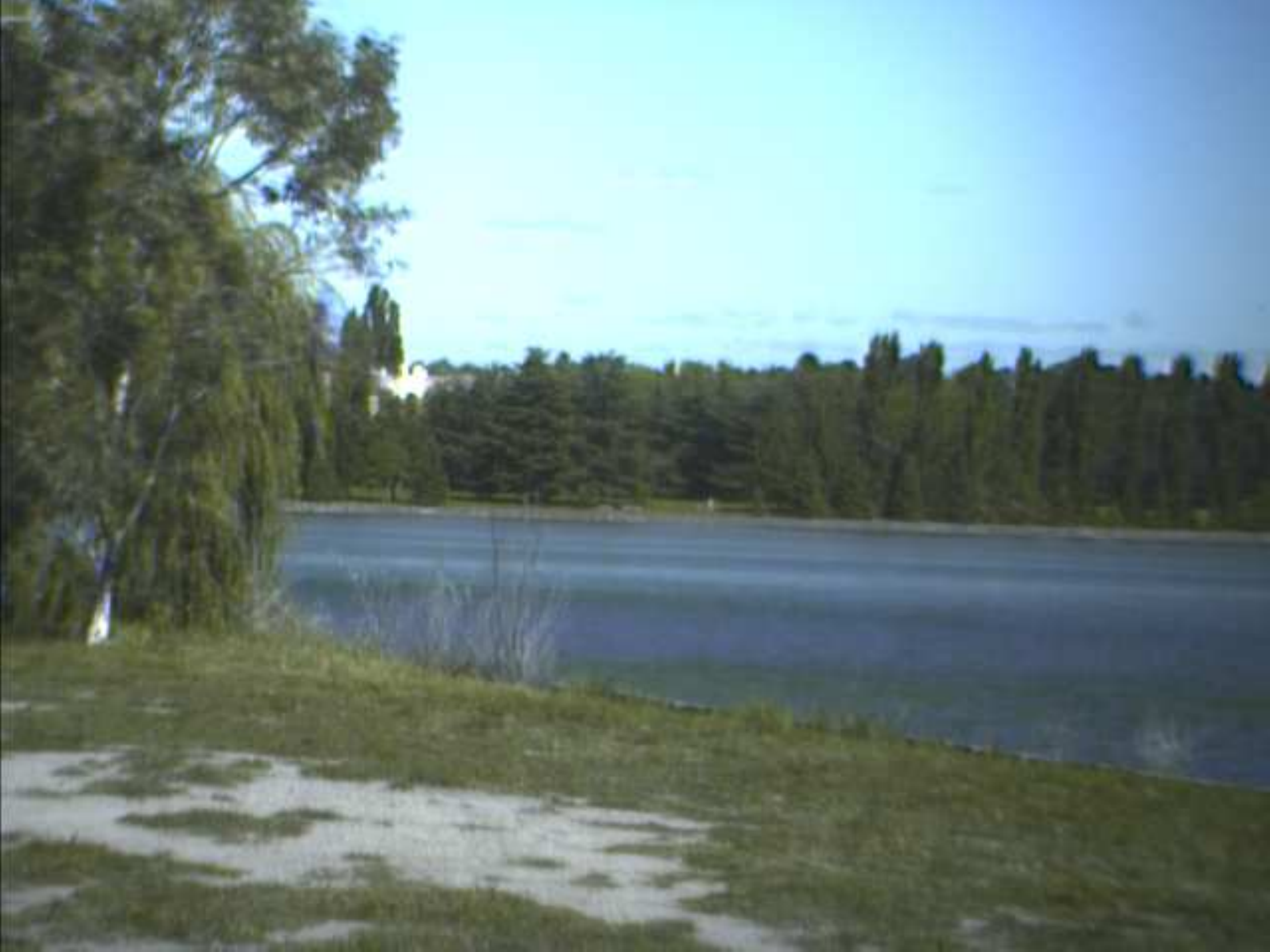}
}
\subfigure[]{
\includegraphics[width=0.45\linewidth]{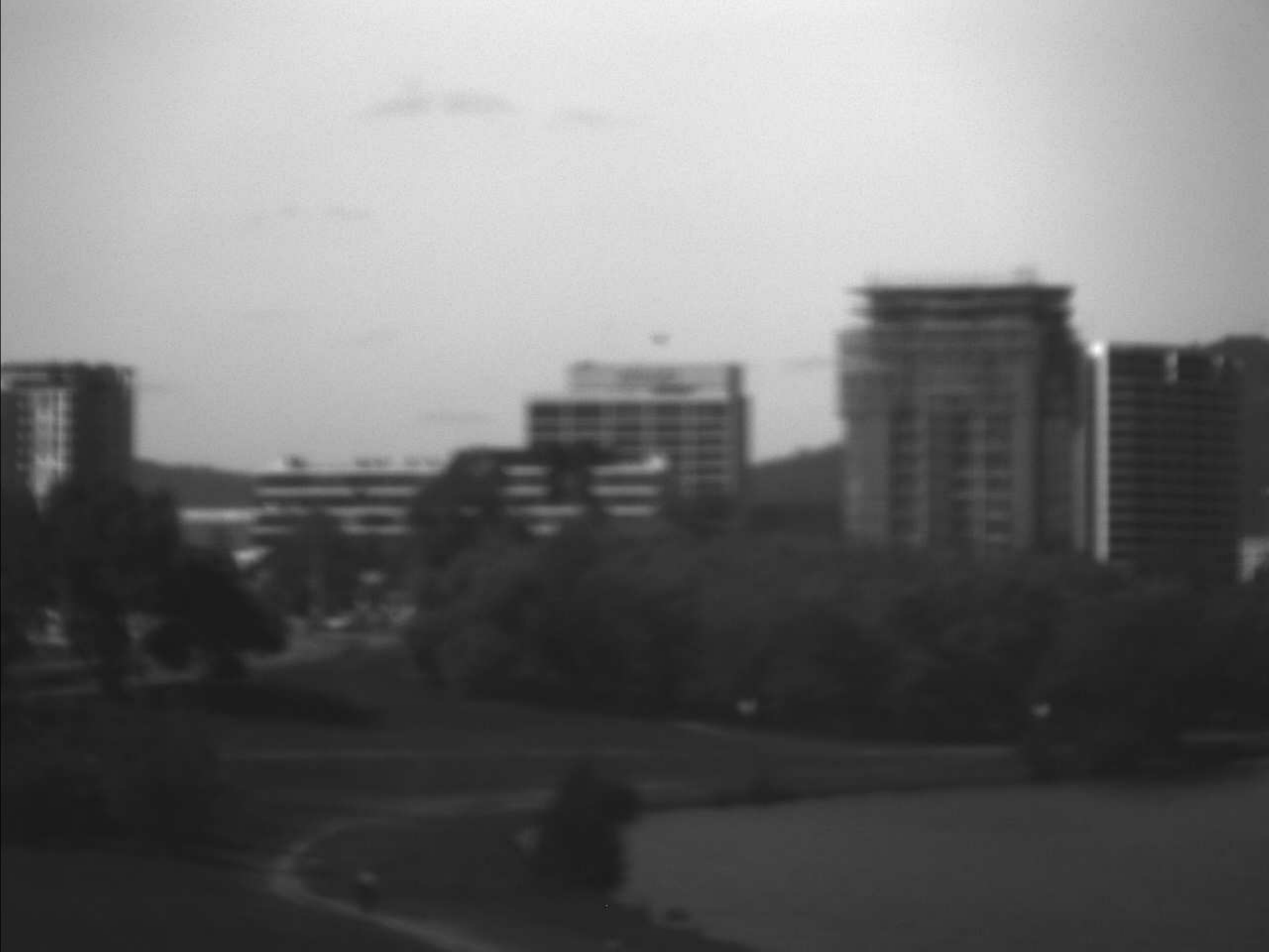}
}
\subfigure[]{
\includegraphics[width=0.45\linewidth]{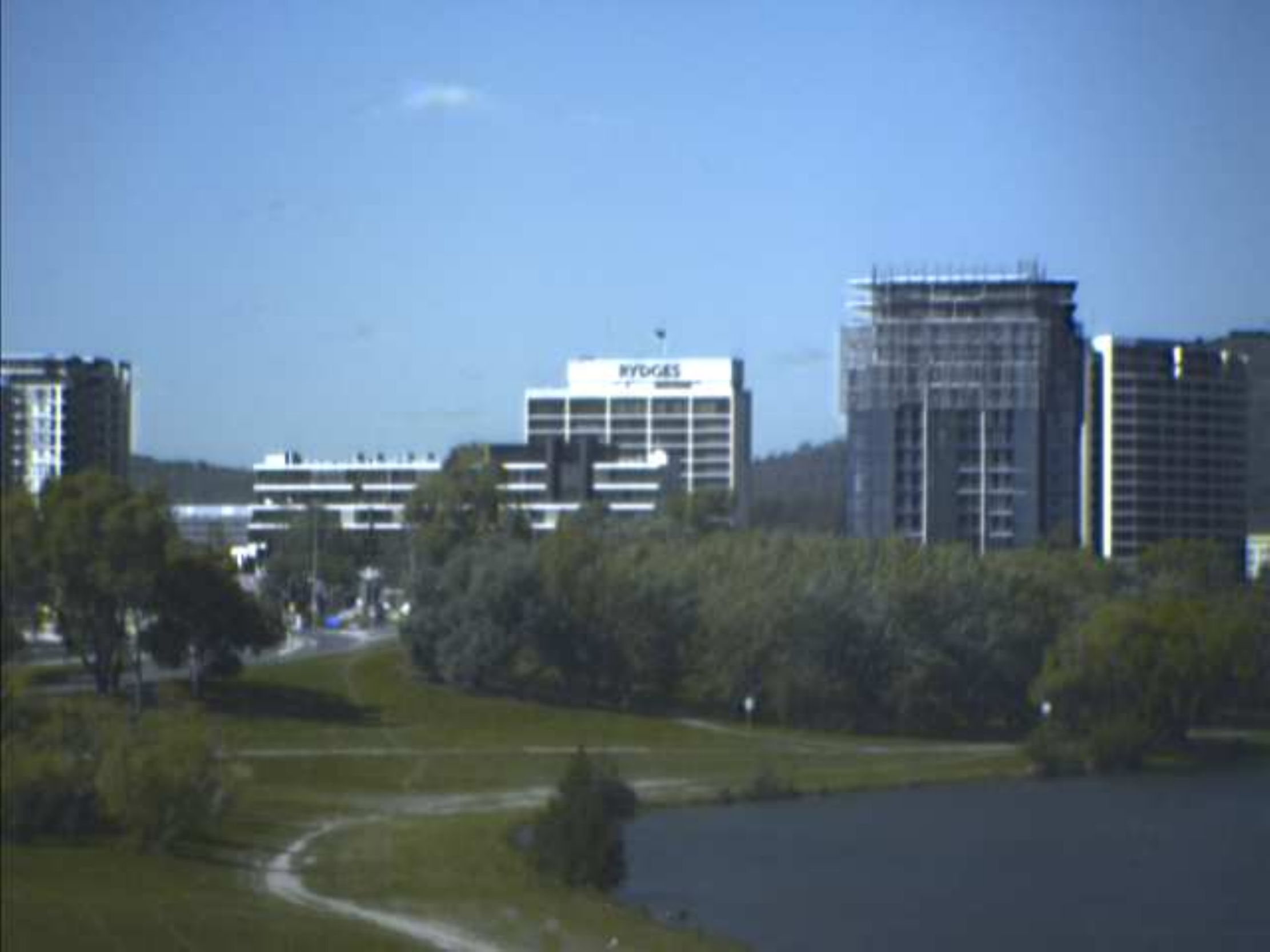}
}
\subfigure[]{
\includegraphics[width=0.45\linewidth]{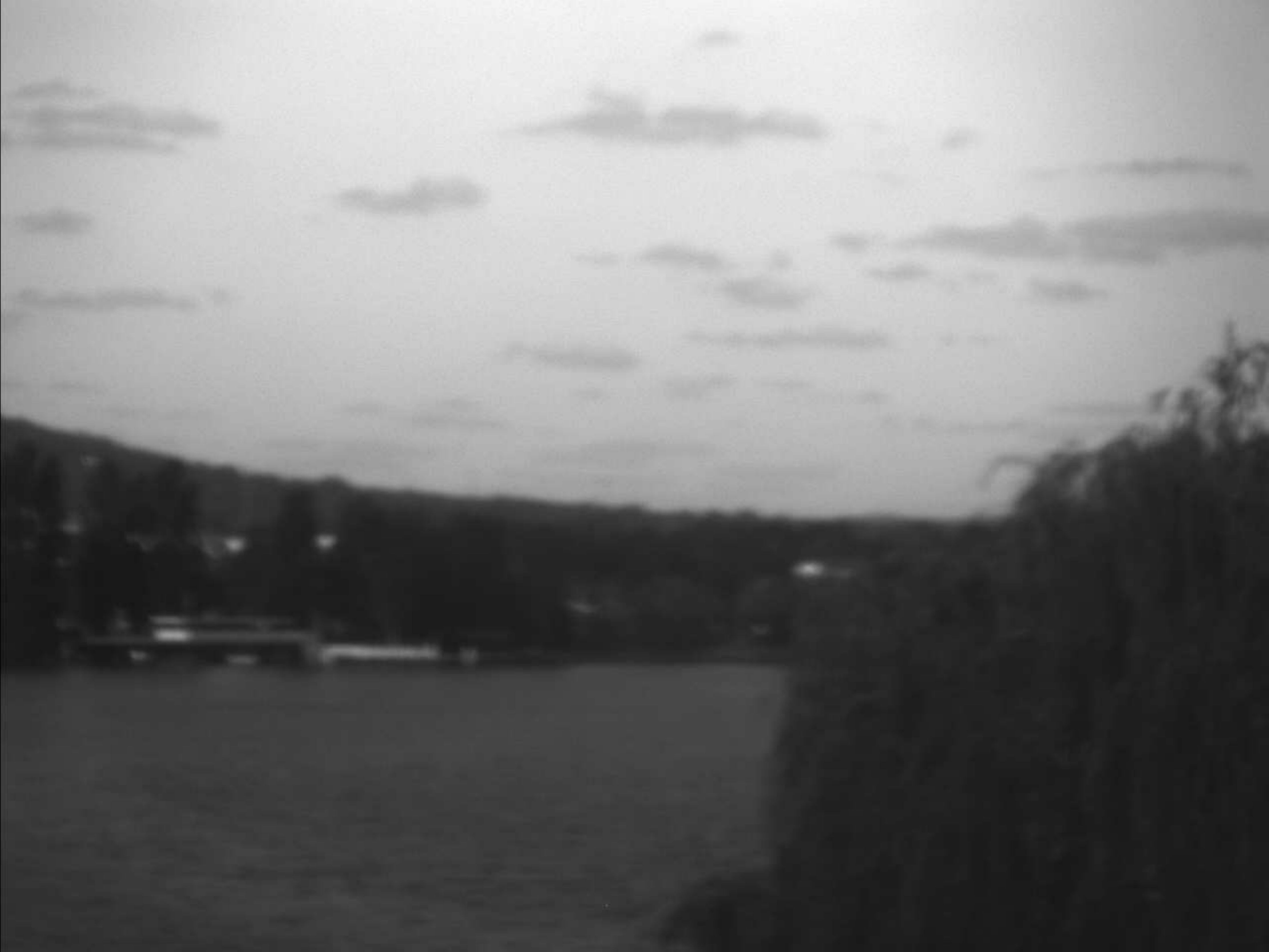}
}
\subfigure[]{
\includegraphics[width=0.45\linewidth]{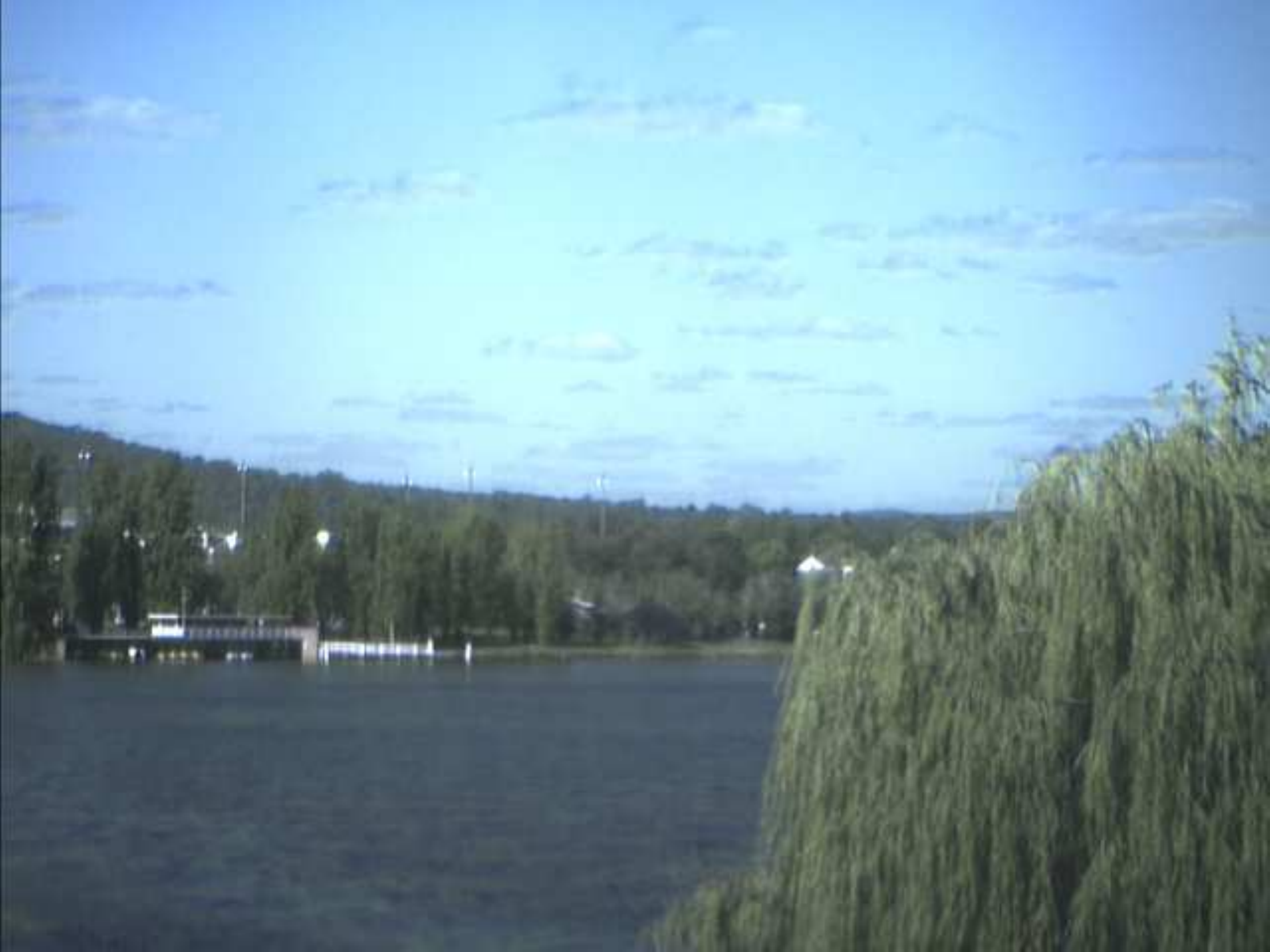}
}
\caption{Visualizations of similar HSIs generated by reusing the projection function of G03.
(a) The first band of F02. (b) The visualization of F02. (c) The first band of G02. (d) The visualization of G02. (e) The first band of G04. (f) The visualization of G04.}
\label{Generalize}
\end{figure}

\begin{figure}[!tp]
\centering
\subfigure[0.1\% matching pairs]{
\includegraphics[width=0.3\linewidth]{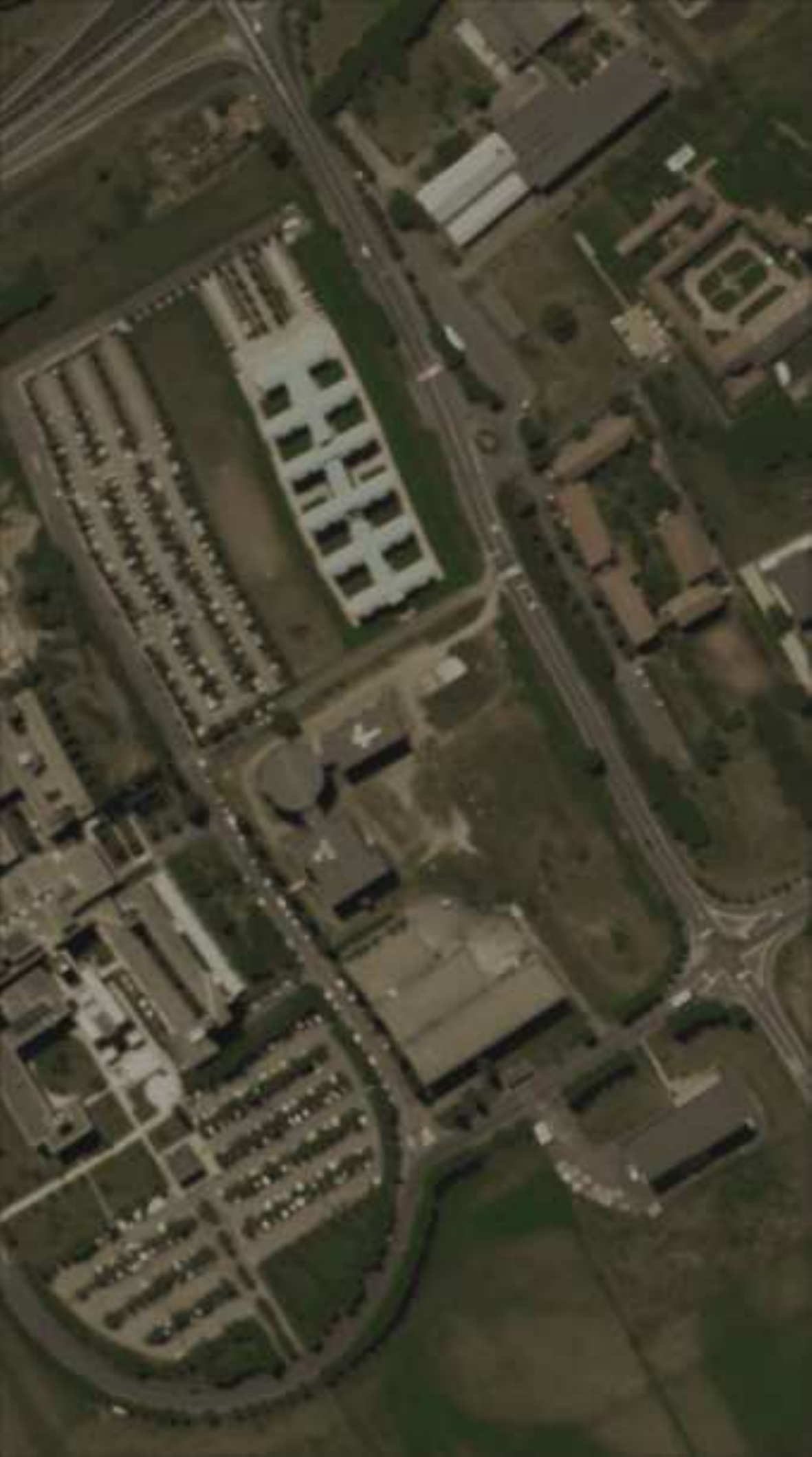}
\label{1matchingnumbers}
}
\subfigure[1\% matching pairs]{
\includegraphics[width=0.3\linewidth]{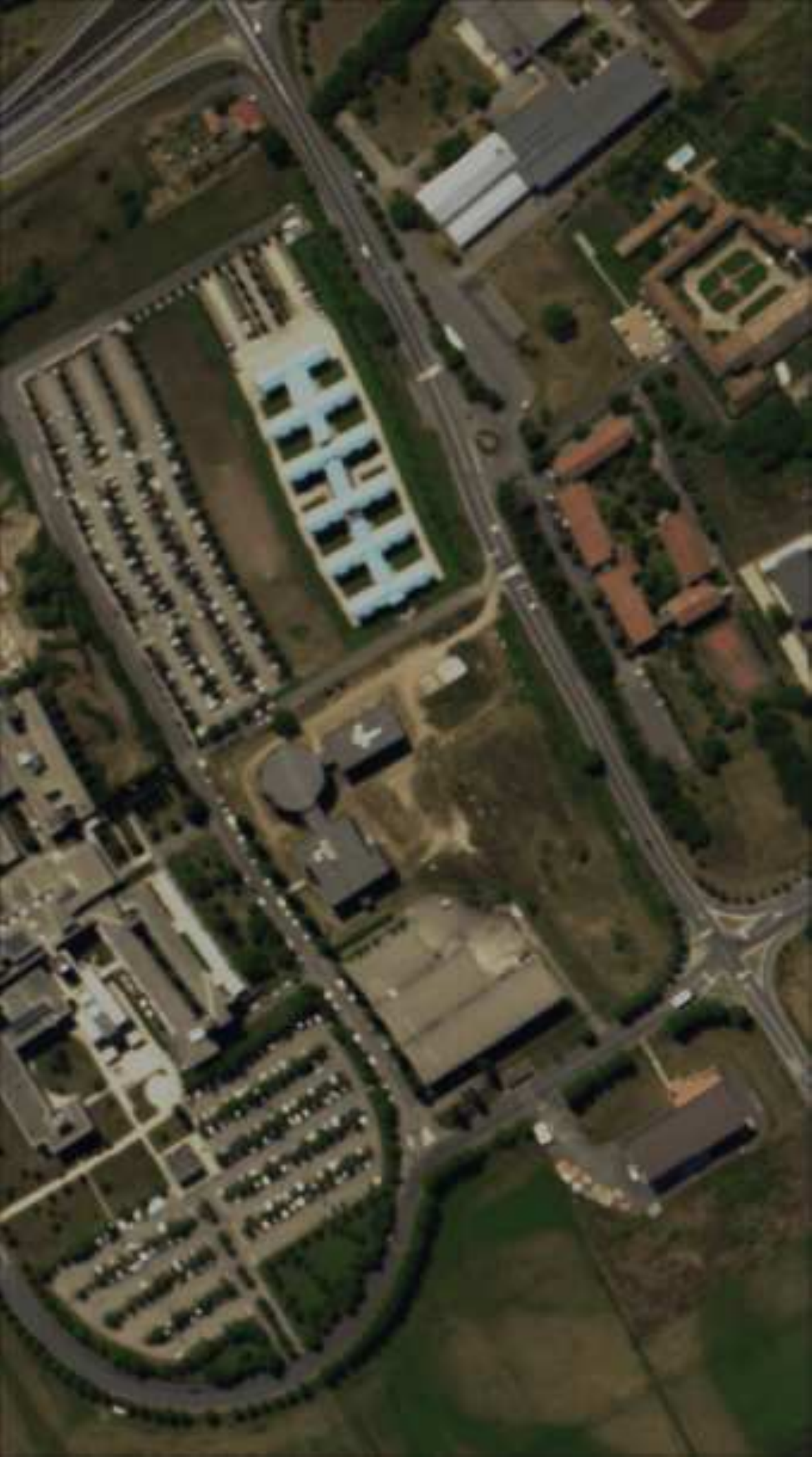}
}
\subfigure[10\% matching pairs]{
\includegraphics[width=0.3\linewidth]{img/paviaU_nonlinear_band_46_31_10_tuned3_blur_step_10spatial_neighbor10_tunedown-eps-converted-to.pdf}
}
\subfigure[100\% matching pairs]{
\includegraphics[width=0.3\linewidth]{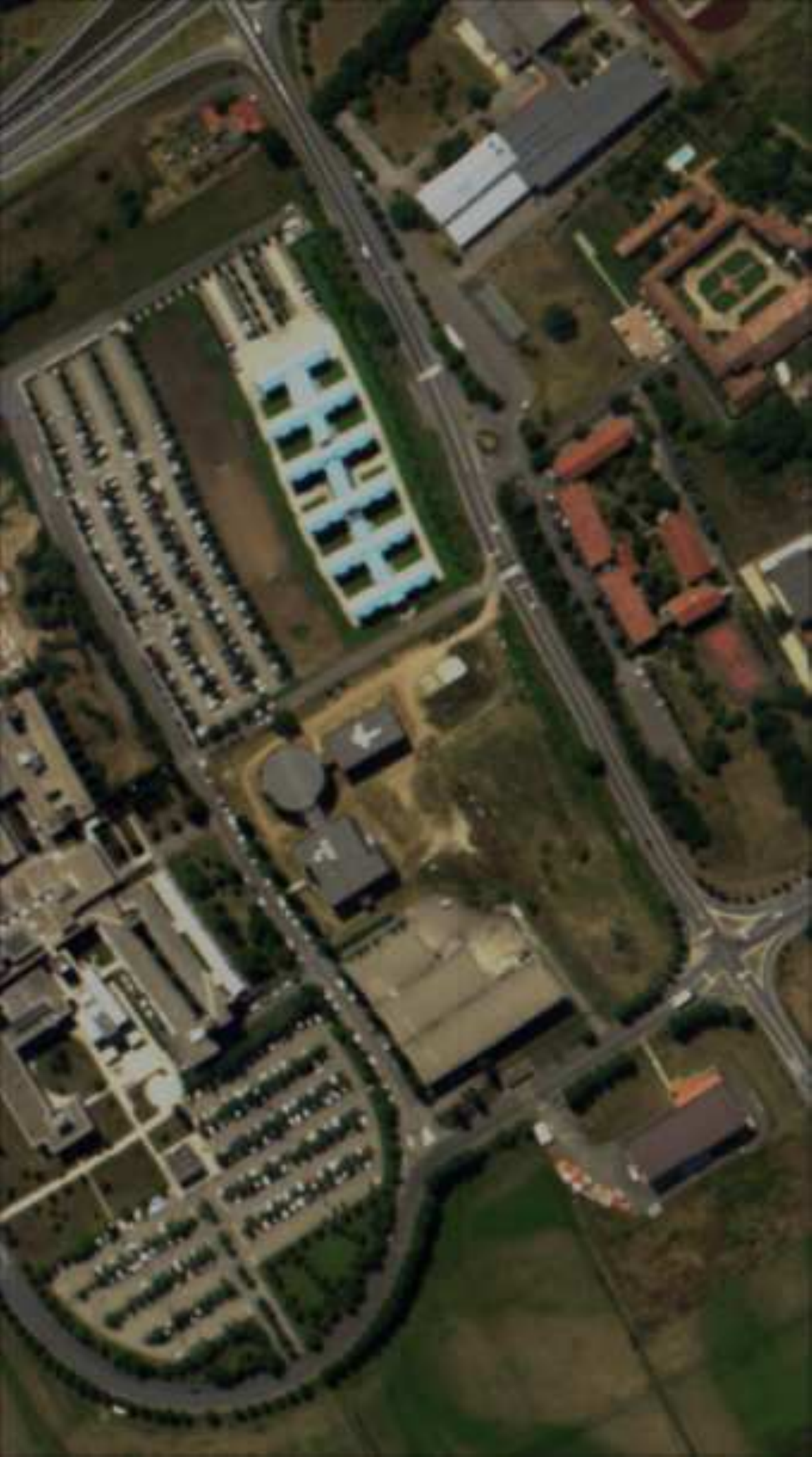}
}
\caption{Visualizations of University of Pavia data set with different numbers of matching pairs between the HSI and the corresponding RGB image.}
\label{differentmatchingnumbers}
\end{figure}

It should be noted that when applied to the Washington DC Mall data set, the manifold learning based methods including Laplacian Eigenmaps, LPP, manifold alignment, and the proposed method have not so good performance in terms of preservation of distance.
One reason we believe is that these manifold learning methods focus on preserving the manifold's local similarity rather than the manifold's global distance distribution, i.e., they can make pixels with similar spectral signatures to be displayed with similar colors, but can not guarantee that pixels with larger spectral differences to be displayed with more distinct colors.
Another reason may be that both the proposed method and the manifold alignment method incorporate the color information of the corresponding RGB image into the visualization while in some cases the pairwise distances between pixels in the RGB images are not consistent with those in the HSI, which leads to the inconsistency of pairwise Euclidean distance between the visualization and the HSI.

\subsection{Running time comparison}
Time costs  of the compared methods are summarized in Table~\ref{runningTime}.
The experiments were carried out on 2.4 GHz Intel Xeon CPU (E5-2680 v4) running Linux with 64GB of RAM.
We can see that manifold learning-based methods (Laplacian Eigenmaps, LPP, manifold alignment and the proposed method) are more time consuming compared with other methods, which is mainly due to the computation of the adjacency matrix $\textbf{W}$ in the high dimensional spectral space.
The proposed feature-level learning is faster than manifold alignment (both are linear models) since it does not need to compute the graph Laplacian of the corresponding RGB image.
Note that once the mapping function from high dimensional spectral space to the color
space is acquired by feature-level learning, the mapping
function can be applied to visualize another HSI captured by the same sensor and this process only takes about 0.02 second.
It should also be noted that the computation of $\textbf{W}$ in manifold learning methods can be efficiently implemented using the parallel
abilities of a graphic processor unit (GPU) and can be greatly accelerated.

\subsection{HSI Visualization with the Corresponding RGB Image Captured by  RGB Camera}
\label{secDifferentCamera}

The corresponding RGB images in the previous experiments were obtained by stacking the R, G, and B channels from the visible wavelength of the HSI.
When the spectral range of an HSI does not cover the visible wavelength,
we can use a RGB image captured on the same or similar site as the HSI through other sources such as a RGB camera mounted on a different platform or the same platform as the HSI sensor.
For example, Fig.~\ref{DCgoogle} shows the RGB image of
Washington DC mall obtained from the Google Earth.
Fig.~\ref{DCgoogleresult} shows the result of the proposed instance-level learning with the color constraint provided by this RGB image.
Another corresponding RGB image (Fig.~\ref{G03RGB}) of the G03 data set was captured by a Nikon D60 SLR digital camera.
The visualization result produced with this image is shown in Fig.~\ref{G03RGBresults}.
We can find that both results have natural colors and are very easy to interpret.

\subsection{Generalizing Projection Function to Visualize Other HSIs}

The proposed feature-level manifold learning method learns an explicit projection function from the high dimensional spectral space to the RGB space.
The projection function can be directly reused to visualize semantically similar HSIs captured by the same hyperspectral imaging sensor.
In this experiment, the projection function derived from
the G03 data set and its corresponding RGB image was applied to
visualize three HSI data sets named ``F02'', ``G02'', and ``G04'', which were captured by the same imaging sensor as G03.
Fig.~\ref{Generalize} shows their first band images and the visualized RGB images.
It can be seen that the visualizations of these HSIs have natural colors after applying the projection function.
Besides, the same objects are presented with consistent colors across these three images.

\subsection{Parameter setting}

There are two main related parameters in our model.
The first one is the number of matching pixel pairs between the HSI and the RGB image, which we denote as $c$.
The second one is the $\lambda$ in Equations~(\ref{lossfunction1}) and~(\ref{lossfunction2}), which is the weight assigned to the color constraint. We empirically set $\lambda = kn/c$ in all our experiments, where $n$ is the number of pixels in the input HSI and $k$ is the number of neighboring pixels in constructing adjacency graph. An intuitive explanation is that when $c$ is small, a relatively large weight is required for the color constraint to be effective. As $c$ grows, the weight of color constraint should be tuned down to better preserve the manifold structure of the HSI.

To analyze the impact of $c$, we did experiments on the University of Pavia data set with different percentage of pixels designated as matching points. Fig.~\ref{differentmatchingnumbers} displays the visualized results of the proposed instance-level learning with different numbers of matching pairs. We can see that when the number of matching pairs is very small ($0.1\%$ as shown in Fig.~\ref{1matchingnumbers}), the colors of the  visualized image is not very desirable. There is no significant visual difference as the number of matching pairs varies from  $1\%$ to $100\%$.
This experiment shows that a small number of matching pairs are sufficient for our approach to generate a fairly good visualization result.

\section{Conclusions}
\label{Conclusions}

A constrained manifold learning approach is proposed to display HSIs with natural colors.
The key idea is to construct a color representation of an input HSI by manifold learning with the color constraint provided by a corresponding RGB image.
We applied a composite kernel in manifold learning to better preserve the image structure of the HSIs.
The proposed manifold learning can be done at instance-level and feature-level.
Instance-level learning can achieve nonlinear embedding results, thus it can preserve more information than feature-level learning, as can be seen from its higher accuracies in the the classification experiments.
Feature-level learning learns a linear mapping function from the high dimensional spectral space to the RGB space. It is faster and the learnt mapping function can be directly applied to visualize similar HSIs captured by the same imaging sensor.
In addition, our methods also perform well in terms of preservation of the spectral distances, especially when comparing with other methods that follow the same goal of representing HSIs in natural colors.

\bibliographystyle{IEEEtran}
\bibliography{IEEEabrv,HSIvisualization}

\begin{IEEEbiography}[{\includegraphics[width=1in,height=1.25in,clip,keepaspectratio]{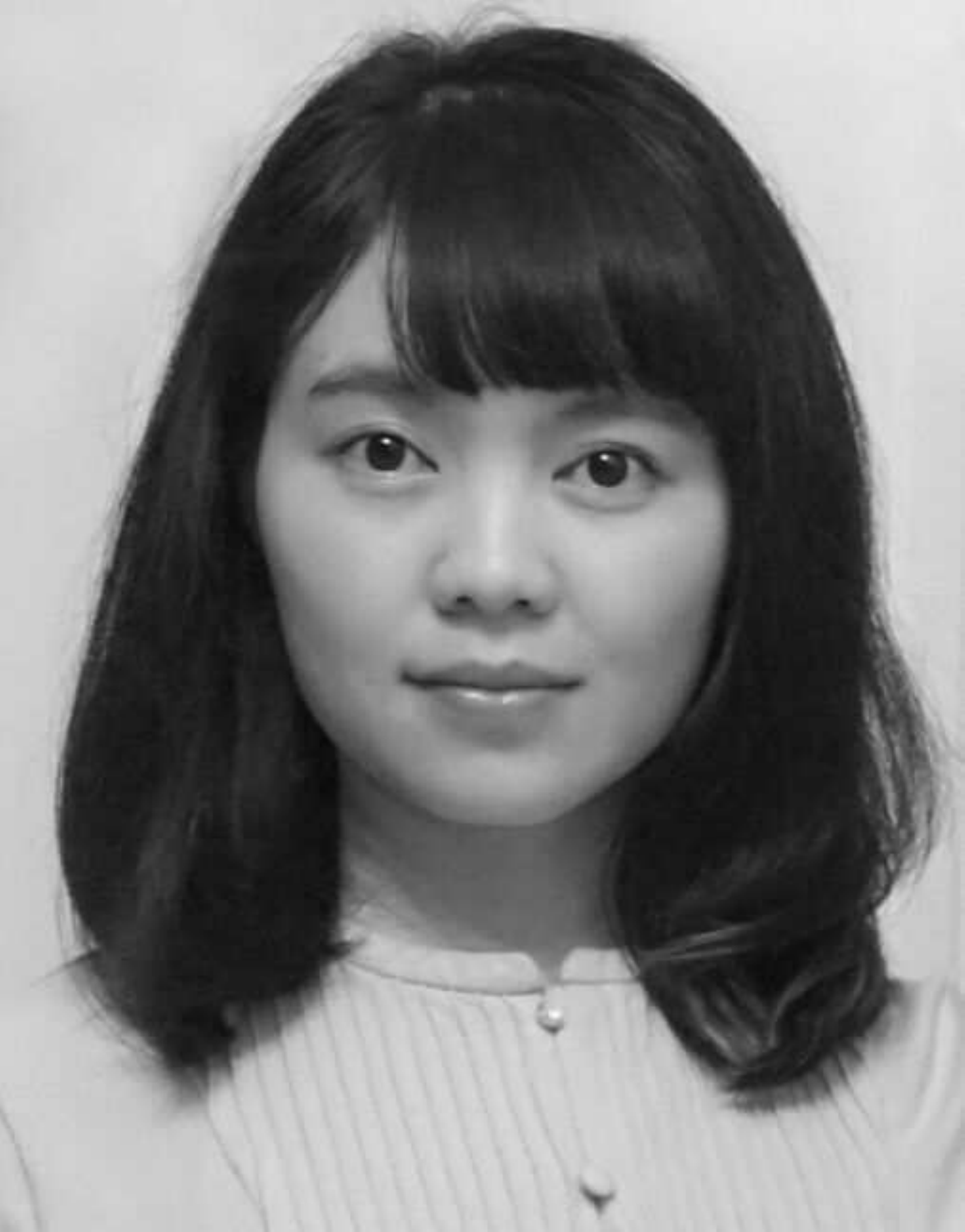}}]{Danping Liao} received her B.E. degree in software engineering from Nankai University, China, in 2011.
She is currently a Ph.D. candidate at College of Computer Science in Zhejiang University, China.
She was a visiting research assistant in the Vision and Media Lab, Simon Fraser University, Canada during 2014 and 2016.
Her research interests include image processing, computer vision and machine learning.
\end{IEEEbiography}
\begin{IEEEbiography}[{\includegraphics[width=1in,height=1.25in,clip,keepaspectratio]{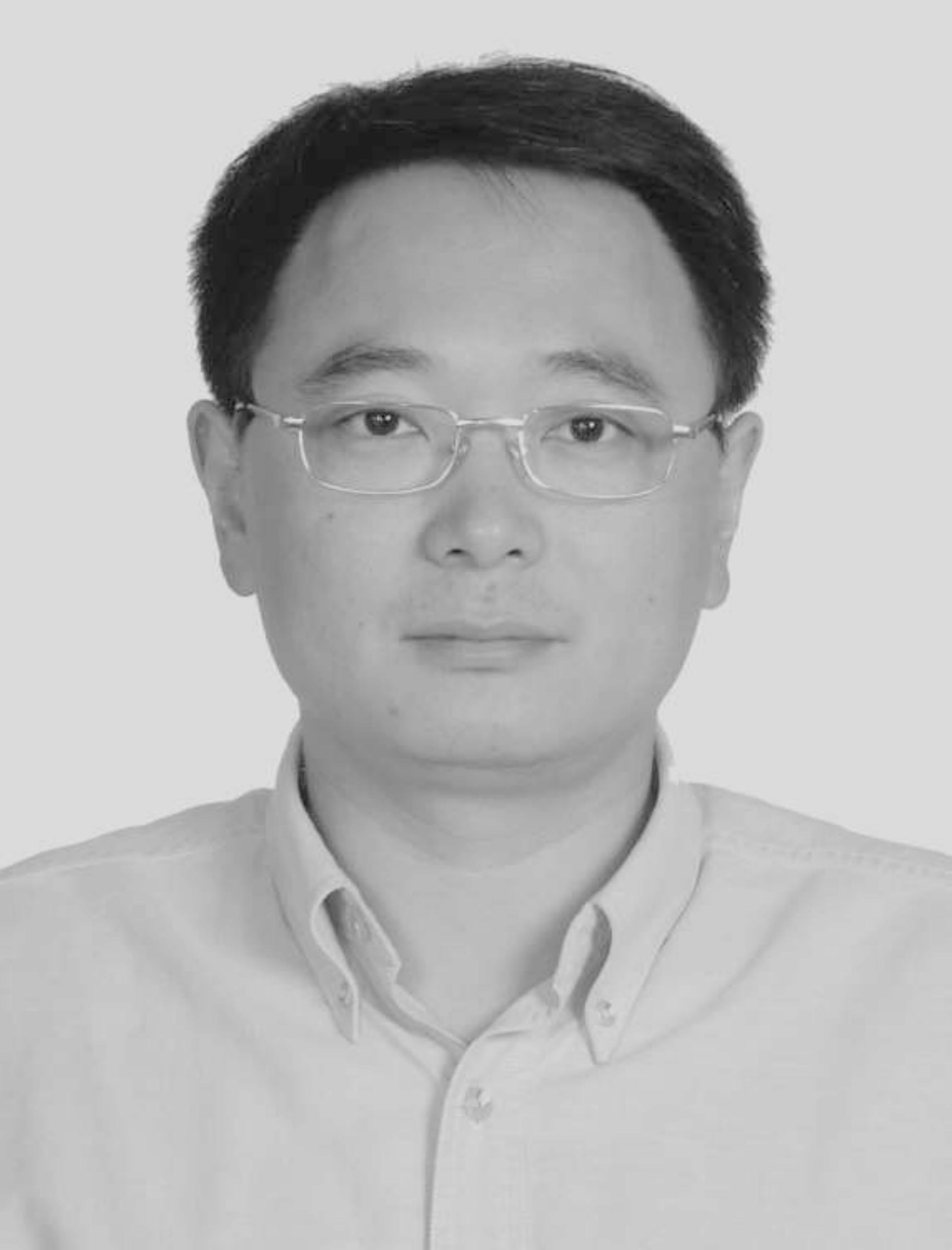}}]{Yuntao Qian} (M'04)
received the B.E. and M.E. degrees in automatic control from Xi'an Jiaotong University, Xi'an, China, in 1989 and 1992, respectively,
and the Ph.D. degree in signal processing from Xidian University, Xi'an, China, in 1996.

During 1996--1998, he was a Postdoctoral Fellow with the Northwestern Polytechnical University, Xi'an, China.
Since 1998, he has been with the College of Computer Science, Zhejiang University, Hangzhou, China,
where he became a Professor in 2002. During 1999--2001, 2006, 2010, 2013, and 2015--2016
he was a Visiting Professor at Concordia University, Hong Kong Baptist University, Carnegie Mellon University,
the Canberra Research Laboratory of NICTA, Macau University, and Griffith University. His current research interests
include machine learning, signal and image processing, pattern recognition, and hyperspectral imaging.

Prof. Qian is an Associate Editor of the \textsc{IEEE Journal of Selected Topics in Applied Earth Observations and Remote Sensing}.
\end{IEEEbiography}
\begin{IEEEbiography}[{\includegraphics[width=1in,height=1.25in,clip,keepaspectratio]{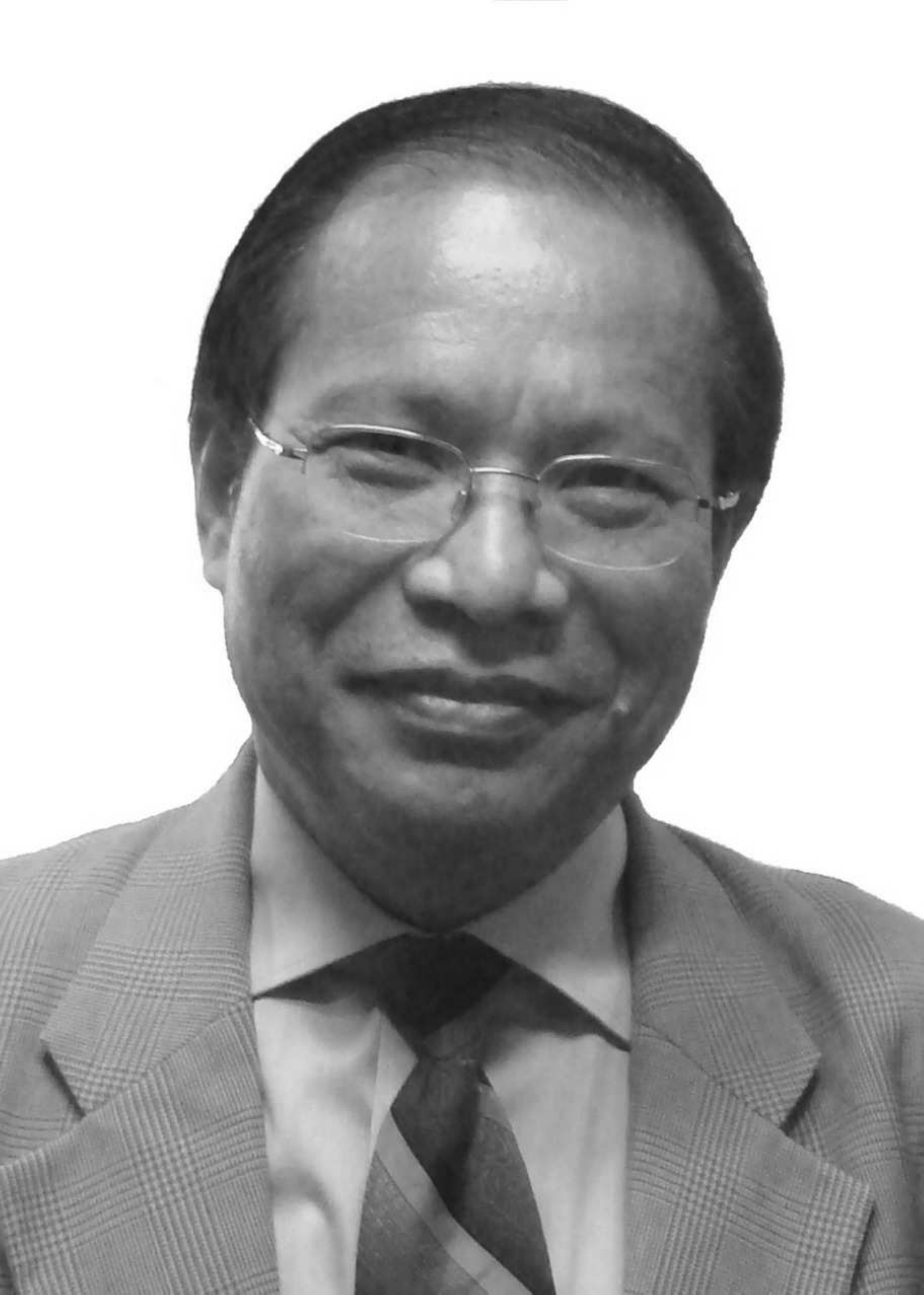}}]{Yuan Yan Tang} (F'04) received the B.S. degree in electrical and computer engineering from Chongqing University, Chongqing, China, the M.S. degree in electrical engineering from the Beijing University of Post and Telecommunications, Beijing, China, and the Ph.D. degree in computer science from Concordia University, Montreal, QC, Canada.

He is currently a Chair Professor with the Faculty of Science and Technology, University of Macau, Macau, China, and a Professor/Adjunct Professor/ Honorary Professor at several institutions including Chongqing University, Chongqing, China, Concordia University, Montreal, QC, Canada, and Hong Kong Baptist University, Hong Kong. He is the Founder and Editor-in-Chief of the International Journal of Wavelets, Multiresolution, and Information Processing and an Associate Editor of several international journals. He has published more than 400 academic papers and is the author/coauthor of over 25 monographs/books/book chapters. His current interests include wavelets, pattern recognition, and image processing. Dr. Tang is a Fellow of the International Association for Pattern Recognition (IAPR). He is the Founder and Chair of the Pattern Recognition Committee of IEEE SMC. He is the Founder and Chair of the Macau Branch of IAPR.

\end{IEEEbiography}

\end{document}